\documentclass[twoside,11pt]{article}

\usepackage{jmlr2e}

\usepackage{algorithm}
\usepackage{algorithmic}
\usepackage{overpic}
\usepackage{subfigure}
\usepackage{url}
\usepackage[colorlinks]{hyperref}
\usepackage{breakurl}
\usepackage{amsmath,amssymb} % define this before the line numbering.
\usepackage{multirow}
\usepackage{overpic}
\usepackage{color}

\def\gx{{g_x}}

\def\gxbf{{g_{\mathbf x}}}
\def\gkbf{{g_{\mathbf k}}}

\def\gvb{{g_{ \mbox{\rm \tiny VB}}}}

\def\grho{g_{ \mbox{\rm \tiny VB}}^} %% have to privide rho after this
\def\hrho{h^} %% have to privide rho after this
\def\psirho{\psi^} %% have to privide rho after this

\def\ffun{\nu} %% have to privide rho after this
\def\gfun{u}

\def\A{{\mathbf A}}
\def\C{{\mathbf C}}

\def\HH{{\mathbf H}}
\def\II{{\mathbf I}}

\def\W{{\mathbf W}}

\def\b{{\mathbf b}}

\def\k{{\mathbf k}}
\def\n{{\mathbf n}}

\def\x{{\mathbf x}}
\def\y{{\mathbf y}}
\def\z{{\mathbf z}}
\def\v{{\mathbf v}}

\def\0{{\mathbf 0}}

\def\GM{{\mathbf \Gamma}}
\def\gam{{\boldsymbol \gamma}}

\newcommand{\myendofproof}[0]{\hfill $\blacksquare$ \newline}

\newtheorem{Thm}{Theorem}
\newtheorem{Def}{Definition}
\newtheorem{Cor}{Corollary}

\ShortHeadings{Revisiting Bayesian Blind Deconvolution}{Wipf and Zhang}
\firstpageno{1}

\begin{document}

\title{Revisiting Bayesian Blind Deconvolution}

\author{\name David Wipf \email davidwipf@gmail.com \\
       \addr Visual Computing Group \\
       Microsoft Research\\
       Beijing, P.R.~China
       \AND
       \name Haichao Zhang \email hczhang1@gmail.com \\
       \addr School of Computer Science \\
       Northwestern Polytechnical University \\
       Xi'an, P.R.~China}

\editor{None}

\maketitle

\begin{abstract}%   <- trailing '%' for backward compatibility of .sty file
Blind deconvolution involves the estimation of a sharp signal or image given only a blurry observation.  Because this problem is fundamentally ill-posed, strong priors on both the sharp image and blur kernel are required to regularize the solution space.  While this naturally leads to a standard MAP estimation framework, performance is compromised by unknown trade-off parameter settings, optimization heuristics, and convergence issues stemming from non-convexity and/or poor prior selections.  To mitigate some of these problems, a number of authors have recently proposed substituting a variational Bayesian (VB) strategy that marginalizes over the high-dimensional image space leading to better estimates of the blur kernel.  However, the underlying cost function now involves both integrals with no closed-form solution and complex, function-valued arguments, thus losing the transparency of MAP.  Beyond standard Bayesian-inspired intuitions, it thus remains unclear by exactly what mechanism these methods are able to operate, rendering understanding, improvements and extensions more difficult.  To elucidate these issues, we demonstrate that the VB methodology can be recast as an unconventional MAP problem with a very particular penalty/prior that couples the image, blur kernel, and noise level in a principled way.  This unique penalty has a number of useful characteristics pertaining to relative concavity, local minima avoidance, and scale-invariance that allow us to rigorously explain the success of VB including its existing implementational heuristics and approximations.  It also provides strict criteria for choosing the optimal image prior that, perhaps counter-intuitively, need not reflect the statistics of natural scenes.  In so doing we challenge the prevailing notion of why VB is successful for blind deconvolution while providing a transparent platform for introducing enhancements and extensions. Experimental results using such derived modifications corroborate our theoretical conclusions.  % reveal improved performance compared with \emph{state-of-the-art} approaches.
\end{abstract}


%%%%%%%%% BODY TEXT
\section{Introduction}

Blind deconvolution problems involve the estimation of some latent sharp signal of interest given only an observation that has been compromised by an unknown filtering process.  Although relevant algorithms and analysis apply in a general setting, this paper will focus on the particular case of blind image deblurring, where an unknown convolution or blur operator, as well as additive noise, corrupt the image capture of an underlying natural scene.  Such blurring is an undesirable consequence that often accompanies the image formation process and may arise, for example, because of camera-shake during acquisition. Blind image deconvolution or deblurring  strategies aim to recover a sharp image from only a blurry, compromised observation, a long-standing problem~\citep{Richardson, Lucy, Kundur96} that remains an active research topic~\citep{Fergus06removingcamera, hqdeblurring_siggraph2008, Understanding_BD, fast_motion_deblur_2009, norm_sparse}.  Moreover, applications extend widely beyond standard photography, with astronomical, bio-imaging, and other signal processing data eliciting particular interest~\citep{Zhu_PAMI, KenigKF10_PAMI}.

Assuming a convolutional blur model with additive noise~\citep{Fergus06removingcamera, hqdeblurring_siggraph2008}, the low quality image observation process  is commonly modeled as
\begin{equation}\label{eq_observation}
\y =  \mathbf{k}\ast \x + \n,
 \end{equation}
where $\mathbf{k}$ is the point spread function (PSF) or blur kernel, $\ast$ denotes the 2D convolution operator, and  $\n$ is a noise term assumed to be zero-mean Gaussian with covariance $\lambda \II$ (although as we shall see, these assumptions about the noise distribution can easily be relaxed via the framework described herein).
The task of blind deconvolution is to estimate both the sharp image $\x$  and blur kernel $\k$ given only the blurry observation $\y$, where we will mostly be assuming that $\x$ and $\y$ represent filtered (e.g., gradient domain) versions of the original pixel-domain images. Because $\mathbf{k}$ is non-invertible, some (typically) high frequency information is lost during the  observation process, and thus even if $\k$ were known, the non-blind estimation of $\x$ is ill-posed.  However, in the blind case where $\k$ is also unknown, the difficulty is exacerbated considerably, with many possible image/kernel pairs explaining the observed data equally well.

To alleviate this problem, prior assumptions must be adopted to constrain the space of candidate solutions, which naturally suggests a Bayesian framework.  In Section~\ref{sec:MAP_vs_MAP_k}, we briefly review the two most common classes of Bayesian algorithms for blind deconvolution used in the literature, (i)~Maximum a Posteriori (MAP) estimation and (ii) Variational Bayes (VB), and then later detail their fundamental limitations, which include heuristic implementational requirements and complex cost functions that are difficult to disentangle.  Section~\ref{sec:EB_BDB} uses ideas from convex analysis to reformulate these Bayesian methods promoting greater understanding and suggesting useful enhancements, such as rigorous criteria for choosing appropriate image priors.  Section \ref{sec:relationship_levin} then situates our theoretical analysis within the context of existing analytic studies of blind deconvolution, notably the seminal work from \citep{Understanding_BD,LevinWDF11_PAMI}, and discusses the relevance of natural image statistics. Learning noise variances is later addressed in Section \ref{sec:learn_lambda}, while experiments are carried out in Section~\ref{sec:Exp} to provide corroborating empirical evidence for some of our theoretical claims.  Finally, concluding remarks are contained in Section~\ref{sec:con}.

\section{MAP versus VB}\label{sec:MAP_vs_MAP_k}

As mentioned above, to compensate for the ill-posedness of the blind deconvolution problem, a strong prior is required for both the sharp image and kernel to regularize the solution space.  Recently, natural image statistics have been invoked to design prior (regularization) models, {e.g.}, \citep{FoE_IJCV, sps_deblur, KrishnanF09_NIPS, Cho_PAMI}, and MAP estimation using these priors has been proposed for blind deconvolution, e.g.,~\citep{hqdeblurring_siggraph2008,norm_sparse}.  While some specifications may differ, the basic idea is to find the mode (maximum) of
\begin{eqnarray}\label{eq:reg_Bayesian_old} \nonumber
\begin{split}
&p(\x, \k|\y) = \frac{p(\y|\x, \k)p(\x) p(\k)}{p(\y)} \propto p(\y|\x, \k)p(\x)p(\k).
% &\propto \exp\left(-\frac{\Vert \y - \k\ast\x\Vert_2^2}{\sigma^2}\right) \exp\left(-\frac{\rho(\x)}{\lambda_1^2}\right) \exp\left(-\frac{\varrho(\k)}{\lambda_2^2}\right).
 \end{split}
\end{eqnarray}
After a $-2\log$ transformation, and ignoring constant factors, this is equivalent to computing
\begin{eqnarray}\label{eq:reg_bd}
\min_{\x, \k} -2\log p(\x, \k|\y) \equiv \min_{\x, \k} \frac{1}{\lambda} \Vert  \mathbf{k} \ast\x -\y \Vert^2_{2} +  \gxbf (\x) + \gkbf (\k),
\end{eqnarray}
where $\gxbf (\x)$ is a penalty term over the desired image, typically of the form $\gxbf (\x) = \sum_i \gx(x_i)$, while $\gkbf (\k)$ regularizes the blur kernel.  Both penalties generally have embedded parameters that must be balanced along with $\lambda$.  It is also typical to assume that $\sum_i k_i = 1$, with $k_i \geq 0$ and we will adopt this assumption throughout (however, Section \ref{sec:other_f} will discuss a type of scale invariance such that this assumption becomes irrelevant anyway).

Although straightforward, there are many problems with existing MAP approaches including ineffective global minima, {e.g.}, poor priors may lead to degenerate solutions like the delta kernel (frequently called the no-blur solution), or too many local minima and subsequent convergence issues.  Therefore, the generation of useful solutions (or to guide the algorithm carefully to a proper local minima) requires a delicate balancing of various factors such as dynamic noise levels, trade-off parameter values, and other heuristic regularizers such as salient structure selection~\citep{hqdeblurring_siggraph2008, fast_motion_deblur_2009,norm_sparse} (we will discuss these issues more in Sections \ref{sec:noise_depend_analysis} and \ref{sec:blur_depend_analysis}).

To mitigate some of these shortcomings of MAP, the influential work by Levin \emph{et al.} and others proposes to instead solve~\citep{Understanding_BD}
\begin{eqnarray}\label{eq:MAP_k}
\begin{split}
\max_{\k} p(\k|\y) \equiv \min_{\k} -2\log p(\y|\k) p(\k),
\end{split}
\end{eqnarray}
where $p(\y|\k) = \int p(\x, \y|\k) d\x$. This technique is sometimes referred to as Type II estimation in the statistics literature.\footnote{To be more specific, Type II estimation refers to the case where we optimize over one set of unknown variables after marginalizing out another set, in our case the image $\x$.  In this context, standard MAP over both $\x$ and $\k$ via (\ref{eq:reg_bd}) can be viewed as Type I.}  Once $\k$ is estimated in this way, $\x$ can then be obtained via conventional non-blind deconvolution techniques.  One motivation for the Type II strategy is based on the inherent asymmetry in the dimensionality of the image relative to the kernel~\citep{Understanding_BD}.  By integrating out (or averaging over) the high-dimensional image, the estimation process can then more accurately estimate the few remaining low-dimensional parameters in $\k$.

The challenge of course with (\ref{eq:MAP_k}) is that the evaluation of $p(\y|\k)$ requires a marginalization over $\x$, which is a computationally intractable integral given realistic image priors.  Consequently a variational Bayesian (VB) strategy is used to approximate the troublesome marginalization~\citep{LevinWDF11}.  A similar idea has also been proposed by a number of other authors~\citep{ensemble_DB, Fergus06removingcamera, BabacanMDK12}.  In brief, VB provides a convenient way of computing a rigorous upper bound on $-\log p(\y|\k)$, which can then be substituted into (\ref{eq:MAP_k}) for optimization purposes leading to an approximate Type II estimator.

% The idea of VB is to find the approximate separable posterior that minimizes the KL divergence from the true posterior.

The VB methodology can be easily applied whenever the image prior $p(\x)$ is expressible as a Gaussian scale mixture (GSM)~\citep{Wipf_VEM_NIPS05}, meaning
\begin{equation}\label{eq:GSM}
%p(\x) =  \int_0^{\infty} \mathcal{N}(\x; \0, \gam^{-1}) d \mu(\gam),
p(\x)  =  \exp \left[-\frac{1}{2}\gxbf(\x)\right] = \prod_i \exp \left[-\frac{1}{2}\gx(x_i)\right]  =   \prod_i \int  \mathcal{N}(x_i;0, \gamma_i) p(\gamma_i) d \gamma_i,
\end{equation}
where each $\mathcal{N}(x_i;0, \gamma_i)$ represents a zero mean Gaussian with variance $\gamma_i$ and prior distribution $p(\gamma_i)$.  The role of this decomposition will become apparent below.  Also, with some abuse of notation, $p(\gamma_i)$ may characterize a discrete distribution, in which case the integral in (\ref{eq:GSM}) can be reduced to a summation.  Note that all prior distributions expressible via (\ref{eq:GSM}) will be supergaussian \citep{Wipf_VEM_NIPS05}, and therefore will to varying degrees favor a sparse $\x$ (we will return to this issue in Sections \ref{sec:EB_BDB} and \ref{sec:relationship_levin}).

Given this $p(\x)$, the negative log of $p(\y|\k)$ can be upper bounded via
\begin{eqnarray}\label{eq:EM} \nonumber
\begin{split}
- \log p(\y|\k) &\le  \underbrace{ -\iint  q(\x,\gam) \log \frac{p(\x, \gam, \y|\k)}{q(\x, \gam)} d\x d\gam}_{F[q(\x,\gam), \k]},
 \end{split}
\end{eqnarray}
where $F[q(\x,\gam), \k]$ is called the \emph{free energy}, $q(\x,\gam)$ is an arbitrary distribution over $\x$, and $\gam = [\gamma_1,\gamma_2,\ldots]^T$, the vector of all the variances from (\ref{eq:GSM}).  Equality is obtained when $q(\x, \gam) = p(\x,\gam|\y,\k)$. In fact, if we were able to iteratively minimize this $F$ over $q(\x,\gam)$ and $\k$ (i.e., a form of coordinate descent), this would be exactly equivalent to the standard expectation-maximization (EM) algorithm for minimizing $- \log p(\y|\k)$ with respect to $\k$, treating $\gam$ and $\x$ as hidden data and assuming $p(\k)$ is flat within the specified constraint set mentioned previously (see  \citep[Ch.9.4]{Bishop} for a detailed examination of this fact).  However, optimizing over $q(\x,\gam)$ is intractable since $p(\x,\gam|\y,\k)$ is generally not available in closed-form.  Likewise, there is no closed-form update for $\k$, and hence no exact EM solution is possible.

The contribution of VB theory is to show that if we restrict the form of $q(\x,\gam)$ via structural assumptions, the updates can now actually be computed, albeit approximately.  For this purpose the most common constraint is that $q(\x,\gam)$ must be factorized, namely, $q(\x,\gam) = q(\x)q(\gam)$, sometimes called a mean-field approximation \citep[Ch.10.1]{Bishop}.  With this approximation we are effectively utilizing the revised (and looser) upper bound
\begin{eqnarray}\label{eq:VB}
\begin{split}
- \log p(\y|\k) &\le -\iint  q(\x) q(\gam) \log \frac{p(\x, \gam, \y|\k)}{q(\x) q(\gam)} d\x d\gam,
%&p(\x, \k|\y) = \frac{p(\y|\x, \k)p(\x) p(\k)}{p(\y)} \propto p(\y|\x)p(\x)p(\k)\\
% &\propto \exp\left(-\frac{\Vert \y - \k\ast\x\Vert_2^2}{\sigma^2}\right) \exp\left(-\frac{\rho(\x)}{\lambda_1^2}\right) \exp\left(-\frac{\varrho(\k)}{\lambda_2^2}\right),
 \end{split}
\end{eqnarray}
which may be iteratively minimized over $q(\x)$, $q(\gam)$, and $\k$ independently while holding the other two fixed.  In each case, closed-form updates are now possible, although because of the factorial approximation, we are of course no longer guaranteed to minimize $- \log p(\y|\k)$.

% This provides a kernel estimate $\k$ that can then be used to estimate $\x$ using conventional non-blind deblurring algorithms.
% To optimize the free energy, we must alternatively minimize  over each variable while holding all others fixed.
Compared to the original Type II problem from (\ref{eq:MAP_k}),  minimizing the bound from  (\ref{eq:VB}) is considerably simplified because the problematic marginalization over $\x$ has been effectively decoupled from $\gam$.  However, when solving for $q(\x)$ at each iteration, it can be shown that a full covariance matrix of $\x$ conditioned on $\gam$, denoted as  $\C$, must be computed.  While this is possible in closed form, it requires $O(m^3)$ operations, where $m$ is the number of pixels in the image.  Because this is computationally impractical for reasonably-sized images, a diagonal approximation to $\C$ must be adopted~\citep{LevinWDF11}.  This assumption is equivalent to incorporating an additional factorization into the VB process such that now we are enforcing the constraint $q(\x,\gam) = \prod_i q(x_i)q(\gamma_i)$. This leads to the considerably looser upper bound
\begin{eqnarray}\label{eq:VB2} \nonumber
\begin{split}
- \log p(\y|\k) &\le  -\iint  \prod_i q(x_i)q(\gamma_i) \log \frac{p(\x, \gam, \y|\k)}{\prod_i q(x_i)q(\gamma_i)} d\x d\gam.
%&p(\x, \k|\y) = \frac{p(\y|\x, \k)p(\x) p(\k)}{p(\y)} \propto p(\y|\x)p(\x)p(\k)\\
% &\propto \exp\left(-\frac{\Vert \y - \k\ast\x\Vert_2^2}{\sigma^2}\right) \exp\left(-\frac{\rho(\x)}{\lambda_1^2}\right) \exp\left(-\frac{\varrho(\k)}{\lambda_2^2}\right),
 \end{split}
\end{eqnarray}
In summary then, the full Type II approach can be approximated by minimizing the VB upper bound via the optimization problem
\begin{equation}\label{eq:VB_DB}
 \min_{q(\x,\gam), \k}  \hspace*{0.2cm} F\left[q(\x,\gam), \k\right],  \hspace*{0.5cm} \mbox{s.t. }  q(\x,\gam) = \prod_i q(x_i)q(\gamma_i).
\end{equation}
The requisite update rules are shown in Algorithm~\ref{algo:algo1}.\footnote{For simplicity we have ignored image boundary effects when presenting the computation for $c_j$ in Algorithm~\ref{algo:algo1}.  In fact, the complete expression for $c_j$ is described in Appendix A in the proof of Theorem \ref{thm:vb_cost_reform}.  Additionally, Algorithm~\ref{algo:algo1} in its present form includes a modest differentiability assumption on $g_x$ for updating the sufficient statistics of $q(\gamma_i)$.}  Numerous methods fall within this category with some implementational differences.  Note also that the full distributions for each $q(x_i)$ and $q(\gamma_i)$ are generally not needed; only certain sufficient statistics are required (certain means and variances, see Algorithm~\ref{algo:algo1}), analogous to standard EM.  These can be efficiently computed using techniques from \citep{Wipf_VEM_NIPS05} for any $p(\x)$ produced by (\ref{eq:GSM}).   In the VB algorithm from \citep{LevinWDF11}, the sufficient statistic for $\gam$ is computed using an alternative methodology which applies only to finite Gaussian scale mixtures.  However, the resulting updates are nonetheless equivalent to Algorithm~\ref{algo:algo1} as shown in the proof of Theorem \ref{thm:vb_cost_reform} presented later.

% Although Algorithm~\ref{algo:algo1} includes an implicit differentiability assumption for updating the sufficient statistics of $q(\gamma_i)$, this is not required.

\begin{algorithm}[tb]
\caption{VB Blind Deblurring~\citep{LevinWDF11,Wipf_VEM_NIPS05,BabacanMDK12}}
\begin{algorithmic}[1]
\STATE {\bf Input:} {a blurry (gradient domain) image $\y$}, noise level reduction factor $\beta$ ($\beta > 1$), minimum noise level $\lambda_0$, an image prior $p(\x) = \exp [-\frac{1}{2}\gxbf(\x)] = \prod_i \exp [-\frac{1}{2}\gx(x_i)]$
%\STATE {\bf Output:} {estimated blur kernel ${\k^*}$ and sharp image ${\boldsymbol{\mu}}$}
\STATE {\bf Initialize:}   blur kernel ${\k}$, noise level $\lambda$
% {\bf Initialization:} sparse vector $\hat{\af}$ recovered from $\y$ in terms of $\D$, and $\hat{\x} = \D\hat{\af}$.\\
\STATE  {\bf While} stopping criteria is not satisfied, do
{
        \begin{itemize}
        \item
            {\bf Update sufficient statistics for $q(\gam)=\prod_i q(\gamma_i)$:}
                 \begin{eqnarray}\nonumber
                 \omega_i \triangleq {\rm E}_{q(\gamma_i)}[\gamma_i^{-1}] \leftarrow \frac{\gx'(\sigma_i)}{2\sigma_i},
                 \end{eqnarray}
              with $\sigma_i^2 \triangleq {\rm E}_{q(x_i)}[x_i^2] = \mu_i^2 + C_{ii}$.
        \item         {\bf Update sufficient statistics for $q(\x)=\prod_i q(x_i)$:}
          \begin{eqnarray}\nonumber
          \begin{split}
            \boldsymbol{\mu}  \triangleq  {\rm E}_{q(\x)}[\x]& \leftarrow  \A^{-1} \b, \hspace*{0.5cm} C_{ii}  \triangleq {\rm Var}_{q(x_i)}[x_i] &\leftarrow A_{ii}^{-1},
          \end{split}
          \end{eqnarray}
          where { $\A = \frac{\HH^T\HH}{\lambda} +{\rm diag} [\boldsymbol{\omega}]$, $\b =  \frac{\HH^T \y}{\lambda}$},
          $\HH$ is the convolution matrix of $\k$.
        \item {\bf Update $\k$:}
        \begin{eqnarray}\label{eq:kernel} \nonumber
 \begin{split}
{\k} &\leftarrow \arg \min_{\k\ge 0} \Vert \y - \W \k\Vert_2^2 +  \sum_{j} c_j k_j^2
 \end{split}
\end{eqnarray}
where {\small $c_j = \sum_i C_{i+j, i+j}$} and $\W$ is the convolution matrix of $\boldsymbol{\mu}$.
%\begin{eqnarray}\label{eq:EM_ker} \nonumber
%\begin{split}
%{\k^*} = \arg \min_{\k} \frac{1}{2} \k^T \A_{\k} \k - \b_{\k}^T \k \quad {\rm s.t.}  \quad \k \ge 0,
%\end{split}
%\end{eqnarray}
%where $\A_{\k}(i_1, i_2) = \sum_i {\x^*}(i+i_1){\x^*}(i+i_2)+\C(i+i_1, i+i_2)$ and $\b_{\k}(i_1) = \sum_i {\x^*}(i+i_1) \y(i)$.
 \item {\bf Noise level reduction:} If $\lambda > \lambda_0$, then $\lambda \leftarrow \lambda/\beta$.
        \end{itemize}

}
\STATE {\bf End}
\end{algorithmic}
\label{algo:algo1}
\end{algorithm}

While possibly well-motivated in principle, the Type~II approach relies on rather severe factorial assumptions which may compromise the original high-level justifications.  In fact, at any minimizing solution denoted $q^*(x_i),q^*(\gamma_i),\forall i, \k^*$, it is easily shown that the gap between $F$ and $-\log p(\y|\k^*)$ is given explicitly by
\begin{equation}
KL\left(\prod_i q^*(x_i)q^*(\gamma_i) || p(\x,\gam|\y,\k^*) \right), % F(\prod_i q(x_i),\prod q(\gamma_i),\k) + \log p(\y|\k) =
\end{equation}
where $KL(p_1 || p_2)$ denotes the standard KL divergence between the distributions $p_1$ and $p_2$. Because the posterior $p(\x,\gam,|\y,\k)$ is generally highly coupled (non-factorial), this divergence will typically be quite high, indicating that the associated approximation is potentially very poor.  We therefore have no reason to believe that this $\k^*$ is anywhere near the maximizer of $p(\y|\k)$, which was the ultimate goal and motivation of Type II to begin with.

Other outstanding issues persist as well.  For example, the free energy cost function, which involves both integration and function-valued arguments, is not nearly as transparent as the standard MAP estimation from (\ref{eq:reg_bd}).  Moreover for practical use, VB depends on an appropriate schedule for reducing the noise variance $\lambda$ during each iteration (see Algorithm~\ref{algo:algo1}), which implements a form of coarse-to-fine, multiresolution estimation scheme \citep{LevinWDF11_PAMI} while potentially improving the convergence rate \citep{LevinWDF11}.  % like most deconvolution algorithms including MAP

It therefore becomes difficult to rigorously explain exactly why VB has often been empirically more successful than MAP in practice (see \citep{BabacanMDK12,LevinWDF11_PAMI,LevinWDF11} for such comparisons), nor how to decide which image priors operate best in the VB framework.\footnote{Note that, as discussed later, certain MAP algorithms can perform reasonably well when carefully balanced with additional penalty factors and tuning paramters added to (\ref{eq:reg_bd}).  However, in direct comparisons using the same basic probabilistic model, VB can perform substantially better, even achieving state-of-the-art performance without additional tuning.}  While Levin \emph{et al.} have suggested that at a high level, marginalization over the latent sharp image using natural-image-statistic-based priors is a good idea to overcome some of the problems faced by MAP estimation~\citep{Understanding_BD,LevinWDF11_PAMI}, this argument only directly motivates substituting (\ref{eq:MAP_k}) for (\ref{eq:reg_bd}) rather than providing explicit rationalization for (\ref{eq:VB_DB}).  Thus, we intend to more meticulously investigate the exact mechanism by which VB operates, explicitly accounting for all of the approximations and assumptions involved by drawing on convex analysis and sparse estimation concepts from \citep{Wipf_VEM_NIPS05,Wipf_Latent_Variable_TIT11} (Section~\ref{sec:relationship_levin} will discuss direct comparisons with Levin \emph{et al.} in detail).  This endeavor then naturally motivates extensions to the VB framework and a simple prescription for choosing an appropriate image prior $p(\x)$.  Overall, we hope that we can further demystify VB providing an entry point for broader improvements such as robust non-uniform blur estimation.

Several surprising, possibly counterintuitive conclusions emerge from this investigation which strongly challenge much of the prevailing wisdom regarding why and how Bayesian algorithms can be advantageous for blind deconvolution.  These include:
\begin{itemize}
\item The optimal image prior for blind deconvolution purposes using VB or MAP is not the one which most closely reflects natural images statistics.  Rather, it is the distribution that most significantly discriminates between blurry and sharp images, meaning a prior such that, for some good sharp image estimate $\hat{\x}$, we have $p(\hat{\x}) \gg p(\k \ast \hat{\x})$.  Natural image statistics typically fail in this regard for explicit reasons, which apply to both MAP and VB, as discussed in Sections \ref{sec:EB_BDB} and \ref{sec:relationship_levin}.
\item The advantage of VB over MAP is not directly related to the dimensionality differences between $\k$ and $\x$ and the conventional benefits of marginalization over the latter.  In fact, we prove in Section \ref{sec:connecting} that the underlying cost functions are formally equivalent in ideal noiseless environments given the factorial assumptions required by practical VB algorithms.  Instead, there is an intrinsic mechanism built into VB that allows bad locally minimizing solutions to be largely avoided even when using the highly non-convex, discriminative priors needed to distinguish between blurry and sharp images.  This represents a completely new perspective on the relative advantages of VB.
\item The VB algorithm can be reformulated in such a way that generic, non-Gaussian noise models and other extensions are easily incorporated, circumventing one important perceived advantage of MAP.
\end{itemize}

While technically these conclusions only apply to the uniform blur model described by (\ref{eq_observation}), many of the underlying principles can nonetheless be applied to more general models.  In fact, we have already obtained success in more complex non-uniform and multi-image models using similar principles, e.g., see \citep{MDB_CVPR13,NBD_inprep}.

\section{Analysis of Variational Bayes}\label{sec:EB_BDB}

Drawing on ideas from \citep{Wipf_VEM_NIPS05,Wipf_Latent_Variable_TIT11}, in this section we will completely reformulate the VB methodology to elucidate its behavior.  More profoundly, we will demonstrate that VB is actually equivalent to using an unconventional MAP estimation-like cost function but with a particular penalty that couples image, blur kernel, and noise in a well-motivated fashion.  This procedure removes the ambiguity introduced by the VB approximation, the subsequent diagonal covariance approximation, and the $\lambda$ reduction heuristic that all contribute still somewhat mysteriously to the effectiveness of VB.  It will then allow us to pinpoint the exact reasons why VB represents an improvement over conventional MAP estimations in the form of (\ref{eq:reg_bd}), and it provides us with a specific criteria for choosing the image prior $p(\x)$.

%================================================
\subsection{Notation and Definitions} \label{sec:notation}

Following~\citep{Fergus06removingcamera} and \citep{LevinWDF11}, we work in the derivative domain of images for ease of modeling and better performance, meaning that $\x$ and $\y$ will now denote the lexicographically ordered image derivatives of sharp and blurry images respectively obtained via a particular derivative filter.  Given that convolution is a commutative operator, the blur kernel is unaltered.

For latent sharp image derivatives of size $M\times N$ and blur kernel of size $P\times Q$,
we denote the lexicographically ordered vector of the sharp image derivatives, blurry image derivatives, and blur kernel as
$\x \in \mathbb{R}^{m}$, $\y \in \mathbb{R}^{n}$ and $\k \in \mathbb{R}^{l}$ respectively, with $m\triangleq MN$, $n\triangleq (M-P+1)(N-Q+1)$, and $l\triangleq PQ$. This assumes a single derivative filter. The extension to multiple filters, typically one for each image dimension, follows naturally.  For simplicity of notation however, we omit explicit referencing of multiple filters throughout this paper, although all related analysis follow through in a straightforward manner.

The likelihood model (\ref{eq_observation}) can be rewritten as
\begin{equation}
\begin{split}
\y & = \HH \x + \n  = \W \k + \n,
\end{split}
 \end{equation}
where $\HH \in \mathbb{R}^{n \times m}$ and $\W \in \mathbb{R}^{n \times l}$   are the convolution matrices constructed from the blur kernel and  sharp image respectively. We introduce a matrix  $\bar{\II} \in \mathbb{R}^{l\times m}$, where the $j$-th row of $\bar{\II}$ is a binary vector with $1$ indicating that the $j$-th element of  $\k$ (i.e. $k_j$) appears in the corresponding column of $\HH$ and $0$ otherwise. We define $\Vert \bar{\k}\Vert_2 \triangleq \sqrt{ \sum_j k_j^2 \bar{I}_{ji} }$, which is equivalent to the norm of the $i$-th column of $\HH$.  It can also be viewed as the effective norm of $\k$ accounting for the boundary effects.\footnote{Technically $\Vert \bar{\k}\Vert_2$ depends on $i$, the index of image pixels, but it  only makes a difference near the image boundaries. We prefer to avoid an explicit notational dependency on $i$ to keep the presentation concise, although the proofs in Appendix A do consider $i$-dependency when it is relevant. The subsequent analysis will also omit this dependency although all of the results carry through in the general case. The same is true for the other quantities that depend on $\Vert \bar{\k}\Vert_2$, e.g., the $\rho$ parameter defined later in (\ref{eq:vb_cost_reform}).
}
The element-wise magnitude of $\x$ is given by $|\x|\triangleq [|x_1|, |x_2|, \cdots]^T$.

Finally we introduce the definition of \emph{relative concavity}~\citep{RC_Palmer} which will serve subsequent analyses:
\begin{Def}\label{def:1}
Let $\gfun$ be a strictly increasing function on $[a, b]$.  The function
$\ffun$ is \textbf{concave relative} to $\gfun$ on the interval $[a,b]$ if and only if
\begin{eqnarray}
\begin{split}
    \ffun(y) &\le \ffun(x) + \frac{\ffun'(x)}{\gfun'(x)} \left[ \gfun(y)-\gfun(x) \right]
     \end{split}
\end{eqnarray}
holds $\forall x,y \in [a, b]$.
\end{Def}

% The notion of {relative concavity} can be understood as follows. The conventional concavity of a function is defined with respect to a linear function.  Specifically,

In the following, we will use $\ffun \prec \gfun$ to denote that $\ffun$ is  concave relative to $\gfun$ on $[0,\infty)$.  This can be understood as a natural generalization of the traditional notion of a concavity, in that a concave function is equivalently \emph{concave relative to a linear function} per Definition \ref{def:1}.  In general, if $\ffun \prec \gfun$, then when $\ffun$ and $\gfun$  are set to have the same  functional value and the same slope at any given point (i.e., by an affine transformation of $\gfun$), then $\ffun$ lies completely under $\gfun$.

%When defined to have the same functional value and slope at any given point, a concave function lies completely below a linear function. In terms of relative concavity, a concave function is \emph{concave relative} to a linear function. This intuition can be transferred to the case of relative concavity seamlessly.

%\begin{figure}
%\centering
%\begin{overpic}[width=7cm]{Lp.eps}
%\end{overpic}
%\caption{An illustration of relative concavity using the \mbox{$\mathcal{\ell}_p$ norm}.}
%\label{fig:Lp}
%\end{figure}

It is well-known that functions concave in $|\x|$ favor sparsity (meaning a strong preference for zero and relatively little distinction between nonzero values) \citep{RaoECPK03,Wipf_Latent_Variable_TIT11}. The notion of relative concavity induces an ordering for many of the common sparsity promoting functions.  Intuitively, a non-decreasing function $\ffun$ of $|x_i|$ is more aggressive in promoting sparsity than some $\gfun$  if it is concave relative to~$\gfun$.  For example, consider the class of $\ell_p$ functionals $\sum_i |x_i|^p$ that are concave in $|x_i|$ whenever $0 < p \leq 1$.  The smaller $p$, the more a sparse $\x$ will be favored, with the extreme case $p \rightarrow 0$ producing the $\mathcal{\ell}_0$ norm (a count of the number of nonzero elements in $\x$), which is the most aggressive penalty for  promoting sparsity.  Meanwhile, using Definition~\ref{def:1} it is easy to verify that, as a function of $|\x|$, $\mathcal{\ell}_{p_1}\prec \mathcal{\ell}_{p_2}$  whenever $p_1<p_2$.

\subsection{Connecting VB with MAP} \label{sec:connecting}

As mentioned previously, the VB algorithm of \citep{LevinWDF11} can be efficiently implemented using any image prior expressible in the form of (\ref{eq:GSM}).  However, for our purposes we require an alternative representation with roots in convex analysis.  Based on \citep{Wipf_VEM_NIPS05}, it can be shown that any prior given by (\ref{eq:GSM}) can also be represented as a maximization over scaled Gaussians with different variances leading to the alternative representation
\begin{equation} \label{eq:convex_prior}
p(x_i)= \exp\left[-\frac{1}{2} g_x(x_i) \right] = \max_{\gamma_i \geq 0} \mathcal{N}(x_i; 0,\gamma_i) \exp\left[-\frac{1}{2} f(\gamma_i) \right],
\end{equation}
\noindent where $f(\gamma_i)$ is some non-negative energy function; the associated exponentiated factor is sometimes treated as a hyperprior, although it will not generally integrate to one.  This $f$, which determines the form of $g_x$ in (\ref{eq:GSM}), will ultimately play a central role in how VB penalizes images $\x$ as will be explored via the results of this section.

\begin{Thm} \label{thm:vb_cost_reform}
Consider the objective function
\vspace*{-0.0cm}
\begin{equation} \label{eq:vb_cost_reform}
\mathcal{L}(\x,\k)  \triangleq \frac{1}{\lambda} \left\|\y - \k \ast \x \right\|_2^2 + \sum_i \gvb(x_i,\rho) + m \log \Vert\bar{\k}\Vert_2^2,
\vspace*{-0.0cm}
\end{equation}
where
\vspace*{-0.0cm}
\begin{equation} \label{eq:general_f_penalty1}
\gvb(x_i,\rho) \triangleq \min_{\gamma_i\ge 0} \left[ \frac{x_i^2}{\gamma_i}  + \log(\rho + \gamma_i ) +  f(\gamma_i) \right], \hspace*{0.1cm} \mbox{\rm s.t. } \rho = \frac{\lambda}{\Vert\bar{\k}\Vert^2_2}.
\end{equation}
\noindent Algorithm 1 minimizing (\ref{eq:VB_DB}) is equivalent to coordinate descent minimization of (\ref{eq:vb_cost_reform}) over $\x$, $\k$, and the latent variables $\gam = [\gamma_1,\ldots,\gamma_m]^T$. % for some penalty function $\gvb$, symmetric about zero, that depends only on the $f$ from (\ref{eq:convex_prior}).
% Moreover, global(local) minima of (\ref{eq:VB_DB}) will correspond with global(local) minima of (\ref{eq:vb_cost_reform}).
\end{Thm}
Proofs will be deferred to the Appendix A.  This reformulation of VB closely resembles (\ref{eq:reg_bd}), with a quadratic data fidelity term combined with additive image and kernel penalties. The penalty on $\k$ in (\ref{eq:vb_cost_reform}) is not unlike those incorporated into standard MAP schemes, meaning $\gkbf (\k)$ from (\ref{eq:reg_bd}).  However, quite unlike MAP, the penalty $\gvb$ on the image pixels $x_i$ is dependent on both the noise level $\lambda$ and the kernel $\k$ through the parameter $\rho$, the ratio of the noise level to the squared kernel norm. Moreover, with a general $\lambda \neq 0$, it is easily shown that the function $\gvb$ is non-separable in $\k$ and $\x$, meaning $\gvb(x_i,\rho) \neq h_1(x_i) + h_2(\k)$ for any possible functions $h_1$ and $h_2$.  The remainder of Section~\ref{sec:EB_BDB} will explore the consequences of this crucial, yet previously unexamined distinction from typical MAP formulations.

In contrast, with $\lambda = 0$, both MAP and VB possess a formally equivalent penalty on each $x_i$ via the following corollary:
\begin{Cor} \label{cor:vb_cost_reform}
If $\lambda = 0$, then $\gvb(x_i,0) = \gx(x_i) \equiv  -2\log p(x_i)$.
%If $\lambda = 0$, then $\gvb(x_i,0) = -\log p(x_i)$.
\end{Cor}
\noindent Therefore the underlying VB cost function \emph{is effectively no different than regular MAP from (\ref{eq:reg_bd}) in the noiseless setting}, a surprising conclusion that seems to counter much of the prevailing understanding of VB deconvolution algorithms.

A simple 1D example described next will serve to motivate why VB can, perhaps paradoxically, still outperform MAP even in ideal noiseless scenarios.  In brief, different solutions between VB and MAP are still possible here because a decreasing sequence of $\lambda$ is used to optimize both techniques, and this may lead to a radically different optimization trajectory terminating at different locally minimizing solutions.   Later, Sections \ref{sec:implicit_penalty}-\ref{sec:blur_depend_analysis} will provide rigorous analysis of $\gvb$, including how it may affect convergence paths, leading to several insights regarding VB performance.  Finally, Section~\ref{sec:other_f} will address the issue of choosing the optimal image prior $p(\x)$, which is equivalent to choosing the optimal $f$ in (\ref{eq:convex_prior}).

\subsection{Illustrative Example using 1D Signals}
\label{sec:1D_exp}

Here we will briefly illustrate the distinction between MAP and VB where other confounding factors have been conveniently removed.  For this purposed we consider a simplified noiseless situation where the optimal $\lambda$ value is zero.  Based on Corollary \ref{cor:vb_cost_reform} we know that whenever $\lambda$, and therefore $\rho$, goes to zero, the MAP and VB cost functions become exactly equivalent when given the same prior selection $p(\x)$, and thus share the same globally optimal solution in the limit $\lambda \rightarrow 0$.

As mentioned in Section \ref{sec:MAP_vs_MAP_k} and shown in Algorithm \ref{algo:algo1} however, essentially all VB and MAP deconvolution algorithms begin with a large value of $\lambda$ and gradually reduce it towards some minimal value as the iterations proceed as part of a multi-resolution approach designed to find good solutions. While the underlying cost functions may be equivalent when $\lambda = 0$, they behave very differently for $\lambda > 0$, and we will argue extensively that here is where the ultimate advantage of VB lies.  Although details will be deferred to later sections, the embedding of $\k$ and $\lambda$ into $\gvb$ leads to a powerful adjustment of the image penalty curvature during each iteration, smoothing out local minima (especially at the beginning of the estimation process) such that bad local solutions can largely be avoided.  In contrast, MAP employs a static image penalty that is easily lured into suboptimal basins of attraction.  Thus even if the global minima are the same when eventually $\lambda \rightarrow 0$, we may expect that VB has a better chance of reaching this solution.

We test this conclusion using perhaps the simplest case where $f$ is constant, i.e., $f(\gamma) = b$ (later Section~\ref{sec:other_f} will argue that this selection is in some sense optimal).  With this assumption, the associated MAP problem from (\ref{eq:reg_bd}) is easily shown to be
\begin{equation} \label{eq:vb_cost_MAP}
\min_{\x,\k}  \frac{1}{\lambda} \left\|\y - \k \ast \x \right\|_2^2 + \sum_i 2\log\left|x_i\right| + 2m\log \Vert\bar{\k}\Vert_2.
\end{equation}
\noindent where the image penalty is obtained by applying a $-2\log$ transformation to (\ref{eq:convex_prior}) giving
\begin{equation}
 -2\log \left[ \max_{\gamma_i \geq 0} \mathcal{N}(x_i; 0,\gamma_i)  \right]  \equiv 2\log\left|x_i\right|.
\end{equation}
\noindent Irrelevant additive constants have been excluded, and we are assuming the same kernel penalty as VB in (\ref{eq:vb_cost_reform}). In the limit as $\lambda \rightarrow 0$, based on the equivalency derived from Corollary~\ref{cor:vb_cost_reform}, both VB and MAP are effectively solving
\begin{equation} \label{eq:vb_cost_MAP2}
\min_{\x,\k} \sum_i \log\left|x_i\right| + m \log \Vert\bar{\k}\Vert_2, \hspace*{0.3cm} \mbox{s.t. } \y = \k \ast \x.
\end{equation}
Moreover, given the stated assumptions $\sum_i k_i = 1, k_i \geq 0$ and arguments made in Section \ref{sec:implicit_penalty} below, it can be shown that: (i) the $\log \Vert\bar{\k}\Vert_2$ term is no longer relevant for determining the globally optimal solution (assuming that the optimal $\x$ is actually sparse), and (ii) the remaining penalty on $\x$ reduces to the $\ell_0$ norm,
which represents a count of the nonzero elements in $\x$ and is sometimes considered as the canonical metric for quantifying sparsity.  Thus, (\ref{eq:vb_cost_MAP2}) effectively reduces to
\begin{equation} \label{eq:vb_cost_MAP3}
\min_{\x,\k} \|\x\|_0, \hspace*{0.3cm} \mbox{s.t. } \y = \k \ast \x.
\end{equation}
Therefore at this simplified, stripped-down level both VB and MAP are merely minimizing the $\ell_0$ norm of $\x$ subject to the linear convolutional constraint.

Of course we do not attempt to solve (\ref{eq:vb_cost_MAP3}) directly, which is a difficult combinatorial problem in nature.  Instead for both VB and MAP we begin with a large $\lambda$ and gradually reduce it towards zero as described in Algorithm \ref{algo:algo1}, where we use $\beta = 1.15$ (this value is taken from Levin \emph{et al.}~\citep{LevinWDF11}) for updating $\lambda$.\footnote{The MAP algorithm can be implemented by simply setting $\C$ to zero before the $q(\gamma_i)$ update in Algorithm~\ref{algo:algo1}, with guaranteed convergence to some local minima.  For both MAP and VB, the $\gamma$ sufficient statistic update is simply $\omega_i = \sigma_i^{-2}$ whenever $f$ is a constant.}  Before $\lambda$ becomes small, the VB and MAP cost functions will behave very differently, since VB is based on the coupled image penalty $\gvb$ in (\ref{eq:vb_cost_reform}) while MAP employs a simple $\sum_i \log|x_i|$ factor.  The superiority of the VB convergence path will be now be demonstrated with a synthetic 1D signal.

\begin{figure*}[ht]
\centering
\vspace{0.1in}
\begin{overpic}[viewport = 5 1 480 400, clip, height= 3.5cm, width=3.7cm]{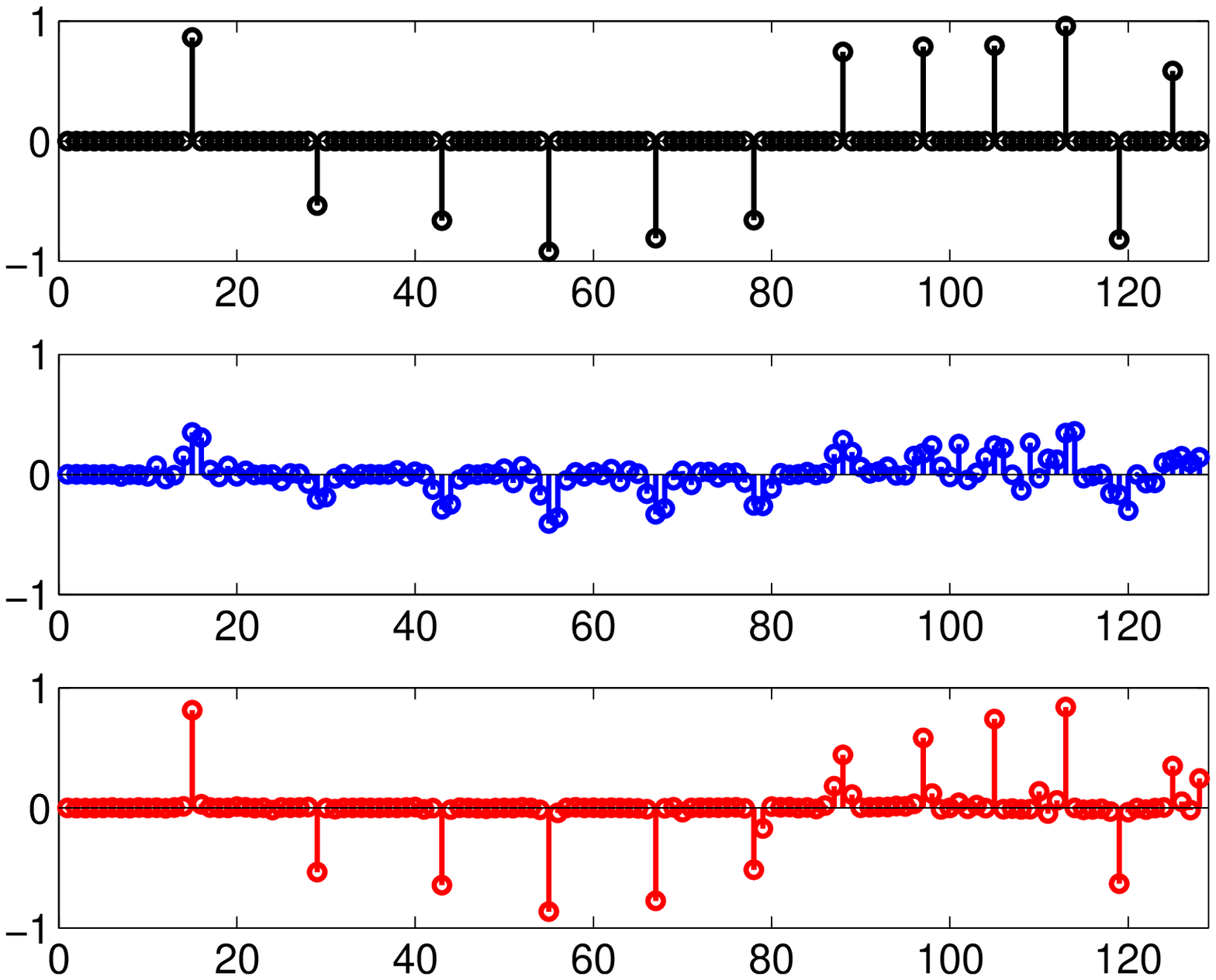}
\put(3,6){ \sffamily \footnotesize{\textcolor{black}{{(a)}}}}
\put(35,90){ \sffamily \footnotesize{\textcolor{black}{{signal}}}}
\put(63,95){ \sffamily \footnotesize{\textcolor{black}{{\textsc{Uniform Kernel}}}}}
\end{overpic}
\begin{overpic}[viewport = 2 1 480 400, clip, height= 3.5cm, width=3.7cm]{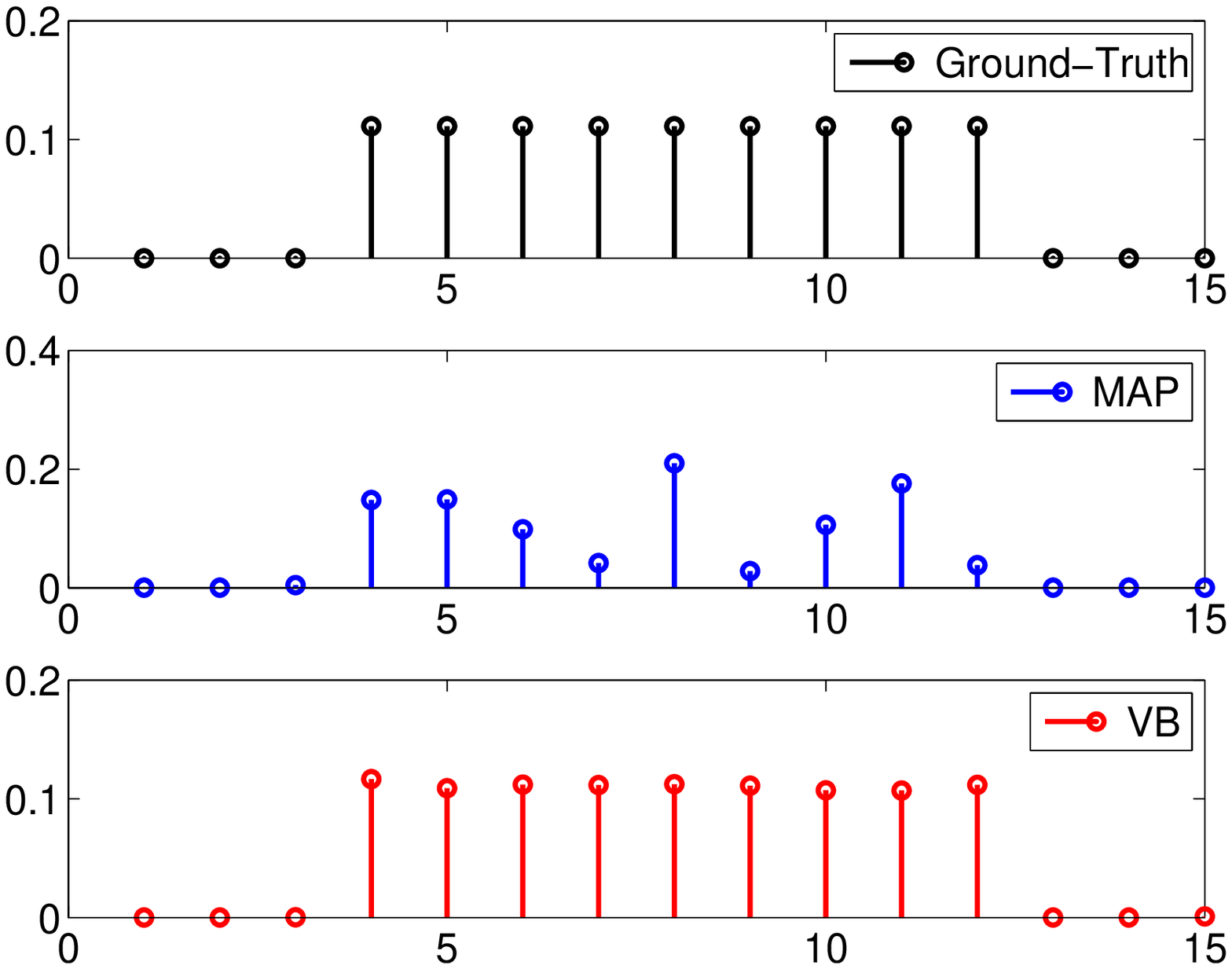}
\put(3,6){ \sffamily \footnotesize{\textcolor{black}{{(b)}}}}
\put(35,90){ \sffamily \footnotesize{\textcolor{black}{{blur kernel}}}}
\end{overpic}
\begin{overpic}[viewport = 5 1 480 400, clip, height= 3.5cm, width=3.7cm]{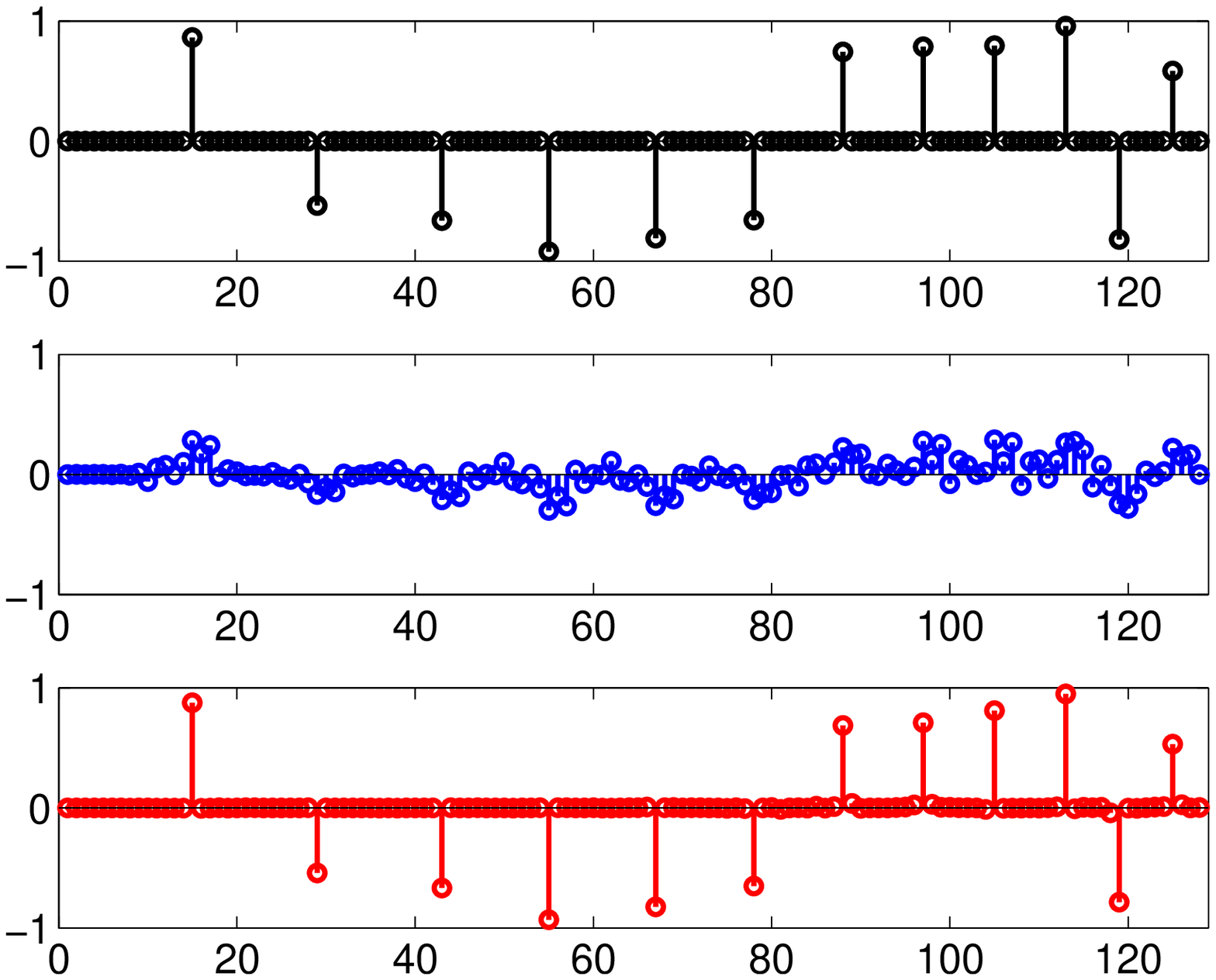}
\put(35,90){ \sffamily \footnotesize{\textcolor{black}{{signal}}}}
\put(63,95){ \sffamily \footnotesize{\textcolor{black}{{\textsc{Random Kernel}}}}}
\put(3,6){ \sffamily \footnotesize{\textcolor{black}{{(c)}}}}
\end{overpic}
\begin{overpic}[viewport = 2 1 480 400, clip, height= 3.5cm, width=3.7cm]{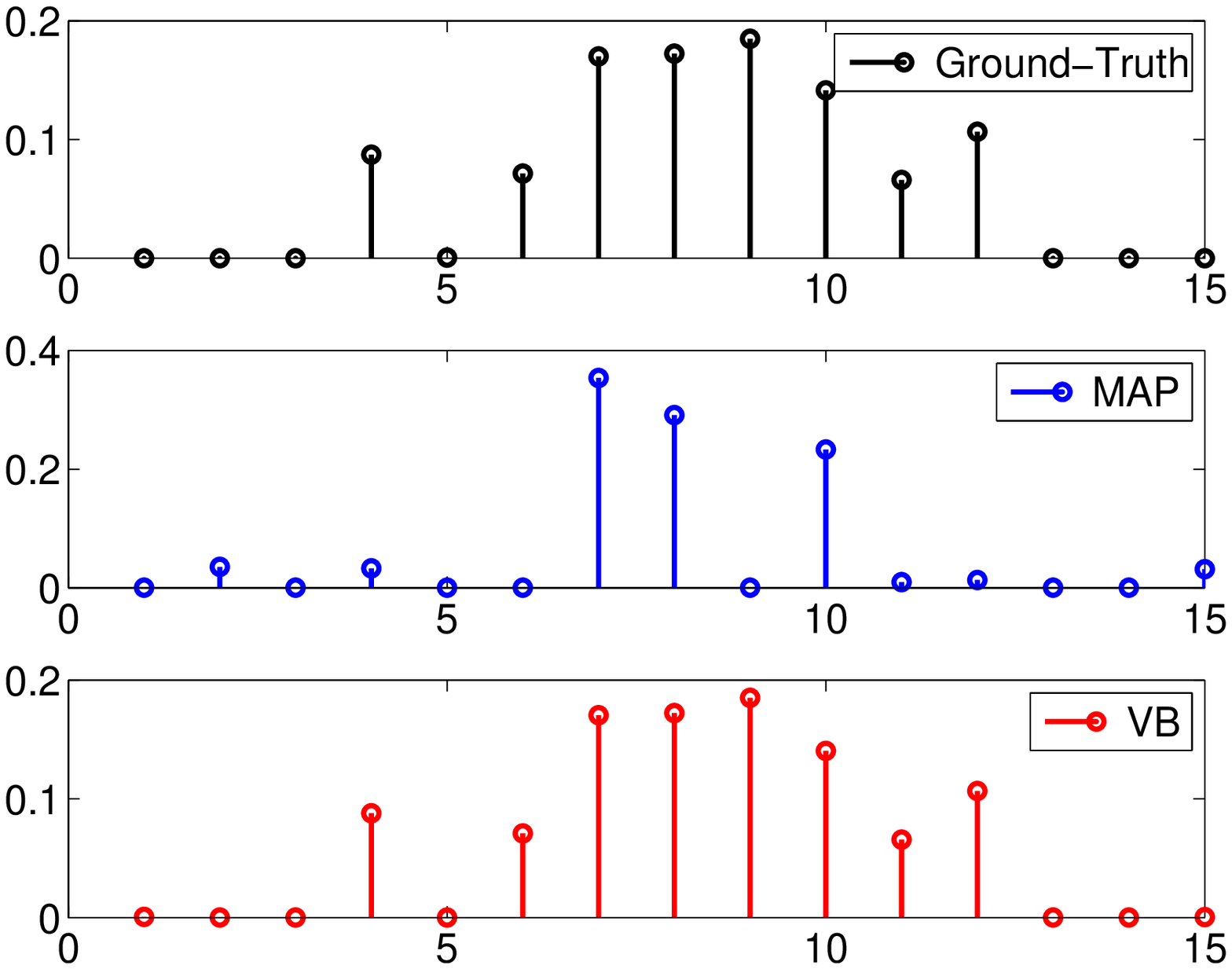}
\put(35,90){ \sffamily \footnotesize{\textcolor{black}{{blur kernel}}}}
\put(3,6){ \sffamily \footnotesize{\textcolor{black}{{(d)}}}}
\end{overpic}
\caption{1D deblurring example  using MAP and VB approaches assuming the same underlying image prior $p(\x)$. (a)-(b) results with a uniform blur kernel; (c)-(d) results with a random blur kernel.
}
\label{fig:1D_eg}
\end{figure*}

In this example, we generate a 1D signal composed of multiple spikes and convolve it  with two different blur kernels, one uniform and one random, creating two different blurry observations.  Refer to Figure~\ref{fig:1D_eg} (first row) for the ground-truth spike signal and associated blur kernels.  We then apply the MAP and VB blind deconvolution algorithms, with the same prior ($f$ equals a constant) and $\lambda$ reduction schedule, to the blurry test signals and compare the quality of the reconstructed blur kernels and signals.  The recovery results are shown in Figure~\ref{fig:1D_eg} (second and third rows), where it is readily apparent that VB produces superior estimation quality of both kernel and image.  Additionally, the signal recovered by VB is considerably more sparse than MAP, indicating that it has done a better job of optimizing (\ref{eq:vb_cost_MAP3}), consistent with subsequent theoretical analysis that will be conducted in Sections \ref{sec:implicit_penalty}-\ref{sec:blur_depend_analysis}.  This is not to say that MAP cannot potentially be effective with careful tuning and initialization (perhaps coupled with additional regularization factors or clever optimization schemes), only that VB is much more robust to suboptimal experimental settings, etc., in its present form.

Note that this demonstrable advantage of VB is entirely based on an improved convergence path, since VB and MAP possess an identical constellation of local minima once $\lambda = 0$.  Moreover, it is unrelated to any putative advantage of solving (\ref{eq:MAP_k}) over (\ref{eq:reg_bd}).  We will revisit this latter point in Section \ref{sec:relationship_levin}.

\subsection{Evaluating the VB Image Penalty $\gvb$} \label{sec:implicit_penalty}

Illustrative examples aside, we will now explore in more depth exactly how the image penalty $\gvb$ in (\ref{eq:vb_cost_reform}) contributes to the success of VB.  While in a few special cases $\gvb$ can be computed in closed-form for general $\rho \neq 0$ leading to greater transparency,  as we shall see below the VB algorithm and certain attendant analyses can nevertheless be carried through even when closed-form solutions for $\gvb$ are not possible.  Importantly, we can assess properties that may potentially affect the sparsity and quality of resulting solutions as $\lambda$ and $\|\bar{\k}\|^2_2$ are varied.

A highly sparse prior, and therefore penalty function, is generally more effective in differentiating sharp images with fine structures from blurry ones (more on this later).  Recall that concavity with respect to coefficient magnitudes is a signature property of such sparse penalties \citep{RaoECPK03,Wipf_Latent_Variable_TIT11}.  A potential advantage of MAP is that it is very straightforward to characterize the associated image penalty; namely, if $\gx$ from (\ref{eq:reg_bd}) is a highly concave, nondecreasing function of each $|x_i|$, then we may expect that sparse image gradients will be heavily favored.  And for two candidate image penalties $\gx^{(1)}$ and $\gx^{(2)}$, if $\gx^{(1)} \prec \gx^{(2)}$, then we may expect the former to promote an even sparser solution than the latter (provided we are not trapped at a bad local solution).  Section \ref{sec:relationship_levin} will argue that $\gx^{(1)}$ will then lead to a better estimate of $\x$ and $\k$.

In contrast, with VB it is completely unclear to what degree $\gvb$ favors sparse solutions, except in the special case from the previous section.  We now explicitly describe sufficient and necessary conditions for $\gvb$ to be a concave, nondecreasing function of $|x_i|$, which turn out to be much stricter than the conditions required for MAP.

\begin{Thm} \label{thm:concave}
 The VB penalty $\gvb$ will be a concave, non-decreasing function of $|x_i|$ for any $\rho$ if and only if $f$ from (\ref{eq:convex_prior}) is a concave, non-decreasing function on $[0,\infty)$.  Moreover, at least $m-n$ elements of $\x$ will equal zero at any locally minimizing solution to (\ref{eq:vb_cost_reform}) (however typically many more will equal zero in practice).
%(i.e., it favors solutions with some coefficients equal to zero, or exactly sparse).  Elaborate on theoretical consequences of this, e.g., at any local minima some elements of $\x$ will be provably equal to zero.
\end{Thm}

Theorem \ref{thm:concave} explicitly quantifies what class of image priors leads to a strong, sparsity-promoting $\x$ penalty when fully propagated through the VB framework.  Yet while this attribute may anchor VB as a legitimate sparse estimator in the image (filter) domain given an appropriate $f$, it does not explain precisely why VB often produces superior results to MAP.  In fact, the associated MAP penalty $\gx$ (when generated from the same $f$) will actually promote sparse solutions under much weaker conditions as follows:

\begin{Cor} \label{cor:concave}
 The MAP penalty $\gx$ will be a concave, non-decreasing function of $|x_i|$ if and only if $\vartheta(z) \triangleq \log(z) + f(z)$ is a concave, non-decreasing function on $[0,\infty)$.
\end{Cor}

The extra $\log$ factor implies that $f$ itself need not be concave to ensure that $\gx$ is concave.  For example, the selection $f(z) = z - \log (z)$ it not concave and yet the associated $\gx$ still will be since now $\vartheta(z) = z$, which is concave and non-decreasing as required by Corollary \ref{cor:concave}.

Moving forward then, to really understand VB we must look deeper and examine the role of $\rho$ in modulating the effective penalty on~$\x$.  First we define the function $\grho{\rho_{\alpha}}:\mathbb{R}^+ \rightarrow \mathbb{R}$ as $\grho{\rho_{\alpha}}(z) = \gvb(z,\rho = \rho_{\alpha})$, with $z\geq 0$.  Note that because $\gvb$ is a symmetric function with respect to the origin, we may conveniently examine its concavity properties considering only the positive half of the real line.

\begin{Thm} \label{thm:general_sparse_promoting}
Let $f$ be a differentiable, non-decreasing function.  Then we have the following:
\begin{enumerate}

\item As $z \rightarrow \infty$, for all $\rho_1$ and $\rho_2$, $\grho{\rho_2}(z) - \grho{\rho_1}(z) \rightarrow 0$.  Therefore, $\grho{\rho_1}$ and $\grho{\rho_2}$ penalize large magnitudes of $\x$ equally.  % , and both $g'_{\rho_1}(z)$, $g'_{\rho_2}(z) \rightarrow C$ for some $C > 0$

\item For any $z$,$z' \geq 0$ and $\rho_2 > \rho_1$, if $z < z'$ then $\grho{\rho_2}(z) - \grho{\rho_1}(z) > \grho{\rho_2}(z') - \grho{\rho_1}(z')$.  Therefore, as $z \rightarrow 0$, $\grho{\rho_2}(z) - \grho{\rho_1}(z)$ is maximized, implying that $\grho{\rho_1}$ favors zero-valued coefficients more heavily than $\grho{\rho_2}$.

%\item For all $z \geq 0$,  if $\rho_2 > \rho_1$, then $g_{\rho_2}(z) - g_{\rho_1}(z) \geq 0$ and $g'_{\rho_1}(z) - g'_{\rho_2}(z) \geq 0$.

\end{enumerate}
\end{Thm}

This result implies that regardless of $\rho$, $\gvb$ penalizes large magnitudes of any $x_i$ nearly equivalently.  In contrast, small magnitudes are penalized much less as $\rho$ becomes smaller. So Theorem~\ref{thm:general_sparse_promoting} loosely suggests that sparse solutions are more heavily favored when $\rho$ is smaller.  However, we would ideally like to  make more rigorous statements about the relative concavity of the various penalty functions involved, allowing us to make stronger claims about sparsity-promotion.

Perhaps the simplest choice for $f$ which satisfies the conditions of Theorems \ref{thm:concave} and \ref{thm:general_sparse_promoting}, and a choice that has been advocated in the sparse estimation literature in different contexts, is to assume a constant value, $f(\gamma) = b$.  This in turn implies that $p(x_i)$ is a Jeffreys non-informative prior on the coefficient magnitudes $|x_i|$ after solving the maximization from (\ref{eq:convex_prior}), and is attractive in part because there are no embedded hyperparameters (the constant $b$ is irrelevant).\footnote{The Jeffreys prior is of the form $p(x) \propto 1/|x|$, which represents an improper distribution that does not integrate to one.}  This selection for $f$ leads to a particularly interesting closed-form penalty $\gvb$ as follows:

% \footnote{f(x) = b is actually both concave and convex, and is also both non-decreasing and non-increasing.}

\begin{Thm} \label{thm:special_case}
In the special case where $f(\gamma_i) = b$, then
\begin{equation}\label{eq:penalty_fun2}
\gvb(x_i, \rho) \equiv  \frac{2|x_i|}{|x_i|+\sqrt{x_i^2 + 4\rho}} + \log \left( 2\rho + x_i^2 + |x_i|\sqrt{x_i^2 + 4\rho} \right).
\end{equation}
\end{Thm}

% **********************

\begin{figure}
\centering
\begin{overpic}[width=7cm]{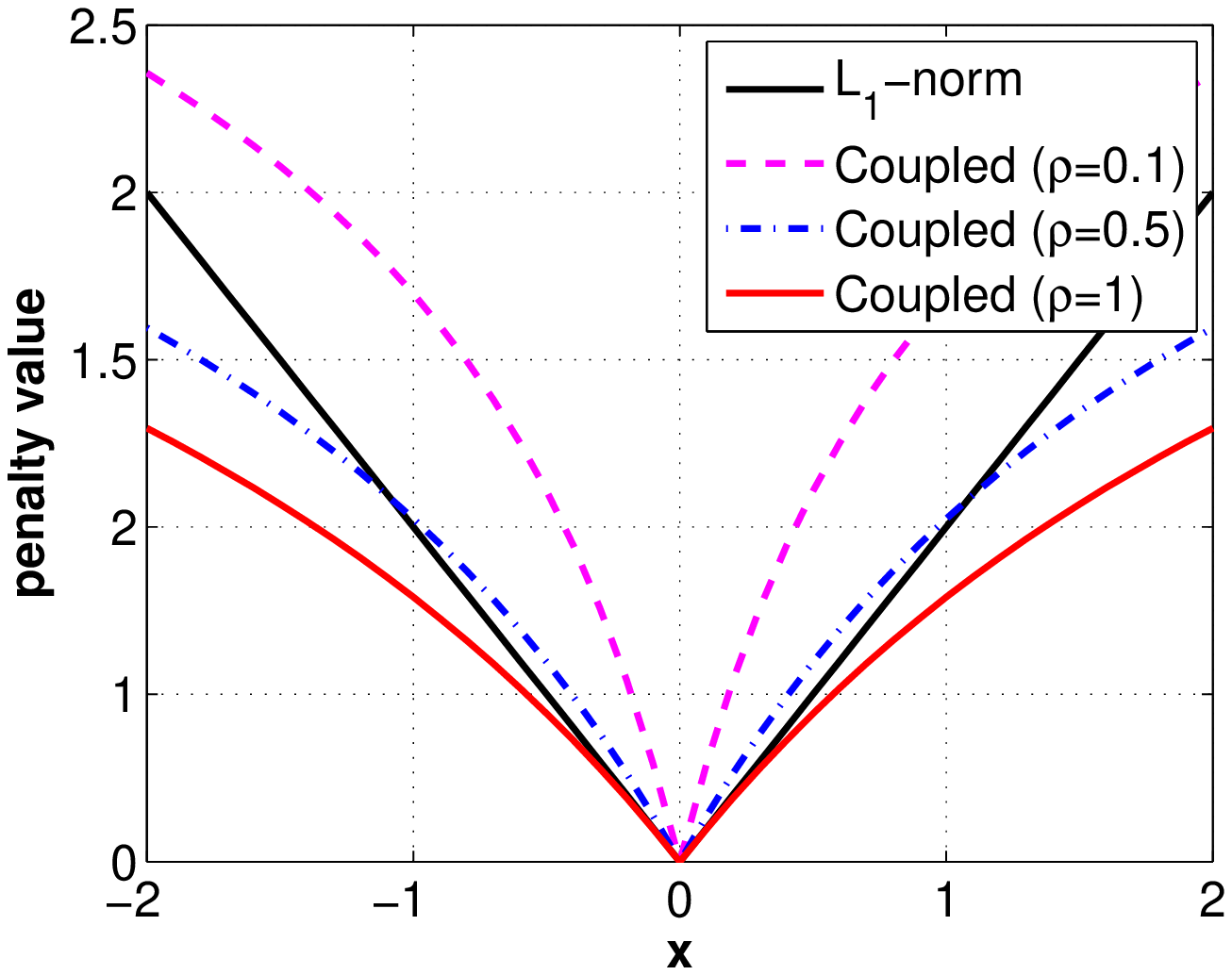}
\put(47,-4){ \sffamily \footnotesize{\textcolor{black}{{(a)}}}}
\end{overpic}
\begin{overpic}[width=7cm]{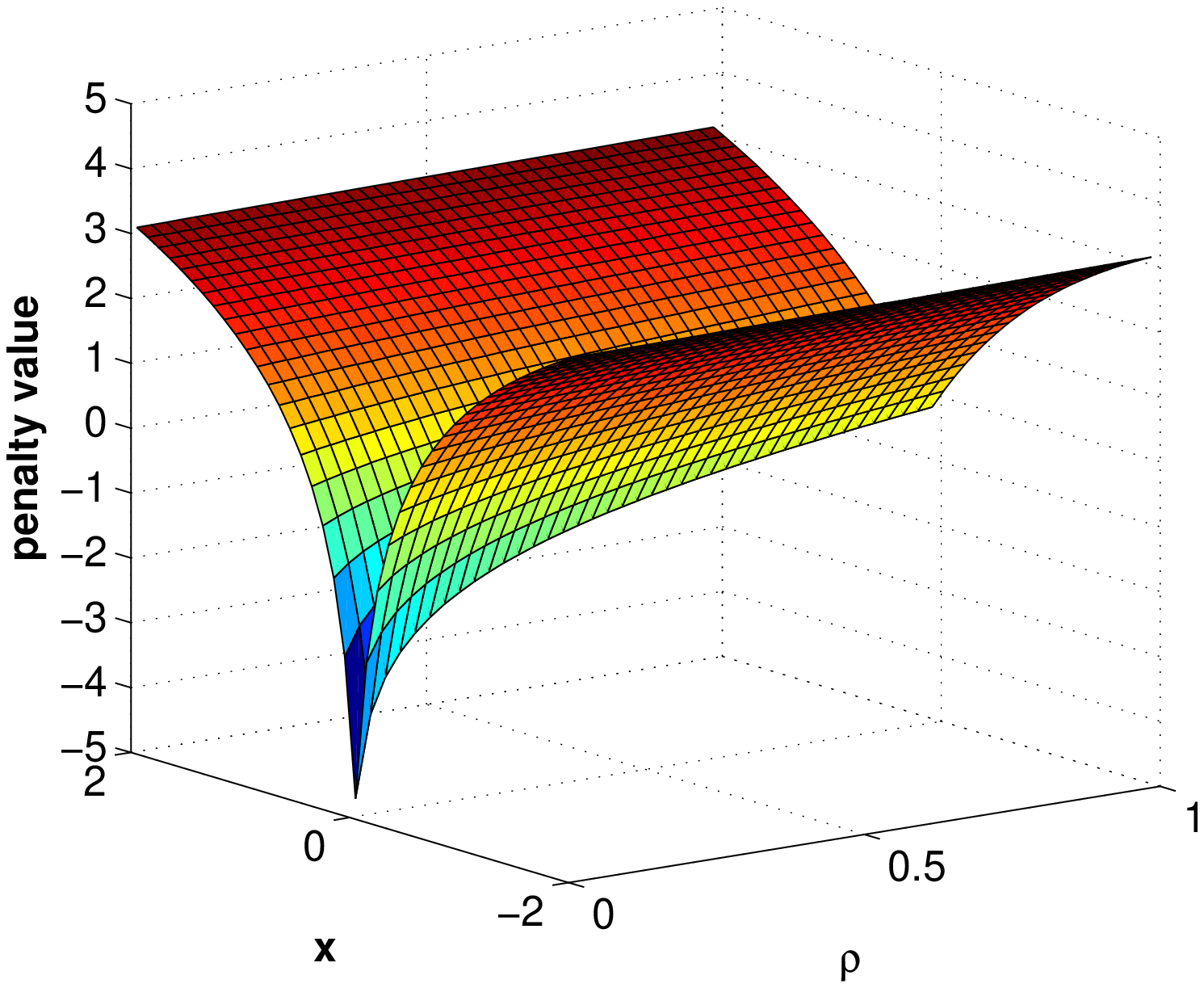}
\put(48,-4){ \sffamily \footnotesize{\textcolor{black}{{(b)}}}}
\end{overpic}
\caption{(a) A 1D example of the coupled penalty $\gvb(x, \rho)$ (normalized) with different $\rho$ values assuming $f$ is a constant. The $\mathcal{\ell}_1$ norm is included for comparison. (b) A 2D example surface plot of the coupled penalty function $\gvb(x, \rho)$; $f$ is a constant.}
\label{fig:penalty_fun}
\end{figure}

Figures \ref{fig:penalty_fun} (a) and (b) display 1D and 2D plots of this penalty function.  It is worth spending some time here to examine this particular selection for $f$ (and therefore $\gvb$) in detail since it elucidates many of the mechanisms whereby VB, with all of its attendant approximations and heuristics, can be effective.

% The first term only depends on the kernel $\k$, while the second and third terms reveal that the image penalty shape is only dependent on the ratio of the noise level to the kernel norm.

In the limit as $\rho \rightarrow 0$, the first term in (\ref{eq:penalty_fun2}) converges to the indicator function $I[x_i \neq 0]$, and thus when we sum over $i$ we obtain the $\ell_0$ norm of $\x$.\footnote{Although with $\rho = 0$, this term reduces to a constant, and therefore has no impact.}  The second term in (\ref{eq:penalty_fun2}), when we again sum over $i$, converges to $\sum_i \log |x_i|$, ignoring a constant factor. Sometimes referred to as Gaussian entropy, this term can also be connected to the $\mathcal{\ell}_0$ norm via the relations $\Vert \x \Vert_0 \equiv \lim_{p\rightarrow0} \sum_i |x_i|^p$ and $\lim_{p\rightarrow 0} \frac{1}{p} \sum_{i} (|x_i|^p -1) = \sum_i \log |x_i|$~\citep{Wipf_Latent_Variable_TIT11}.  Thus the cumulative effect when $\rho$ becomes small is an image prior that closely mimics the highly non-convex (in $|x_i|$) $\ell_0$ norm.  In contrast, when $\rho$ becomes large, it can be shown that both terms in (\ref{eq:penalty_fun2}), when combined for all $i$, approach scaled versions of the convex $\ell_1$ norm.  Additionally, if we assume a fixed kernel and ignore boundary effects, this scaling turns out to be optimal in a particular Bayesian sense described in \citep{Wipf_Dualspace_Sparse_NIPS12} (this technical point will be addressed further in a future publication).

For intermediate values of $\rho$ between these two extremes, we obtain a $\gvb$ that becomes \emph{less} concave with respect to each $|x_i|$ as $\rho$ increases (in the formal sense of relative concavity discussed in Section~\ref{sec:notation}).  In particular, we have the following:

\begin{Cor} \label{cor:special_case}
If $f(\gamma_i) = b$, then $\grho{\rho_1} \prec \grho{\rho_2}$ for $\rho_1 < \rho_2$.
\end{Cor}

Thus, as the noise level $\lambda$ is increased, $\rho$ increases and we have a penalty that behaves more like a convex (less sparse) function, and so becomes less prone to local minima.  In contrast, as $\|\bar{\k}\|^2_2$ is increased, meaning that $\rho$ is now reduced, the penalty actually becomes \emph{more} concave with respect to $|x_i|$.  This phenomena is in some ways similar to certain homotopy sparse estimation schemes (e.g., \citep{ChartrandY08}), where heuristic hyperparameters are introduced to gradually introduce greater non-convexity into canonical compressive sensing problems, but without any dependence on the noise or other factors.  The key difference here with VB however is that penalty shape modulation is explicitly dictated by both the noise level $\lambda$ and the kernel $\k$ in an entirely integrated fashion.

To summarize then, the ratio $\rho$ can be viewed as modulating a smooth transition of the penalty function shape from something akin to the non-convex $\ell_0$ norm to a properly-scaled $\ell_1$ norm.  In contrast, all conventional MAP-based penalties on $\x$ are independent from $\k$ or $\lambda$, and thus retain a fixed shape.  The crucial ramifications of this coupling and $\rho$-controlled shape modification/augmentation exclusive to the VB framework will be addressed in the following two subsections.  Other choices for $f$, which exhibit a partially muted form of this coupling, will be considered in Section \ref{sec:other_f}, which will also address a desirable form of invariance that only exists when $f$ is a constant.

\subsection{Noise Dependency Analysis} \label{sec:noise_depend_analysis}

The success of practical VB blind deconvolution algorithms is heavily dependent on some form of stagewise coarse-to-fine approach, whereby the kernel is repeatedly re-estimated at successively higher resolutions.  At each stage, a lower resolution version is used to initialize the estimate at the next higher resolution.  One way to implement this approach is to initially use large values of $\lambda$ such that only dominant, primarily low-frequency image structures dictate the optimization \citep{Understanding_BD}.  During subsequent iterations as the blur kernel begins to reflect the correct coarse shape, $\lambda$ can be gradually reduced to allow the recovery of more detailed, fine structures.

A highly sparse (concave) prior can ultimately be more effective in differentiating sharp images and fine structures than a convex one. Detailed supported evidence for this claim can be found in \citep{Fergus06removingcamera, sps_deblur, KrishnanF09_NIPS, Cho_PAMI}, as well as in Section~\ref{sec:relationship_levin} below.  However, if such a prior is applied at the initial stages of estimation, the iterations are likely to become trapped at suboptimal local minima, of which there will always be a combinatorial number.  Moreover, in the early stages, the effective noise level is actually high due to errors contained in the estimated blur kernel, and exceedingly sparse image penalties are likely to produce unstable solutions.  Given the reformulation outlined above, we can now argue that VB implicitly avoids these problems by beginning with a large $\lambda$ (and therefore a large $\rho$), such that the penalty function is initially nearly convex in $|x_i|$ (see Figure~\ref{fig:penalty_fun}).  As the iterations proceed and fine structures need to be resolved, the penalty function becomes less convex as $\lambda$ is reduced, but the risk of local minima and instability is ameliorated by the fact that we are likely to be already in the neighborhood of a desirable basin of attraction.  Additionally, the implicit noise level (or modeling error) is now substantially less.

This kind of automatic `resolution' adaptive penalty shaping is arguably superior to conventional MAP approaches based on (\ref{eq:reg_bd}), where the concavity/shape of the induced separable penalty function is kept fixed regardless of the variation in the noise level or scale, {i.e.}, at different resolutions across the coarse-to-fine hierarchy.  In general, it would seem very unreasonable that the same penalty shape would be optimal across vastly different noise scales.    This advantage over MAP can be easily illustrated by simple head-to-head comparisons where the underlying prior distributions are identical (such as the previous example from Section~\ref{sec:1D_exp}).  Additionally, this phenomena can be further enhanced by automatically learning $\lambda$ as discussed in Section \ref{sec:learn_lambda}.

%(In fact, this phenomena is not at all unlike the homotopy sparse estimation scheme from \citep{ChartrandY08}, where a heuristic hyperparameter is introduced to gradually introduce greater non-convexity into canonical compressive sensing problems.)

\subsection{Blur Dependency Analysis} \label{sec:blur_depend_analysis}
The shape parameter $\rho$ is also affected by the kernel norm, with larger values of $\Vert \bar{\k}\Vert_2^2$ leading to less convexity of the penalty function $\gvb$ while small values increase the convexity as can also be observed from Figure~\ref{fig:penalty_fun}. With the standard assumptions $\sum_j k_j=1$ and $k_j \geq 0$, $\Vert \bar{\k} \Vert_2^2$ is bounded between $1/l$ and $1$, where $l$ is the number of pixels in the kernel.\footnote{Actually, because of natural invariances embedded into the VB cost function, the assumption $\sum_j k_j=1$ is not needed for what follows.  See Section \ref{sec:other_f} for more details.}  An increase of $\Vert \bar{\k} \Vert_2^2$  indicates that the kernel is more sparse, with the extreme case of $\k=\delta$ leading to $\Vert \bar{\k} \Vert_2^2=1$.  In this situation, $\gvb$ is the most concave in $|x_i|$ (per the analysis of Section \ref{sec:implicit_penalty}), which is reasonable, as this is the easiest kernel type to handle so the sparsest penalty function can be used without much concern over local minima. In contrast, $\Vert \bar{\k} \Vert_2^2$ is the smallest when all elements are equal, which is the more challenging case corresponding with a broad diffuse image blur, with many local minima. In this situation, the penalty function is more convex and conservative.  In general, a highly concave prior is not needed to disambiguate a highly blurred image from a relatively sharp one.

Additionally, at the beginning of the learning process when $\lambda$ is large and before any detailed structures have been resolved, the $\log \Vert \bar{\k} \Vert_2^2$ penalty on $\k$ from (\ref{eq:vb_cost_reform}) will naturally favor a blurry, diffuse kernel in the absence of additional information.  This will help ensure that $\gvb$ is relatively convex and less aggressive during the initial VB iterations.  However, as the algorithm proceeds, $\lambda$ is reduced, and some elements of $\x$ are pushed towards zero, the penalty $\gvb$, with its embedded $\k$ dependency, will gradually become less convex and can increasingly dominate the overall cost function (since for small $\lambda$ and large $\Vert \bar{\k} \Vert_2^2$ the lower bound on $\gvb$ can drop arbitrarily per the above described concavity modulation).  Because $\gvb$ is minimized as $\k$ becomes relatively sparse, a more refined $\k$ can be explored at this stage to the extent that $\x$ can be pushed towards greater sparsity as well (if $\x$ is not sparse, then there is no real benefit to refining $\k$).  Again, this desirable effect occurs with relatively limited risk of local minima because of the gradual, intrinsically-calibrated introduction of increased concavity.

We may also consider these ideas in the context of existing MAP algorithms, which adopt various structure selection heuristics, implicitly or explicitly, to achieve satisfactory performance~\citep{hqdeblurring_siggraph2008, fast_motion_deblur_2009, XuJ10_ECCV}.  This can be viewed as adding additional image penalty terms and trade-off parameters to (\ref{eq:reg_bd}). For example, \citep{hqdeblurring_siggraph2008} incorporates an extra local penalty on the latent image, such that the gradients of small-scale structures in the recovered image are close to those in the blurry image.  Thus they will actually contribute less to the subsequent kernel estimation step, allowing larger structures to be captured first.  Similarly, a bilateral filtering step is used for pruning out small scale structures in~\citep{fast_motion_deblur_2009}. Finally, \citep{XuJ10_ECCV} develop an empirical structure selection  metric designed such that small scale structures can be pruned away by thresholding the corresponding response map, allowing  subsequent kernel estimation to be dominated by only large-scale structures.

Generally speaking, existing MAP strategies face a trade-off:  either they must adopt a highly sparse image prior needed for properly resolving fine structures (see Section \ref{sec:relationship_levin}) and then deal with the attendant constellation of problematic local minima,\footnote{Appropriate use of continuation methods such as the algorithm from \citep{ChartrandY08} may help in this regard.} or rely on a more smooth image prior augmented with compensatory structure-selection measures such as those described above to avoid bad global solutions.  In contrast, we may interpret the coupled penalty function intrinsic to VB as a principled alternative with a transparent, integrated functionality for estimation at different resolutions without any additional penalty factors or complexity.

\subsection{Other Choices for $f$} \label{sec:other_f}

Because essentially any sparse prior on $\x$ can be expressed using the alternative variational form from (\ref{eq:convex_prior}), choosing such a prior is tantamount to choosing $f$ which then determines $\gvb$.  The results of Theorems~\ref{thm:concave} and \ref{thm:general_sparse_promoting} suggest that a concave, non-decreasing $f$ is useful for favoring sparsity (assumed to be in the gradient domain).  Moreover, Theorem \ref{thm:special_case} and subsequent analyses suggest that the simplifying choice where $f(\gamma) = b$ possesses several attractive properties regarding the relative concavity of the resulting $\gvb$.  But what about other selections for $f$ and therefore $\gvb$?  % It is possible to address this question

While directly working with $\gvb$ can sometimes be limiting (except in certain special cases like $f(\gamma) = b$ from before),
the variational form of (\ref{eq:general_f_penalty1}) allows us to closely examine the relative concavity of a useful proxy.  Let
%$\psi(\gamma_i, \rho) \triangleq \log(\rho + \gamma_i ) +  f(\gamma_i)$.
\begin{equation} \label{eq:general_f_penalty2}
\psi(\gamma_i, \rho) \triangleq \log(\rho + \gamma_i ) +  f(\gamma_i).
\end{equation}
Then for fixed $\lambda$ and $\k$ the VB estimation problem can equivalently be viewed as solving
\begin{equation} \label{eq:vb_cost_reform_gamma_space}
\min_{\x,\gam\ge0} \frac{1}{\lambda} \left\|\y - \k \ast \x \right\|_2^2 + \sum_i \left[ \frac{x_i^2}{\gamma_i} + \psi(\gamma_i,\rho) \right].
\end{equation}
%it is possible to express $\gvb$ in a variational form that allows us to closely examine the relative concavity of a useful proxy.  In particular, any $\gvb$ can be expressed as
%\begin{equation} \label{eq:general_f_penalty1}
%\gvb(x_i,\rho) \triangleq \min_{\gamma_i\ge 0} \frac{x_i^2}{\gamma_i} + \psi(\gamma_i,\rho),
%\end{equation}
%where
%\begin{equation} \label{eq:general_f_penalty2}
%\psi(\gamma_i, \rho) \triangleq \log(\rho + \gamma_i ) +  f(\gamma_i).
%\end{equation}
%See the proof of Theorem 1 in the Appendix A for details.  Thus, for fixed $\lambda$ and $\k$ the VB estimation problem can equivalently be viewed as solving
%\begin{equation} \label{eq:vb_cost_reform_gamma_space}
%\min_{\x,\gam\ge0} \frac{1}{\lambda} \left\|\y - \k \ast \x \right\|_2^2 + \sum_i \left[ \frac{x_i^2}{\gamma_i} + \psi(\gamma_i,\rho) \right]
%\end{equation}
It now becomes  clear that the sparsity of $\x$ and $\gam$ are intimated related.  More concretely, assuming $f$ is concave and non-decreasing (as motivated by Theorems 2 and 3), then there is actually a one-to-one correspondence in that whenever $x_i = 0$, the optimal $\gamma_i$ equals zero as well, and vice versa.\footnote{To see this first consider $x_i = 0$.  The $x_i^2/\gamma_i$ term can be ignored and so the optimal $\gamma_i$ need only minimize $\log(\rho + \gamma_i ) +  f(\gamma_i)$, which is concave and non-decreasing whenever $f$ is.  Therefore the optimal $\gamma_i$ is trivially zero.  Conversely if $\gamma_i = 0$, then there is effectively an infinite penalty on $x_i$, and so the optimal $x_i$ must also be zero.}  Therefore we may instead examine the relative concavity of $\psi$ for different $\rho$ values, which will directly determine the sparsity of $\gam$ and in turn, the sparsity of $\x$.  This then motivates the following result:

\begin{Thm} \label{thm:relative_concavity}
Let $\rho_1 < \rho_2$ and assume that $f$ satisfies the conditions of Theorems \ref{thm:concave} and \ref{thm:general_sparse_promoting}.   Then $\psirho{\rho_1} \prec \psirho{\rho_2}$ if and only if $f(\gamma) = a \gamma + b$, with $a\geq0$.
\end{Thm}

Thus, although we have not been able to formally establish a relative concavity result for all general $\gvb$ directly, Theorem~\ref{thm:relative_concavity} provides a nearly identical analog allowing us to draw similar conclusions to those detailed in Sections \ref{sec:noise_depend_analysis} and \ref{sec:blur_depend_analysis} whenever a general affine $f$ is adopted.  Perhaps more importantly, it also suggests that as $f$ deviates from an affine function, we may begin to lose some of the desirable effects regarding the described penalty shape modulation.

While previously we closely scrutinized the special affine case where $f(\gamma) = b$, it still remains to examine the more general affine form $f(\gamma) = a\gamma + b$, $a > 0$.  In fact, it is not difficult to show that as $a$ is increased, the resulting penalty on $\x$ increasingly resembles an $\ell_1$ norm with lesser dependency on $\rho$, thus severely muting the effect of the shape modulation that appears to be so effective (see arguments above and empirical results section below).  So there currently does not seem to be any advantage to choosing some $a > 0$ and we are left, out of the multitude of potential image priors, with the conveniently simple choice of $f(\gamma) = b$, where the value of $b$ is inconsequential.  Experimental results support this conclusion: namely, as $a$ is increased from zero performance gradually degrades (results not shown for space considerations).

As a final justification for simply choosing $f(\gamma) = b$, there is a desirable form of invariance that uniquely accompanies this selection.
\begin{Thm} \label{thm:invariance}
If $\x^*$ and $\k^*$ represent the optimal solution to (\ref{eq:vb_cost_reform}) under the constraint $\sum_i k_i = 1$, then $\alpha^{-1} \x^*$ and $\alpha \k^*$ will always represent the optimal solution under the modified constraint $\sum_i k_i = \alpha$ if and only if $f(\gamma) = b$.
\end{Thm}
This is unlike the myriad of MAP estimation techniques or VB with other choices of $f$, where the exact calibration of the constraint can fundamentally alter the form of the optimal solution beyond a mere rescaling.  Moreover, if such a constraint on $\k$ is omitted altogether, these other methods must then carefully tune associated trade-off parameters, so in one way or another this lack of invariance will require additional tuning.

Interestingly, Babacan \emph{et al.}~\citep{BabacanMDK12} experiment with a variety of VB algorithms using different underlying image priors, and empirically find that $f$ as a constant works best; however, no rigorous explanation is given for why this should be the case.\footnote{Based on a strong simplifying assumption that the covariance $\C$ from Algorithm \ref{algo:algo1} is a constant, Babacan \emph{et al.}~\citep{BabacanMDK12} provide some preliminary discussion regarding possibly why VB may be advantageous over MAP.  However, this material mostly exists in the sparse estimation literature (e.g., see \citep{Wipf_VEM_NIPS05,Wipf_Latent_Variable_TIT11} and related references) and therefore the behavior of VB blind deconvolution remains an open question, including why a constant $f$ might be optimal.}  Thus, our results provide a powerful theoretical confirmation of this selection, along with a number of useful attendant intuitions.

\subsection{Analysis Summary}

To summarize this section, we have shown that the shape of the effective VB image penalty is explicitly controlled by the ratio of the noise variance to the squared kernel norm, and that in many circumstances this leads to a desired mechanism for controlling relative concavity and balancing sparsity, largely mitigating issues such as local minima that compromise the convergence of more traditional MAP estimators.  We have then demonstrated a unique choice for the image prior (i.e., when $f$ is constant) such that this mechanism is in some sense optimal and scale-invariant.  Of course we readily concede that different choices for the image prior could still be useful when other factors are taken in to account.  We also emphasize that none of this is meant to suggest that real imaging data follows a Jeffreys prior distribution (which is produced when $f$ is constant).  We will return to this topic in  Section \ref{sec:relationship_levin} below.  Overall, this perspective provides a much clearer picture of how VB is able to operate effectively and how we might expect to optimize performance.

While space precludes a detailed treatment, many natural extensions to VB are suggested by these developments.  For example, in the original formation of VB given by (\ref{eq:VB_DB}) it is not clear the best way to incorporate alternative noise models because the required integrations are no longer tractable.  However, when viewed alternatively using (\ref{eq:vb_cost_reform}) it becomes obvious that different data-fidelity terms can easily be substituted in place of the quadratic likelihood factor.  Likewise, given additional prior knowledge about the blur kernel, there is no difficulty in substituting for the $\ell_2$-norm on $\k$ or the uniform convolutional observation model to reflect additional domain knowledge.  Thus, the proposed reformulation allows VB to inherit most of the transparent extensibility previously reserved for MAP.

\section{The Trouble with Natural Image Statistics} \label{sec:relationship_levin}

Levin \emph{et al.}~\citep{Understanding_BD,LevinWDF11,LevinWDF11_PAMI}, which represents the primary inspiration for our work, presents a compelling and highly influential case that joint MAP estimation over $\x$ and $\k$ generally favors a degenerate, no-blur solution, meaning that $\k$ will be a delta function, even when the assumed image prior $p(\x)$ reflects the true underlying distribution of $\x$, meaning $p(\x) = p_{{\rm true}}(\x)$, and $p(\k)$ is assumed flat.\footnote{Note that Levin \emph{et al.} frequently use $\mbox{MAP}_{x,k}$ to refer to joint MAP estimation over both $\k$ and $\x$ (Type I) while using $\mbox{MAP}_{k}$ for MAP estimation of $\k$ alone after $\x$ has been marginalized out (Type II).  In this terminology, $\mbox{MAP}_{k}$ then represents the inference ideal that VB purports to approximate, equivalent to (\ref{eq:MAP_k}) herein.}  In turn, this is presented as a primary argument for why MAP is inferior to VB.  As this line of reasoning is considerably different from that given in Section \ref{sec:EB_BDB}, here we will take a closer look at these orthogonal perspectives in the hopes of providing a clarifying resolution.

To begin, it helps to revisit the formal analysis of MAP failure from \citep{LevinWDF11_PAMI}, where the following specialized scenario is presented.  Assume that a blurry image $\y$ is generated by $\y = \k^* \ast \x^*$, where $\|\k^*\|_2 \ll 1$ and each image gradient $x^*_i$ is drawn iid from the generalized Gaussian distribution $p_{{\rm true}}(x_i^*) \sim \exp(-|x_i^*|^{p})$, $0 < p \leq 1$.  Now consider the minimization problem
\begin{equation} \label{eq:levin_analysis_noiseless}
\min_{\x,\k} \sum_i |x_i|^{p} \hspace*{0.3cm} \mbox{s.t. } \y = \k \ast \x, \hspace*{0.2cm} \sum_i k_i = 1,\ k_i \geq 0.
\end{equation}
Solving (\ref{eq:levin_analysis_noiseless}) is equivalent to MAP estimation over $\x$ and $\k$ under the true image prior $p_{{\rm true}}(\x)$ (and a flat prior on $\k$ within the specified constraint set).  In the limit as the image grows arbitrarily large, \citep[Claim 2]{LevinWDF11_PAMI} proves that the no-blur delta solution $\{\x = \y$, $\k = \delta \}$ will be favored over the true solution $\{\x = \x^*$, $\k = \k^*\}$.  Intuitively, this occurs because the blurring operator $\k$ contributes two opposing effects:
\begin{enumerate}
\item It reduces a measure of the image \emph{sparsity}, which increases $\sum_i |y_i|^{p}$, and
\item It broadly reduces the overall image \emph{variance}, which reduces $\sum_i |y_i|^{p}$.
\end{enumerate}
Depending on the relative contributions, we may have the situation where the second effect dominates such that $\sum_i |y_i|^{p}$ may be less than $\sum_i |x^*_i|^{p}$, meaning the cost function value at the blurred image is actually lower than at the true, sharp image. Consequently, MAP estimation may not be reliable. % even when armed with the true image prior.

%(provided the true blur kernel is large enough)

%%===this is new====
\begin{figure*}[ht]
\begin{center}
\begin{minipage}[b]{0.3\linewidth}
\begin{overpic}[viewport =  0 1 550 420, clip,  width=5.cm]{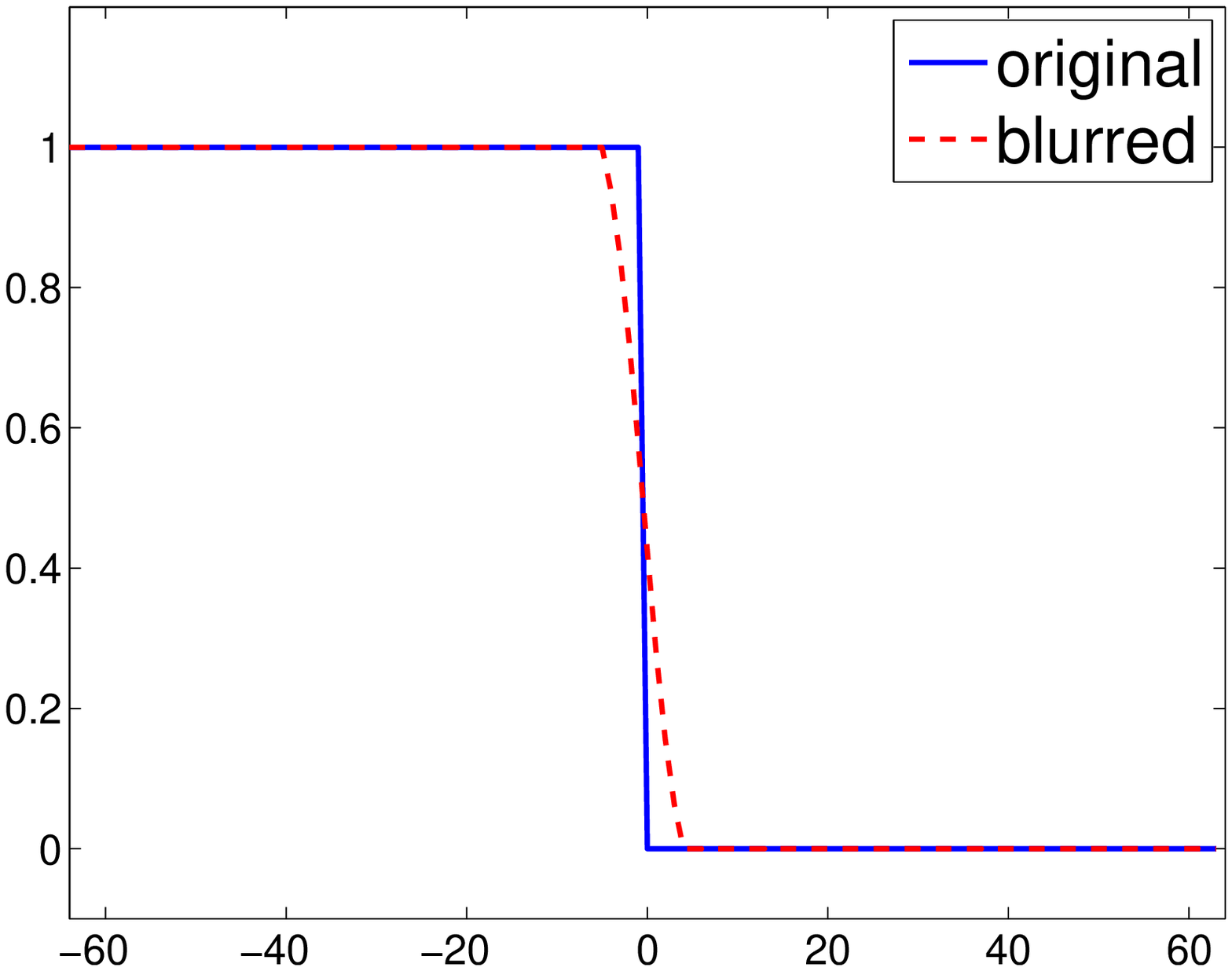}
\end{overpic}
\begin{overpic}[viewport =  -5 1 545 420, clip,  width=5.cm]{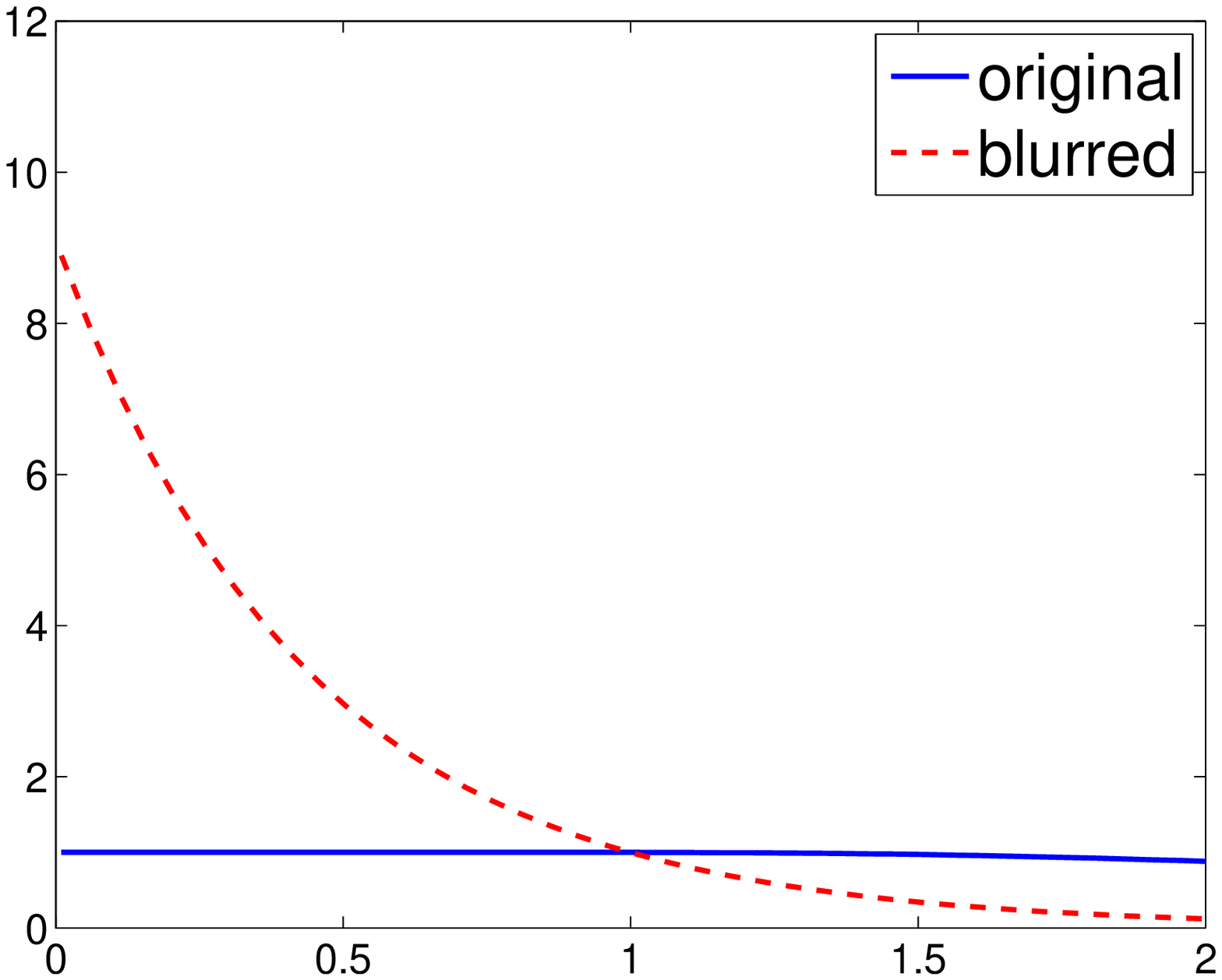}
%\put(3,1){ \sffamily {\textcolor{black}{{$\alpha$}}}}
\put(-3,25){\sffamily\footnotesize\rotatebox{90}{cost value}}
\end{overpic}
\put(-94,-6){  \footnotesize{\textcolor{black}{{$p$}}}}
\put(-85,-20){ \sffamily \footnotesize{\textcolor{black}{{(a)}}}}
\end{minipage}
\begin{minipage}[b]{0.3\linewidth}
\begin{overpic}[viewport =  0 1 550 420, clip,  width=5.cm]{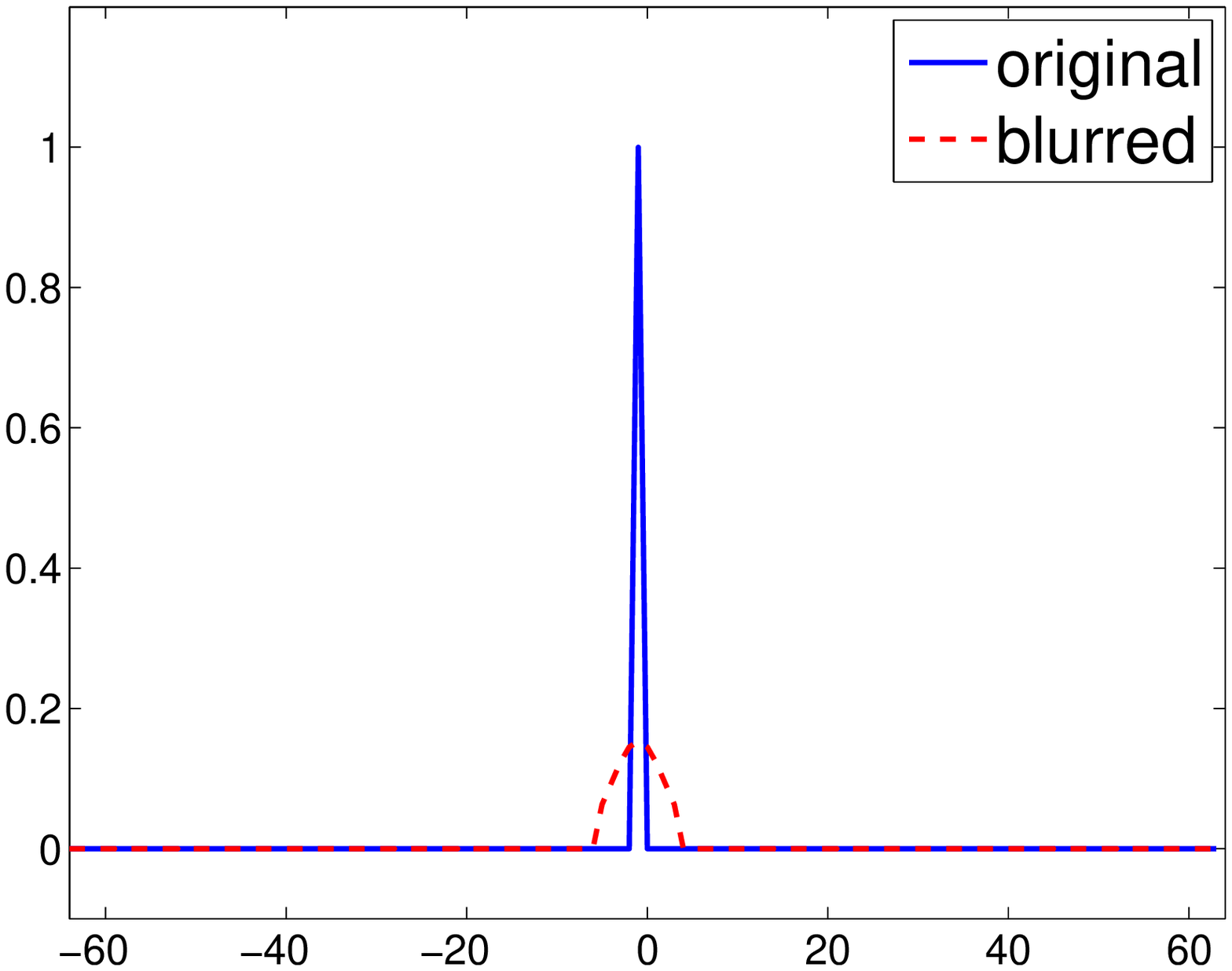}
\end{overpic}
\begin{overpic}[viewport =  -5 1 545 420, clip,  width=5.cm]{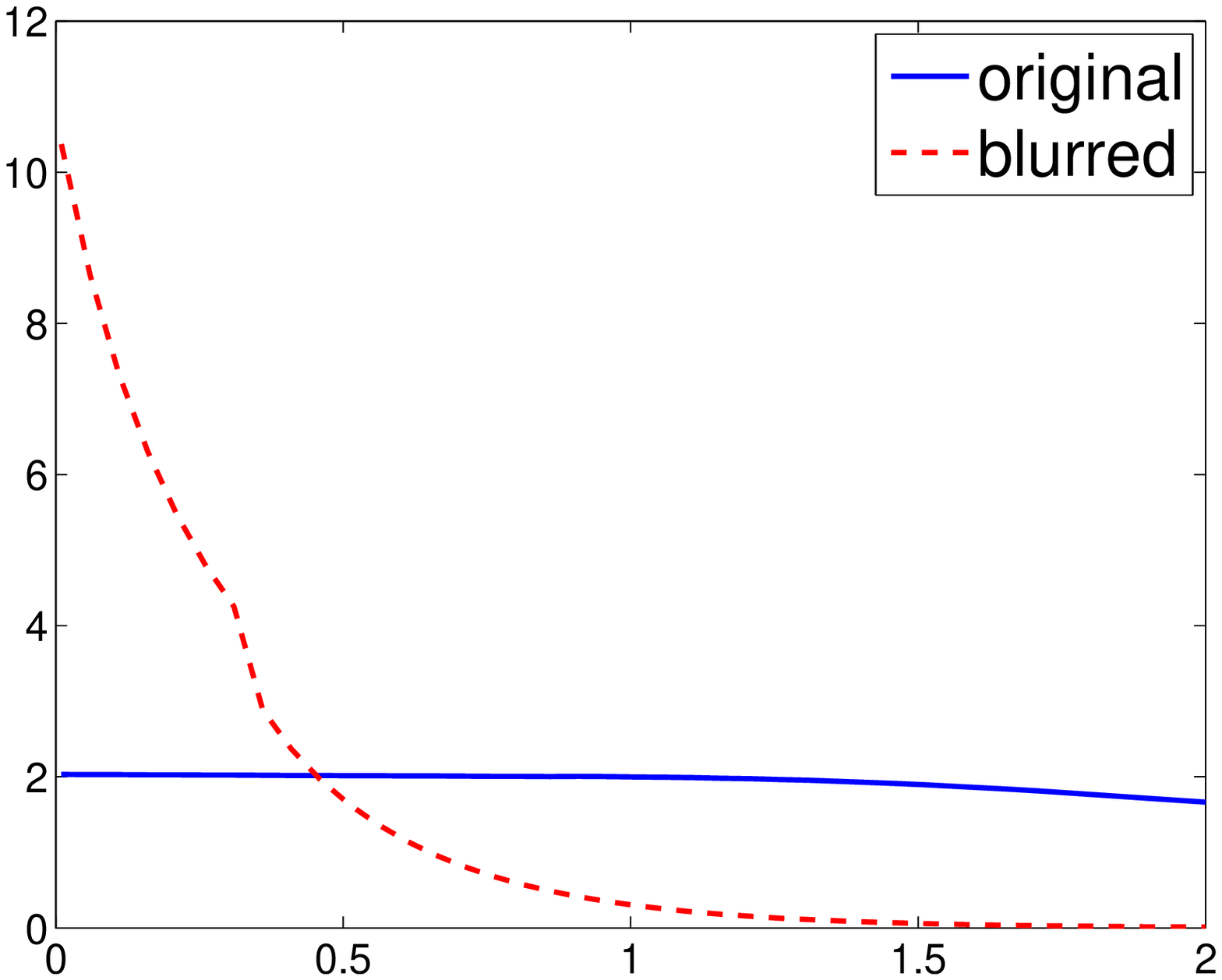}
%\put(3,1){ \sffamily \footnotesize{\textcolor{black}{{(b)}}}}
\put(-3,25){\sffamily\footnotesize\rotatebox{90}{cost value}}
\end{overpic}
\put(-94,-6){  \footnotesize{\textcolor{black}{{$p$}}}}
\put(-85,-20){ \sffamily \footnotesize{\textcolor{black}{{(b)}}}}
\end{minipage}
\begin{minipage}[b]{0.3\linewidth}
\begin{overpic}[viewport =  0 1 550 420, clip,  width=5.cm]{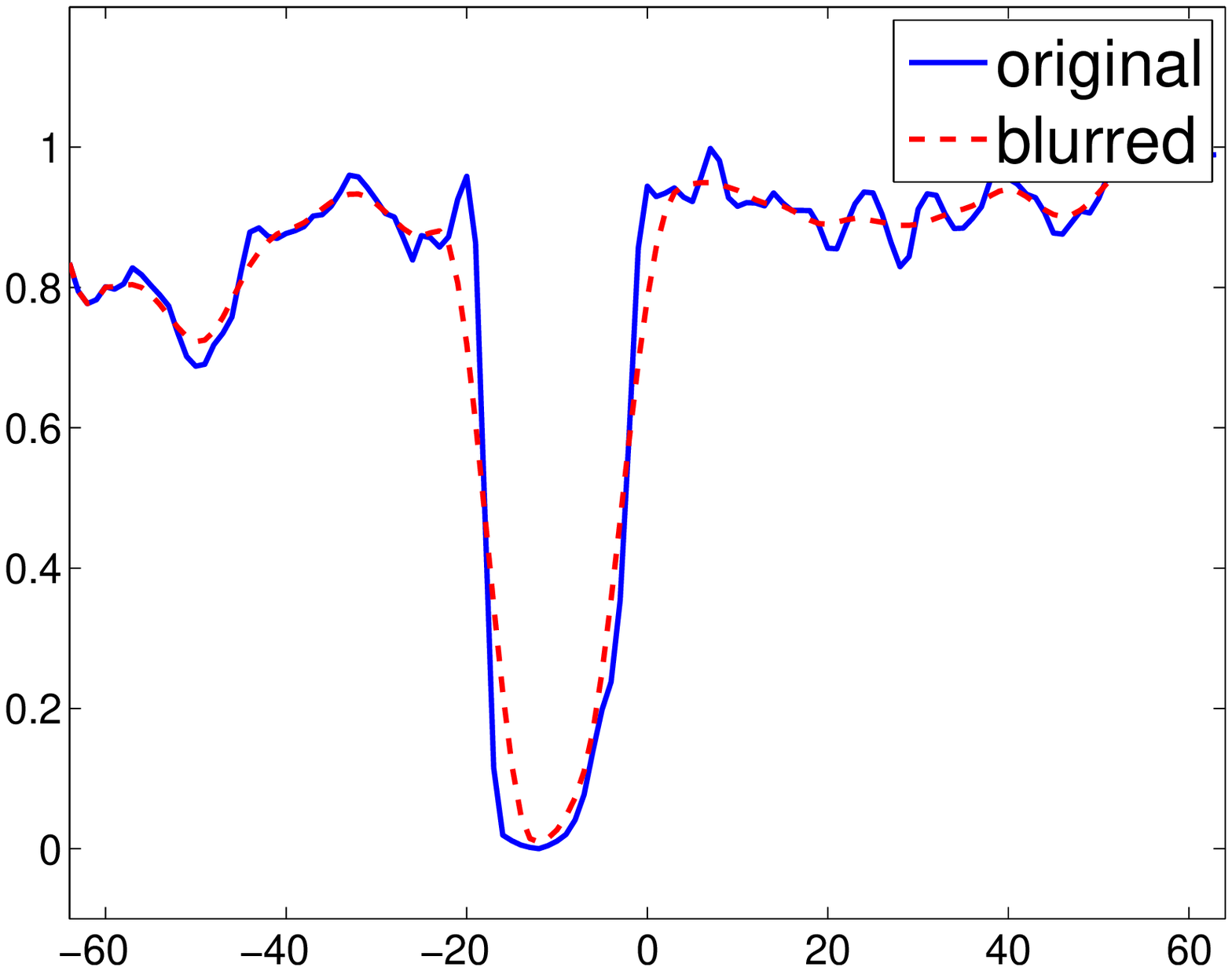}
\end{overpic}
\begin{overpic}[viewport = 0 1 550 420, clip,  width=5.cm]{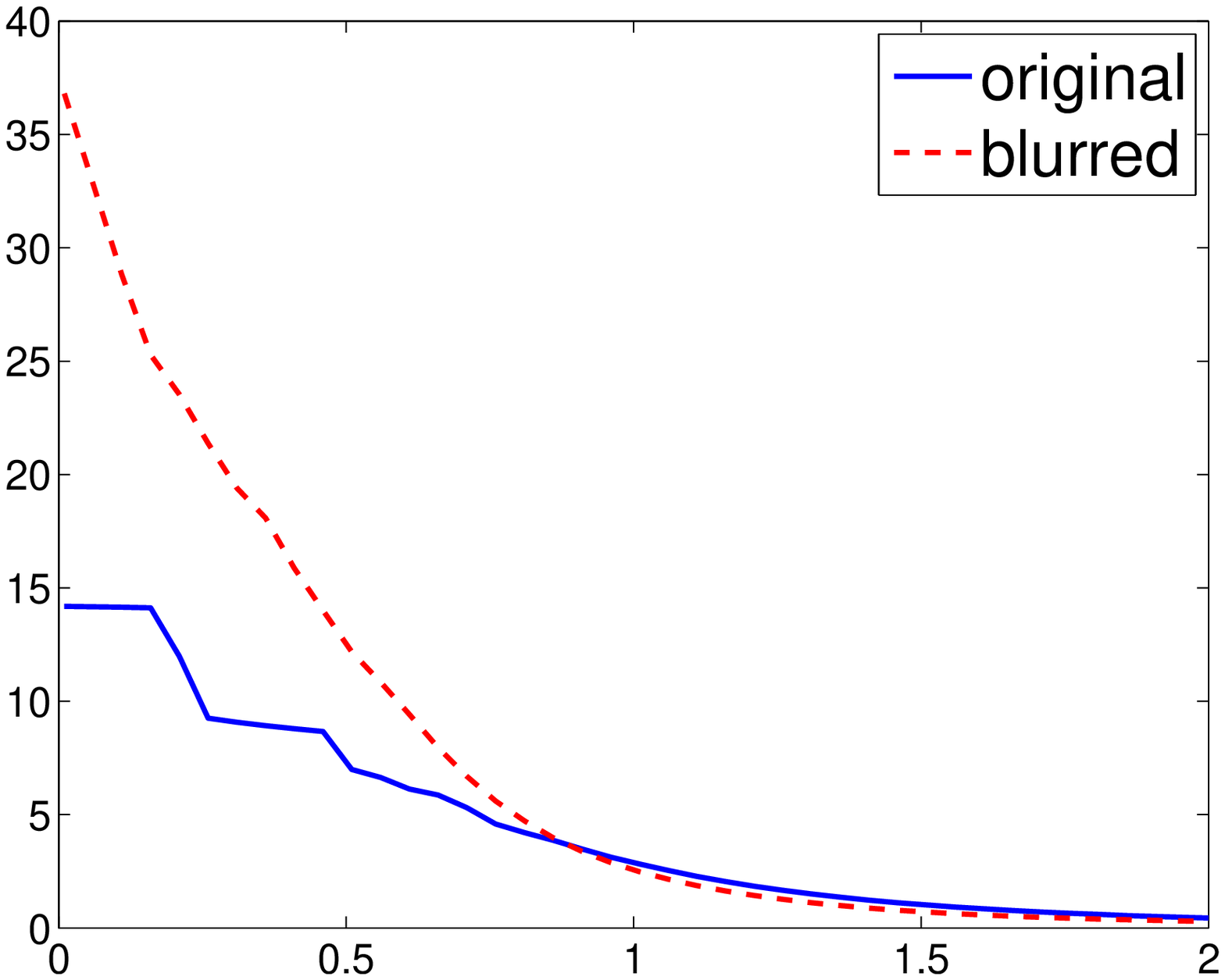}
%\put(3,1){ \sffamily \footnotesize{\textcolor{black}{{(c)}}}}
\put(-5,25){\sffamily\footnotesize\rotatebox{90}{cost value}}
\end{overpic}
\put(-94,-6){  \footnotesize{\textcolor{black}{{$p$}}}}
\put(-85,-20){ \sffamily \footnotesize{\textcolor{black}{{(c)}}}}
\end{minipage}
\end{center}
\vspace{-0.3in}
\caption{Cost function value ($\min_{\x} \frac{1}{\lambda} \Vert \k\ast\x - \y \Vert_2^2 + \sum_i |x_i|^p$) evaluated on toy 1D signals where $\k$ is either the ground-truth kernel (denoted as `original') or the delta kernel (denoted as `blurred'). Top: Sharp and blurred signals. Bottom: The cost function value (optimized over $\x$) as a function of $p$. As can be observed from the figures, for strong edges as in (a), the cost function value using a sparse $\mathcal{\ell}_p$-norm with $p \approx 1$ can already favor the original sharp image. For smaller, more-refined structures as in (b), a smaller $p$ (meaning a more concave penalty function) is required. For a real world image (c), which is a composition of both large and small scale structures, the cost function values of both the original and blurred signals are similar when the $p$ value is large. With a smaller $p$, the cost function value favors the original sharp image as desired.
}
\label{fig:1D_sig_cost_val}
\end{figure*}

The results presented herein then suggest a sort of paradox:  in Section \ref{sec:EB_BDB} we have argued that VB is actually equivalent to an unconventional form of MAP estimation over $\x$, but with an intrinsic mechanism for avoiding bad local minima, increasing the chances that a good global or near-global minima can be found.  Moreover, at least in the noiseless case ($\lambda \rightarrow 0$), any such minima will be exactly equivalent to the standard MAP solution by virtue of Corollary \ref{cor:vb_cost_reform}. However, based on the noiseless analysis from Levin \emph{et al.} above, any global MAP solution is unlikely to involve the true sharp image when the true image statistics are used for $p(\x)$, meaning that VB performance should be poor as well at a global solution.  Thus how can we reconcile the positive performance of VB actually observed in practice, and avoidance of degenerate no-blur solutions, with Levin \emph{et al.}'s characterization of the MAP cost function?

First, when analyzing MAP Levin \emph{et al.} consider only a flat prior on $\k$ within the constraint set $\sum_i k_i = 1$ and $k_i \geq 0$.  However, MAP estimation may still avoid no-blur solutions when equipped with an appropriate non-flat kernel prior.  Likewise VB naturally produces an explicit penalty factor on $\k$ given by $\log\|\bar{\k} \|_2^2$ (see Theorem \ref{thm:vb_cost_reform}) that favors blurry explanations (non-delta kernel), since the delta kernel will maximize the $\ell_2$ norm.  Moreover, VB introduces this prior in a convenient form devoid of additional tuning parameters, whereas a traditional MAP estimator would generally require some form of cross-validation.

Secondly, and perhaps more importantly, the question of whether $\y$ (the blurry image) or $\x^*$ (the true sharp image) is more heavily favored by the true image prior $p_{{\rm true}}(\x)$ is not really the most relevant issue to begin with.  A more pertinent question is whether there exists some sparse $\hat{\x}$ with $\| \hat{\x} \|_0 < \| \y \|_0$ such that $\x^* \approx \hat{\x}$.  If so, then the solution to the relaxation
\begin{equation} \label{eq:levin_analysis}
\min_{\x,\k} \| \y - \k \ast \x \|_2^2 + \lambda \sum_i |x_i|^{p}, \hspace*{0.3cm} \mbox{s.t. } \sum_i k_i = 1, \ k_i \geq 0
\end{equation}
with $p \ll 1$ is very \emph{unlikely} to be $\x = \y$ and $\k = \delta$.  And this is guaranteed to be true as $p$ becomes sufficiently small, assuming $\lambda$ is set appropriately.  It is crucial to understand here that the exponent $p$ from (\ref{eq:levin_analysis}) need not correspond with the true distribution $p_{{\rm true}}(\x)$, as long as $\x^*$ is reasonably close to some sparse solution.  The point then is that with $p$ small, regardless of $p_{{\rm true}}(\x)$, maximally sparse solutions will be favored, and this is very unlikely to involve the no-blur solution.  Therefore, just because $\x^*$ may not be exactly sparse, we may nonetheless locate a sparse approximation $\hat{\x}$ that is sufficiently reasonable such that the unknown $\k^*$ can still be estimated accurately.

In general, it is more important that \emph{the assumed image prior} $p(\x) = \exp[-\frac{1}{2} \gxbf(\x)]$ \emph{be maximally discriminative with respect to blurred and sharp images, as opposed to accurately reflecting the statistics of real images}.  Mathematically, this implies that it is much more important that we have $p(\k \ast \x^*) \ll p(\x^*)$, or under less than ideal circumstances $p(\k \ast \x^*) \ll p(\hat{\x})$, than we enforce $p(\x) = p_{{\rm true}}(\x)$, even if $p_{{\rm true}}(\x)$ were known exactly.  This is because the sparsity/variance trade-off described above implies that it may often be the case that $p_{{\rm true}}(\k \ast \x^*) > p_{{\rm true}}(\x^*)$ and/or $p_{{\rm true}}(\k \ast \x^*) > p_{{\rm true}}(\hat{\x})$.

To achieve the best results then, we must counteract the negative effects of variance reduction when choosing $p(\x)$.  For example, the penalty selection $\gxbf(\x) = \sum_i |x_i|^{p}$ becomes less sensitive to the image variance as $p$ becomes small, and with $p \rightarrow 0$, it actually becomes completely invariant.  So we may expect that smaller $p$ values are more appropriate for disambiguating blurred from unblurred images.   Importantly, the optimal estimator will generally not equal the generalized Gaussian distribution with $p \approx [0.5,0.8]$, a commonly-reported estimate of true image statistics.

To illustrate these points, we reproduce \citep[Figure 1]{LevinWDF11_PAMI}, where three 1D image slices (an ideal edge, an ideal spike, and a real image slice) are compared with respect to $\sum_i |x_i|^{p}$ both before and after blurring.  However, unlike \citep[Figure 1]{LevinWDF11_PAMI} we do not strictly enforce $\y = \k \ast \x$, but compare sharp and blurred images using the relaxed criterion from (\ref{eq:levin_analysis}).  Figure \ref{fig:1D_sig_cost_val} displays the results, where the optimal value of (\ref{eq:levin_analysis}) is computed as $p$ is varied using both the delta kernel $\k = \delta$ and the true blur kernel $\k = \k^*$.   For strong edges (the simplest case), any $p \leq 1$ is sufficient for avoiding the delta solution, while when finer structures need to be resolved, we require that $p\ll1$.  While panels (a) and (b) reflect the results and conclusions from \citep[Figure 1]{LevinWDF11_PAMI}, panel (c) tells a very different story.  Basically, whereas \citep[Figure 1]{LevinWDF11_PAMI} shows the delta kernel (blurred image) being favored even for small values of $p$, the relaxed condition (\ref{eq:levin_analysis}) strongly prefers the true blur kernel $\k^*$ for a wide range of $p$.

\begin{figure}[t]
\centering
\begin{overpic}[width=2.5cm]{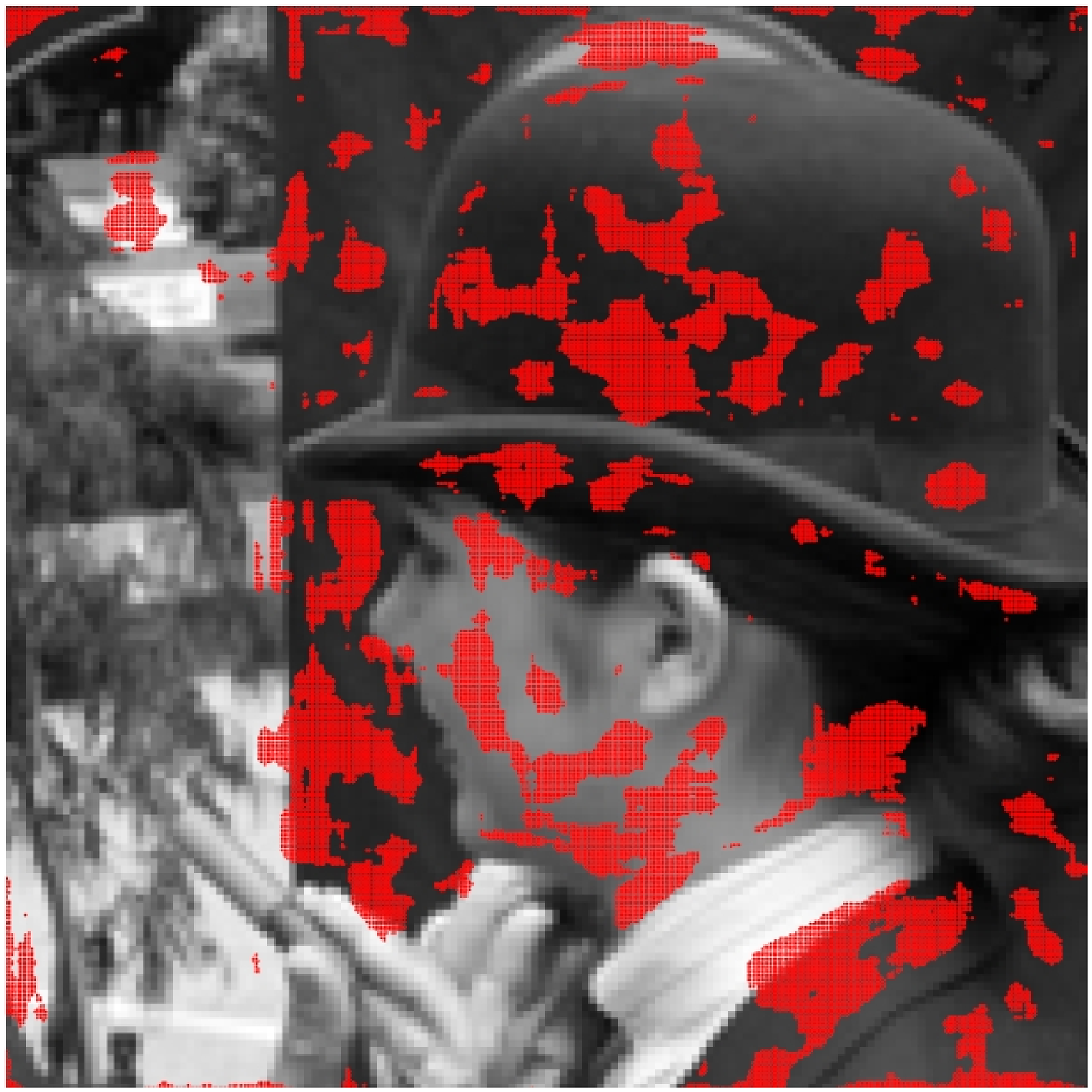}
\put(92,103){ \sffamily \footnotesize{\textcolor{black}{{\textsc{{Separate Penalty}}}}}}
\put(25,-10){ \sffamily \footnotesize{\textcolor{black}{$p = 0.5$}}}
\end{overpic}
\begin{overpic}[width=2.5cm]{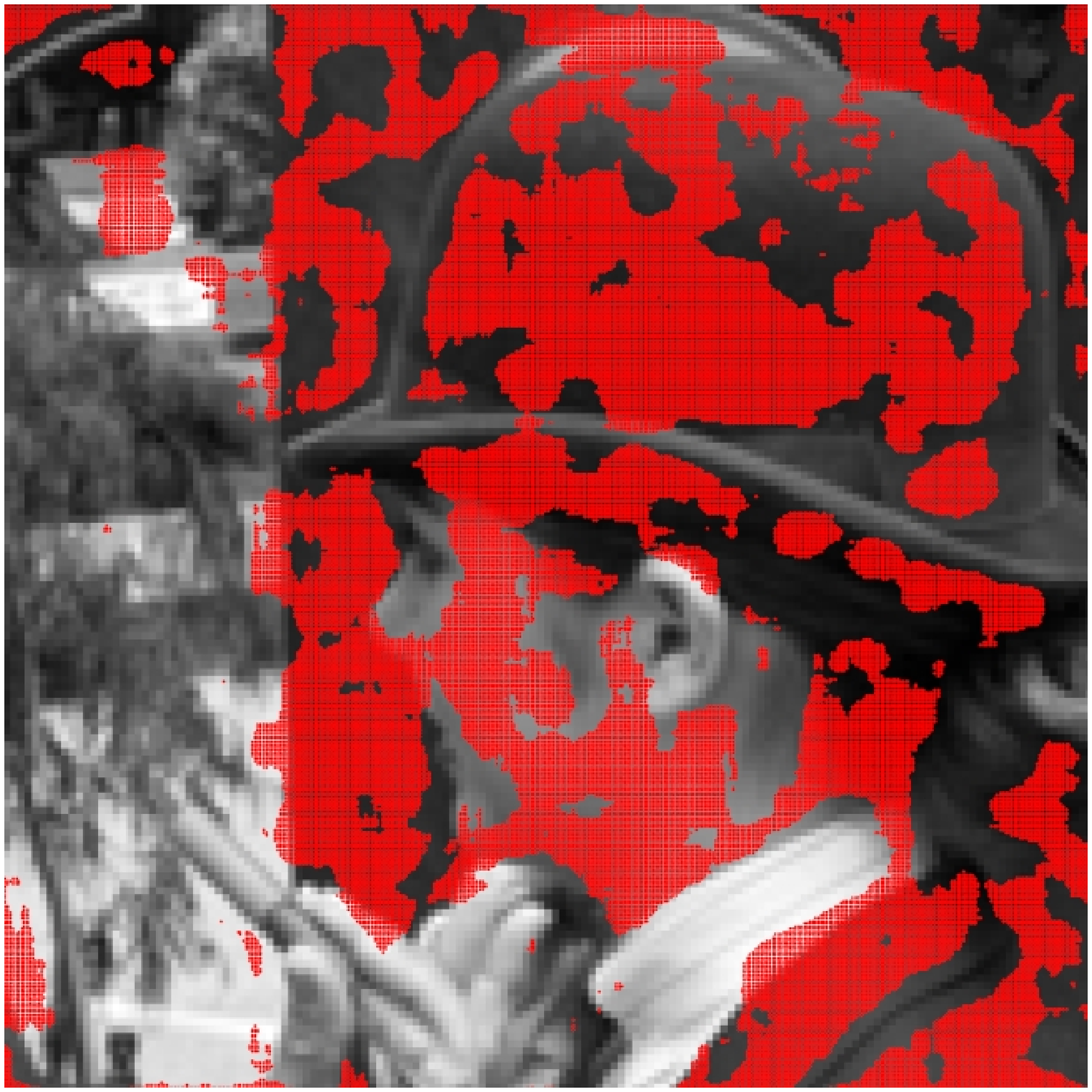}
\put(25,-10){ \sffamily \footnotesize{\textcolor{black}{$p = 0.3$}}}
\end{overpic}
\begin{overpic}[width=2.5cm]{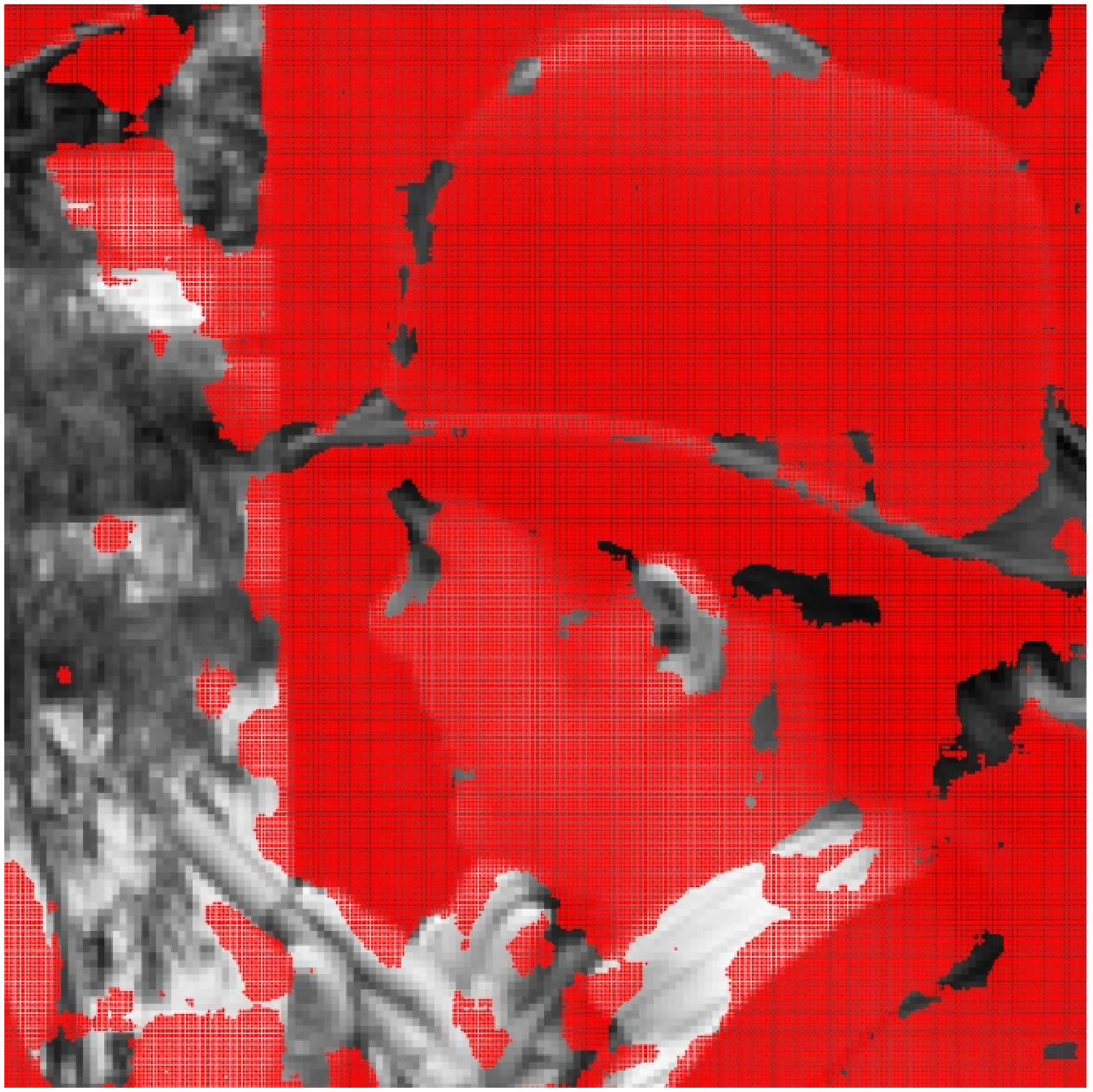}
\put(25,-10){ \sffamily \footnotesize{\textcolor{black}{$p = 0.1$}}}
\end{overpic}
\hspace{0.2in}
\begin{overpic}[ width=2.5cm]{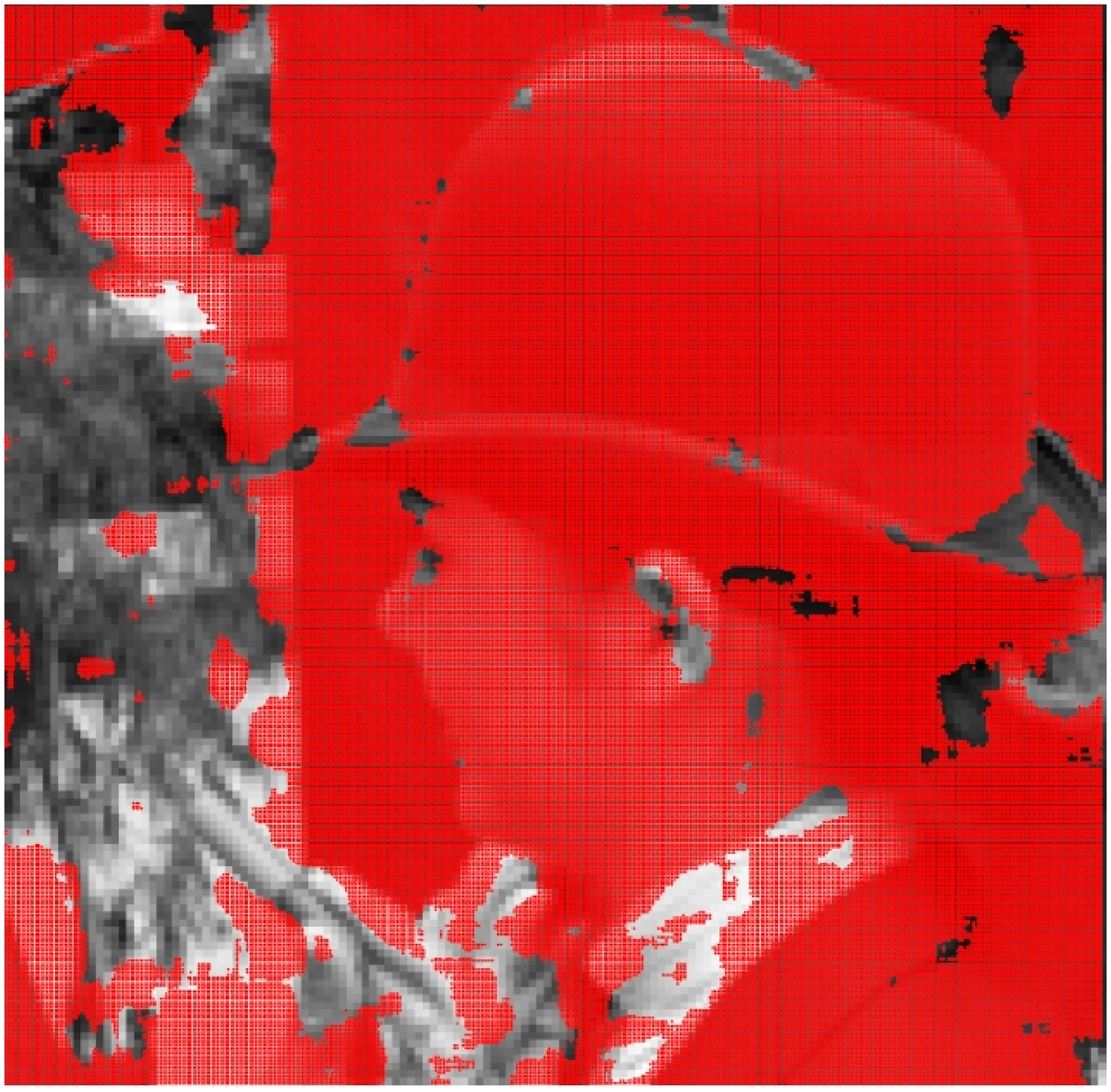}
\put(-10,103){ \sffamily \footnotesize{\textcolor{black}{{\textsc{{Coupled Penalty}}}}}}
\end{overpic}
\caption{A 2D illustrative example. $15 \times 15$ patches in which the true solution (sharp explanation) is favored are marked in red.
Top row shows the results using the penalty function $\sum_i |x_i|^{p}$ with different ${p}$ values. Note that here $\x$ denotes the gradient image.
The bottom row shows the result from the coupled penalty emerging from VB per our analysis in Section \ref{sec:EB_BDB}. Clearly, the percentage of  local patches favoring the true solution increases with decreasing $p$. For sufficiently small $p$ value, the pattern of favored patches is similar to that from the coupled penalty function.}
\label{fig:2D_image_cost_val}
\end{figure}

Generalizing to real 2D images, to ensure practical success and capture high-resolution details, lower values of $p$ are definitely preferred for limiting the undesirable effects of variance reduction mentioned above.  To visualize this claim, we now present a revised version of \citep[Figure~2]{LevinWDF11_PAMI}, which depicts the regions of a real image where the sharp image is preferred to the undesirable blurred solution, where relative preference is rated by $\sum_i |x_i|^{p}$, with $p = 0.5$.  In Figure~\ref{fig:2D_image_cost_val} we update \citep[Figure 2]{LevinWDF11_PAMI} to include both lower $p$ values as well as the analogous ranking using the $\gvb$ from (\ref{eq:penalty_fun2}), with $\rho$ near zero.  From this figure, it is readily apparent that smaller $p$ values are more effective, and that the VB penalty function does indeed behave like the $\ell_p$ norm when $\rho$ is small.

Of course from a practical standpoint solving (\ref{eq:levin_analysis}) represents a difficult, combinatorial optimization problem with numerous local minima when $p$ is small, and we speculate that many have tried such direct minimization and concluded that smaller $p$ values were inadequate.  However, the penalty function shape modulation intrinsic to VB ultimately provides a unique surrogate for circumventing this problem, hence its strong performance.  Thus we can briefly summarize largely why VB can be superior to MAP:  \emph{VB allows us to use a near-optimal image penalty, one that is maximally discriminative between blurry and sharp images, but with a reduced risk of getting stuck in bad local minima during the optimization process}.  Overall, these conclusions provide a more complete picture of the essential differences between MAP and VB.

Before proceeding to the next section, we emphasize that none of the arguments presented herein discredit the use natural image statistics when directly solving (\ref{eq:MAP_k}).  In fact Levin \emph{et al.}~\citep{LevinWDF11_PAMI} prove that when $p(\x) = p_{{\rm true}}(\x)$, then in the limit as the image grows large the MAP estimate for $\k$, after marginalizing over $\x$ (Type II), will equal the true $\k^*$.  But there is no inherent contradiction with our results, since it should now be readily apparent that VB is fundamentally different than solving $\min_{\k} p(\k|\y)$, and therefore justification for the latter cannot be directly transferred to justification for the former.  This highlights the importance of properly differentiating various forms of Bayesian inference.

Natural image statistics are ideal in cases where $\y$ and $\x$ grow large and we are able to integrate out the unknown $\x$, benefitting from central limit arguments when estimating $\k$ alone.  However, when we jointly compute MAP estimates of both $\x$ and $\k$ (Type I) as in (\ref{eq:reg_bd}), we enjoy no such asymptotic welfare since the number of unknowns increases proportionally with the sample size.  One of the insights of our paper is to show that, at least in this regard, VB is on an exactly equal footing with Type I MAP, and thus we must look for theoretical VB justification elsewhere, leading to the analysis of relative concavity, local minima, invariance, maximal sparsity, etc. presented herein.

% ***************************

\section{Learning $\lambda$} \label{sec:learn_lambda}

While existing VB blind deconvolution algorithms typically utilize some preassigned decreasing sequence for $\lambda$ as described in Section \ref{sec:noise_depend_analysis} and noted in Algorithm \ref{algo:algo1}, it may be preferable to have $\lambda$ learned automatically from the data itself as is common in other applications of VB.  This alternative strategy also has the conceptual appeal of an integrated cost function that is universally reduced even as $\lambda$ is updated, unlike Algorithm \ref{algo:algo1} where the $\lambda$ reduction step may in fact increase the overall cost, unlike all of the other updates.  However, current VB deblurring papers either do not mention such a seemingly obvious alternative (perhaps suggesting that the authors unsuccessfully tried such an approach) or explicitly mention that learning $\lambda$ is problematic but without concrete details.  For example, \citep{LevinWDF11_PAMI} observed that the noise level learning used in \citep{Fergus06removingcamera} represents a source of problems as the optimization  diverges  when the estimated noise level decreases too much. But there is no mention of why $\lambda$ might decrease too much, and further details or analyses are absent.

Interestingly, the perspective presented herein provides a clear picture for why learning $\lambda$ may be difficult and suggests some potential fixes.  The problem stems from a degenerate global minimum to (\ref{eq:vb_cost_reform}) and therefore (\ref{eq:VB_DB}) occurring at $\lambda = 0$.  The explanation for this is as follows:  Consider minimization of (\ref{eq:vb_cost_reform}) over $\x$, $\k$, and $\lambda$.  Because the combined dimensionality of $\k$ and $\x$ is larger than $\y$, there are an infinite number of candidate solutions such that $\y = \k\ast\x$.  Therefore the term $\frac{1}{\lambda} \left\|\y - \k \ast \x \right\|_2^2$ in the VB cost function (\ref{eq:vb_cost_reform}) can be minimized to exactly zero even in the limit as $\lambda \rightarrow 0$.  Moreover, by Theorem \ref{thm:general_sparse_promoting}, we observe that $\gvb(x_i,\rho)$ becomes increasingly small as $\rho \rightarrow 0$ around solutions where $x_i$ is small or near zero.  In fact, it can actually be shown that $\gvb(0,\rho) \rightarrow -\infty$ as $\rho \rightarrow 0$ for all non-decreasing $f$.\footnote{Based on (\ref{eq:general_f_penalty1}), it is clear that the optimizing $\gamma_i$ value for computing $\gvb(0,\rho)$ will be $\gamma_i = 0$.  When $\rho \rightarrow 0$, we then have $\log(\gamma_i + \rho) \rightarrow -\infty$, and therefore $\gvb(0,\rho) \rightarrow -\infty$.  Graphically, Figure \ref{fig:penalty_fun} (b) also reveals this effect, showing that if we were to jointly minimize over both $x$ and $\rho$, the $\{0,0\}$ solution is heavily favored.}

Now because of the disparity in dimensionality mentioned above, there will always be feasible solutions to $\y = \k \ast \x$ with at least $m-n$ or more elements of $\x$ equal to zero.  Thus, at any one of these solutions the the VB cost function (\ref{eq:vb_cost_reform}) can then be driven to $-\infty$ with $\lambda \rightarrow 0$.  Unless the true $\x$ actually has many exactly zero-valued elements, this will represent a globally degenerate minimizing solution for a broad class of $f$.  And even for other choices for $f$, a slightly more subdued form of this same degeneracy will still exist since the VB-specific regularization fundamentally favors $\lambda$ being small: essentially the $\log(\gamma_i + \rho)$ factor in (\ref{eq:general_f_penalty1}) will always favor $\rho$, and therefore $\lambda$ being small.  The $1/\lambda$ weighting of $\left\|\y - \k \ast \x \right\|_2^2$ is not sufficient for counteracting this effect given the multitude of feasible solutions such that $\y = \k\ast\x$. Thus our framework makes it very clear that we should never expect to optimize over $\lambda$ and expect to achieve satisfactory results with VB, explaining the empirical observation from \citep{LevinWDF11_PAMI} mentioned above. In contrast, the point is considerably obfuscated when we examine the previous VB free energy-based cost function (\ref{eq:VB_DB}) directly.

Fortunately, this new view of VB naturally offers some potential fixes for noise level estimation while avoiding these types of undesirable degenerate solutions.  Perhaps the simplest approach is to include an additional $d/\lambda$ penalty factor in (\ref{eq:vb_cost_reform}), where $d > 0$ is a user-specified constant.  As justification for this inclusion, note that this added factor is proportional to $1/\lambda \left\|\y - \k \ast \x \right\|_2^2$, but acts as a barrier preventing $\lambda$ from ever going to zero even if $\y = \k \ast \x$.  In fact it is easily shown (see Appendix B) that any $\lambda$ minimizing the cost function (\ref{eq:vb_cost_reform}) augmented with the penalty $d/\lambda$ must satisfy $\lambda \geq d/n$, which can be viewed as a lower-bound on what $1/n\left\|\y - \k \ast \x \right\|_2^2$ should be.  % Appendix B contains the details, as well as the associated updates for estimating $\lambda$

% While admittedly the extra hyperparameter $d$ must be tuned to some degree,

In practice, we have found the fixed value $d = n\times10^{-4}$ to be highly effective across a wide range of images and testing scenarios, including all reported results in Section \ref{sec:Exp} and numerous real-world experiments not shown \citep{MDB_CVPR13,NBD_inprep}.  Regardless, use of a single, fixed value is likely to be less burdensome than producing an entire $\lambda$ reduction schedule, which also requires a user-specified minimal $\lambda$ value anyway.  Moreover, the VB update rules only require a slight modification to account for this additional term while retaining existing convergence properties (see Appendix B for the derivation).  By estimating the noise level together with the image and kernel, we not only make the deblurring algorithm more noise-aware and mostly parameter-free. More profoundly, by initializing with a large value and allowing the iterations to learn the optimal reduction schedule, it offers a natural coarse-to-fine process for blind deblurring, which has been found as one of the crucial factors for blind deblurring algorithms as discussed above.  The experimental results from Section \ref{sec:Exp} support this conclusion. % , although further testing is warranted.

\begin{algorithm}[tb]
\caption{VB Blind Deblurring with Jeffreys Prior and Learned $\lambda$ (VB-Jeffreys).}
\begin{algorithmic}[1]
\STATE {\bf Input:} {a blurry image $\y$}, noise level estimation hyper-parameter $d = n\times10^{-4}$
%\STATE {\bf Output:} {estimated blur kernel ${\k^*}$ and sharp image ${\boldsymbol{\mu}}$}
\STATE {\bf Initialize:}   blur kernel ${\k}$, noise level $\lambda$
% {\bf Initialization:} sparse vector $\hat{\af}$ recovered from $\y$ in terms of $\D$, and $\hat{\x} = \D\hat{\af}$.\\
\STATE  {\bf While} stopping criteria is not satisfied, do same as Algorithm \ref{algo:algo1} except for the following changes:
{
 \begin{itemize}

\item $\omega_i  \leftarrow \sigma_i^{-2}, \forall i$
\vspace*{0.3cm}
\item ${\lambda} \leftarrow \frac{\Vert \y - \boldsymbol{\mu}\ast\k \Vert_2^2 + \sum_i (\Vert \bar{\k} \Vert^2_2 \cdot C_{ii}) + d}{n}$

\end{itemize}
}
\STATE {\bf End}
\end{algorithmic}
\label{algo:algo_EB_BDB}
\end{algorithm}

% \label{eq:update_noise_level}

%%==================

\section{Experimental Results}\label{sec:Exp}

We emphasize that the primary purpose of this paper is the formal analysis of existing state-of-the-art VB blind deconvolution methodology, not the development of a new practical system per se.  Empirical support for recent VB algorithms, complementary to our theoretical presentation, already exist in \citep{BabacanMDK12,LevinWDF11,MDB_CVPR13}.  Nonetheless, motivated by our results herein, we will briefly evaluate two simple refinements of Algorithm \ref{algo:algo1} that help corroborate some of our analytical findings while demonstrating that an extremely simplified version of VB, albeit with theoretically sound underpinnings, can perform well against published state-of-the-art MAP and VB algorithms with considerably more complexity and/or manual parameters.  In doing so, we hope to motivate the optimal usage of VB for more sophisticated and realistic blind deblurring problems.

To this end  we will (i) use an image prior obtained when $f$ is flat (Jeffreys prior) as motivated in Section \ref{sec:EB_BDB} instead of a prior based on natural image statistics, and (ii) we will learn the $\lambda$ parameter automatically per the discussion in Section \ref{sec:learn_lambda}.  The revised estimation steps are summarized in Algorithm~\ref{algo:algo_EB_BDB}, which is obtained by adopting the same procedure from Algorithm 1 under the special case $f(\gamma) = b$ and with the $\lambda$ updates derived in Appendix B. We will refer to this algorithm as  VB-Jeffreys (since the underlying image prior is based on the improper Jeffreys distribution as described previously).  Note that estimation is performed in the gradient domain; however, the recovered kernel is applied to a non-blind deconvolution step to obtain the final latent image estimate.  This final non-blind step, taken from \citep{LevinWDF11}, is standardized across all algorithms compared in this section.

Given this variant of VB, we reproduce the experiments from \citep{LevinWDF11} using the useful benchmark test data from~\citep{Understanding_BD}.\footnote{This data is available online at \url{ http://www.wisdom.weizmann.ac.il/~levina/papers/LevinEtalCVPR09Data.rar} }  This consists of 4 base images of size $255\times 255$ and 8 different blurring effects, leading to a total of 32 blurry images.  Ground truth blur kernels were estimated by recording the trace of focal reference points on the boundaries of the sharp images (see \citep[Figure 7]{LevinWDF11_PAMI} and related text for details of the experimental setup and data collection).  The kernel sizes range from $13\times 13$ to $27\times 27$.  All evaluations are based on the SSD (Sum of Squared Difference) metric defined in~\citep{Understanding_BD}, which quantifies the error between estimated and the ground-truth images. To normalize for the fact that harder kernels give a larger image reconstruction error even when the true kernel is known (because the corresponding non-blind deconvolution problem is also harder), the SSD ratio between the image deconvolved with the estimated kernel and the image deconvolved with the ground-truth kernel is used as the final evaluation measure.

\begin{figure}[t]
\centering
\begin{overpic}[viewport = 10 0 415 300, clip, width=7cm]{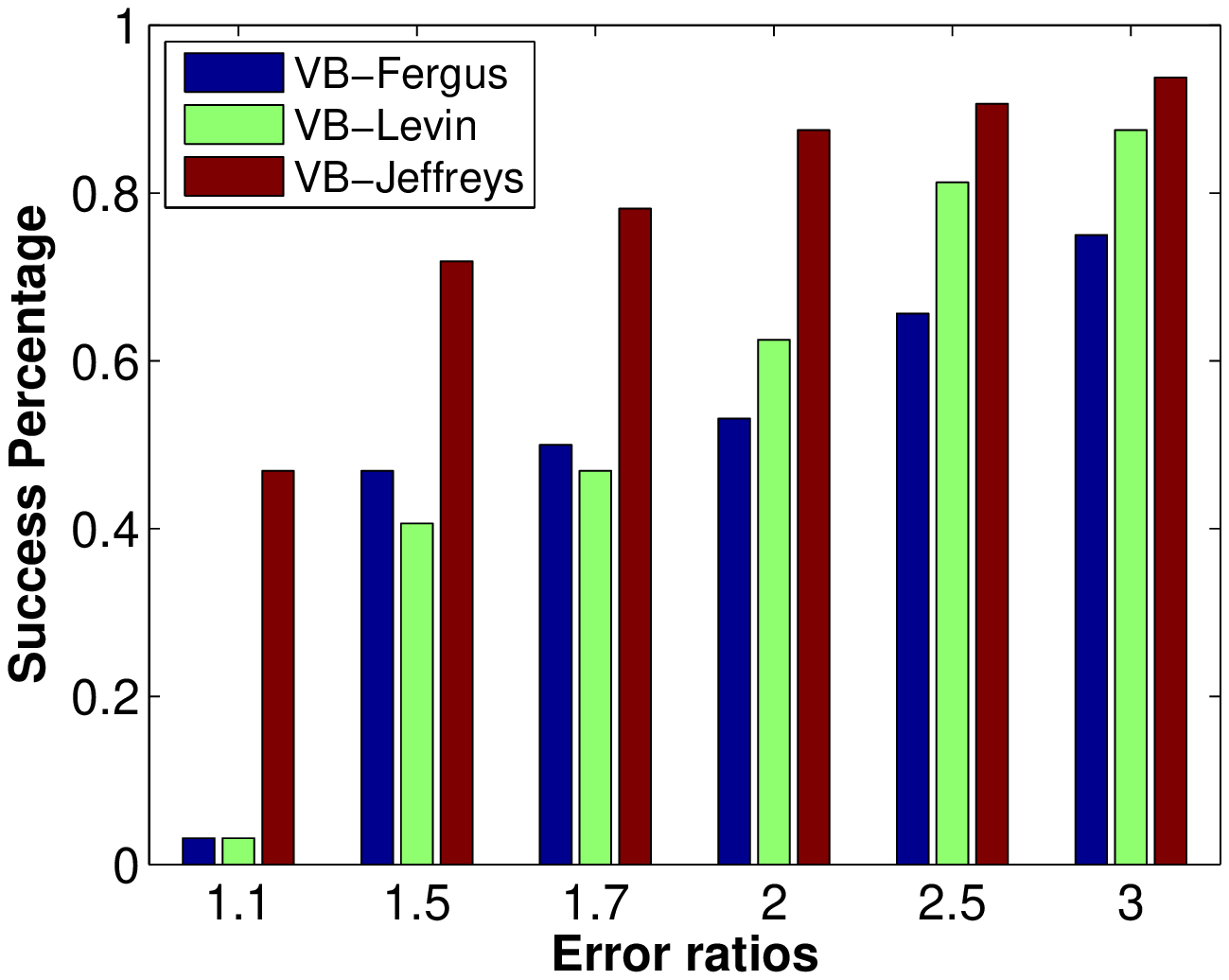}
\put(47,-3.5){ \sffamily \footnotesize{\textcolor{black}{{(a)}}}}
\end{overpic}
\begin{overpic}[viewport = 10 0 415 300, clip,  width=7cm]{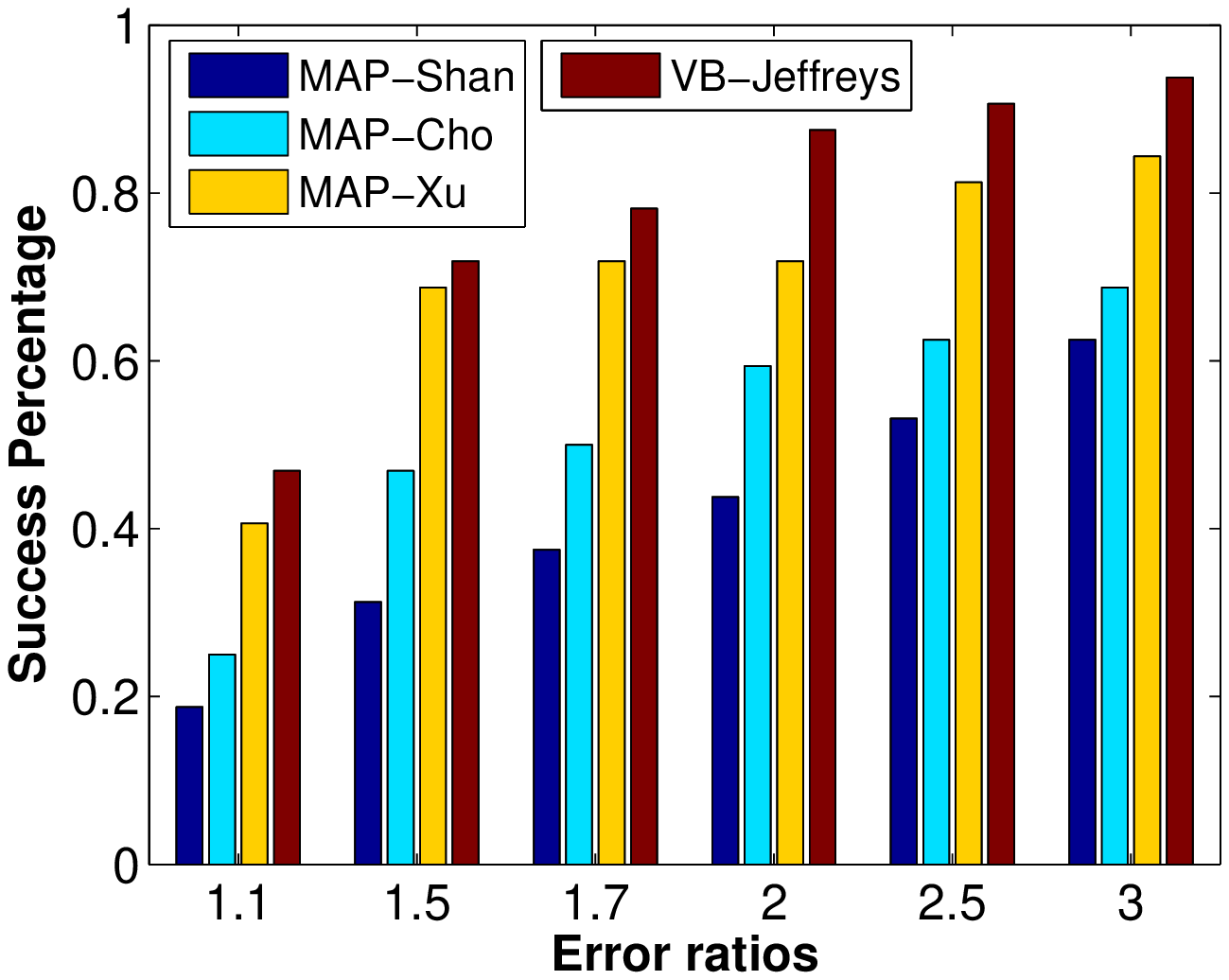}
\put(47,-3.5){ \sffamily \footnotesize{\textcolor{black}{{(b)}}}}
\end{overpic}
\caption{Evaluation of the restoration results: Cumulative histogram of the deconvolution error ratio across 32 test examples. The height of the bar indicates the percentage of images having error ratio below that level. High bars indicate better performance. (a) comparison with several other VB algorithms.  (b) comparison with several state-of-the-art MAP algorithms. }
\label{fig:acc_hist}
\end{figure}

We first compare the VB-Jeffreys method described in Algorithm~\ref{algo:algo_EB_BDB} with the related variational Bayesian methods from Fergus \emph{et al.}~\citep{Fergus06removingcamera} and Levin \emph{et al.}~\citep{LevinWDF11}, labeled VB-Fergus and VB-Levin respectively, which accompany the dataset.  While they can both be effective in practice, they have not been optimized with respect to the considerations provided herein, and specific prior selections have not been rigorously motivated.  Instead, these priors are loosely based on the statistics of natural scenes and, as we have argued in Sections \ref{sec:EB_BDB} and \ref{sec:relationship_levin}, may not be optimal.  The cumulative histogram of the SSD error ratios is shown in Figure~\ref{fig:acc_hist} (a).
The height of the bar indicates the percentage of images having error ratio below that level. High bars indicate better performance.
As mentioned by Levin \emph{et al.}, the results with error ratios above $2$ may already have some visually implausible regions~\citep{Understanding_BD}.
The VB-Jeffreys algorithm can achieve close to $90\%$ success with error ratio below $2$, significantly higher than the others.

Regardless, all of the VB algorithms still exhibit reasonable performance, especially given that they do not benefit from any additional prior information or regularization heuristics that facilitate blur-adaptive structure selection (meaning the additional regularization based on domain knowledge added to (\ref{eq:reg_bd}) that boost typical MAP algorithms as discussed previously). However, one curious phenomenon is that both VB-Fergus and VB-Levin experience a relatively large drop-off in performance when the error ratio reduces from 1.5 to 1.1.  While it is difficult to be absolutely certain, one very plausible explanation for this decline relates to the prior selection employed by these algorithms.  In both cases, the prior is based on a finite mixture of zero mean Gaussians with different variances roughly matched to natural image statistics.  While such a prior does heavily favor approximately sparse signals, it will never produce any exactly sparse estimates at any resolution of the course-to-fine hierarchy, and hence, especially at high resolutions the penalty shape modulation effect of VB will be highly muted, as will be the beneficial sparsity/variance trade-off that accompanies more strongly sparse priors. Thus these algorithms may not be optimal for resolving extremely fine details, which is required for reliably producing image estimates with low error ratios.  In contrast, to achieve high error ratios only lower resolution features need be resolved, and in this regime VB-Levin, which is the closest algorithmically to VB-Jeffreys, performs nearly as well as VB-Jeffreys.  Again, this reinforces the notion that natural image statistics may not be the optimal basis for image priors within the VB framework.

% ***************

We next compare VB-Jeffreys with several state-of-the-art MAP algorithms from Shan \emph{et al.}~\citep{hqdeblurring_siggraph2008}, Xu \emph{et al.}~\citep{XuJ10_ECCV}, and Cho \emph{et al.}~\citep{fast_motion_deblur_2009}.  Shan \emph{et al.} (denoted MAP-Shan) adopts an additional local smoothness prior designed to reduce ringing artifacts.
Xu \emph{et al.}  (MAP-Xu) includes two phases for kernel estimation and incorporates an explicit scheme for edge structure selection.
Finally, Cho \emph{et al.} (MAP-Cho) is also a carefully-engineered MAP approach coupled with structure selection and sharp edge prediction schemes, which help the algorithm to avoid the degenerate delta solution.  Recall that previously we have argued that standard MAP algorithms may suffer from one of two problems: either the pixel-wise image prior is highly sparse and convergence to sub-optimal local solutions becomes a problem, or the prior is less sparse and global solutions do not sufficiently distinguish blurry from sharp images.  All of the MAP algorithms tested here can be viewed as addressing this conundrum by including additional regularization schemes (priors) such that global or near global minima favor sharp images even when the basic pixel-wise image prior is convex (i.e., minimally sparse).  This is a very different strategy than VB, which adopts a simpler underlying model with no additional regularizers beyond the canonical pixel-wise sparse prior.  Figure~\ref{fig:acc_hist} (b) reveals that the simple VB strategy, when properly implemented, can still outperform specially tuned MAP estimates.  Note that the results of MAP-Cho are from the dataset accompanying \citep{LevinWDF11} directly, while the results of MAP-Shan and MAP-Xu are produced using the software provided by the authors, for which we adjust the parameters carefully.  For all algorithms we run every test image with the same parameters, similar to~\citep{Understanding_BD, LevinWDF11_PAMI}.  Overall, VB-Jeffreys obtains the highest reported result of any existing algorithm on this important benchmark.

\section{Conclusion}\label{sec:con}
This paper presents an insightful reformulation and subsequent analysis of MAP and VB blind deconvolution algorithms revealing why practical success is possible and suggesting valuable improvements for the latter.    We summarize the contributions of this perspective as follows:

\begin{itemize}

\item Beginning with their influential work from \citep{Understanding_BD},  Levin \emph{et al.}  have provided an interesting analysis of VB and related MAP algorithms.  We push the limits of understanding much further with a thorough, complementary investigation.

\item We demonstrate that rigorous evaluation of VB and its associated priors cannot be separated from implementation heuristics, and we have meticulously examined the interplay of the relevant underlying algorithmic details employed by practical VB systems.  Consequently, what may initially appear to be a plausible rationale for achieving high performance may have limited applicability given the assumptions required to implement scalable versions of VB.

\item We have proven that in an ideal, noiseless setting, VB and MAP have an identical underlying cost function once the requisite approximations are accounted for.  This is in direct contrast to conventional assumptions explaining the presumed performance advantages of VB.

    % nor can it be chosen without regard for the specific algorithmic context in which it is used

\item We carefully examine the underlying VB objective function in a transparent form, leading to principled criteria for choosing the optimal image prior.  It is crucial to emphasize that this image prior need not, and generally should not, reflect the most accurate statistics of real imaging data.  Instead, the preferred distribution is one that is most likely to guide VB iterations to high quality global solutions by strongly differentiating between blurry and sharp images.  In this context, we have motivated a unique selection, out of the infinite set of possible sparse image priors, that simultaneously allows for maximal discrimination between $\k \ast \x$ and $\x$, displays a desirable form of scale invariance, and leads to an intrinsic coupling between the blur kernel, noise level, and image penalty such that bad local minima can largely be avoided.  To the best of our knowledge, this represents a completely new viewpoint for understanding VB algorithms.

\item The cause of failure when using standard MAP algorithms depends on the choice of image prior.  If $-2\log p(\x)$ is only marginally concave in $|\x|$, or is tuned to natural image statistics, then the problem is often that global or near-global solutions do not properly differentiate blurry from sharp images.  In contrast, if $p(\x)$ is highly sparse, while global solutions may be optimally selective for sharp images, convergence to bad local solutions is more-or-less inevitable.  It is with the latter that VB offers a compelling advantage.

% \item We provide several analytical arguments, and supporting empirical results, as to why the optimal prior need not follow from the statistics of natural images.

\item We have shown why VB deconvolution algorithms are fundamentally ill-equipped to learn the noise level, and proposed at least one potential way to work around this problem.

\item By reframing VB as a nearly parameter-free sparse regression problem in standard form, we demonstrate that it is no longer difficult to enhance performance and generality by inheriting additional penalty functions (such as those from \citep{hqdeblurring_siggraph2008}) or noise models (e.g., Laplacian, Poisson, etc.) commonly reserved for MAP.  Moreover, we anticipate that these contributions will lead to a wider range of principled VB applications, such as non-uniform deconvolution~\citep{WhyteSZP12,Zhu_PAMI} and multi-frame and video deblurring~\citep{SroubekM12,TakedaM11}.  Preliminary results show tremendous promise \citep{MDB_CVPR13,NBD_inprep}. Additionally, the analysis we conducted for blind deconvolution may well be relevant to other related problems like robust dictionary learning in the presence of noise. % We will investigate these and other topics in  future work.

% \item By recasting VB as a nearly parameter-free, yet unconventional form of MAP estimation, we show that VB is uniquely positioned for numerous extensions.

\end{itemize}

Overall, we hope that these observations will ensure that VB is not under-utilized in blind deconvolution and related tasks.  We conclude by mentioning that, given the new perspective on VB provided herein, it may be possible to derive new blind deblurring algorithms and penalty functions that deviate from the VB script but nonetheless adopt some of its attractive properties.  This is a direction of ongoing research.

\section*{Appendix A: Proofs}

\subsection*{Proof of Theorem \ref{thm:vb_cost_reform}}
We begin with the cost function
\begin{equation} \label{eq:cost_fun_L2}
\mathcal{L}(\x, \k, \gam) \triangleq  \frac{1}{\lambda} \Vert \y - \k\ast \x \Vert_2^2 +  \sum_i \left[ \frac{x_i^2}{\gamma_i}  + \log(\lambda + \Vert \bar{\k}\Vert^2_2 \gamma_i ) +  f(\gamma_i) \right],
\end{equation}
%\begin{split}
%\mathcal{L}(\x, \k, \gam) & \triangleq  \frac{1}{\lambda} \Vert \y - \k\ast \x \Vert_2^2   \\
%&+  \sum_i \left[ \frac{x_i^2}{\gamma_i}  + \log(\lambda + \Vert \bar{\k}\Vert^2_2 \gamma_i ) +  f(\gamma_i) \right],
%\end{split}
%\end{eqnarray}
\noindent which is obtained starting with (\ref{eq:vb_cost_reform}) and then simply removing the minimization over $\gam$ from the definition of $\gvb$ in (\ref{eq:general_f_penalty1}), plugging in the value of $\rho$, and simplifying.  The basic strategy here will be to use a majorization-minimization approach \citep{MM} akin to the concave-convex procedure \citep{CCCP_NIPS03} to derive coordinate-wise updates that are guaranteed to reduce or leave unchanged $\mathcal{L}(\x, \k, \gam)$, and then show that these are in fact the same updates as Algorithm \ref{algo:algo1}.  In doing so we show that (\ref{eq:vb_cost_reform}) is an equally valid explanatory cost function with which to interpret VB.

As an initial proposition, we may attempt to directly minimize $\mathcal{L}(\x, \k, \gam)$ over $\x$, $\k$, and $\gam$ independently, in each case while holding the other two variables fixed.  Beginning with $\x$, we collect relevant terms and find that we must solve
\begin{eqnarray}\label{eq:image_update}
  \min_{\x} \frac{1}{\lambda}\Vert \y - \k\ast \x \Vert_2^2 +  \sum_i \frac{x_i^2}{\gamma_i},
\end{eqnarray}
which has a convenient closed-form  solution $\x^{\rm opt}$ given by
\begin{eqnarray} \label{eq:x_update_app}
\begin{split}
\x^{\rm opt} & =  \left[\frac{1}{\lambda} \HH^T \HH +  \GM^{-1} \right]^{-1}   \frac{1}{\lambda} \HH^T \y,
\end{split}
\end{eqnarray}
where $\GM \triangleq \mbox{diag}[\gam]$ and $\HH$ is the convolution matrix of the blur kernel defined in Section~\ref{sec:notation}.

%It can be seen that this update rule is the same as the one used in the VB approach.  Note that as the optimal ${\x}$  corresponds to the mean of the Gaussian distribution  $q(\x)$, thus   $\boldsymbol{\mu}$ is used for denoting the optimal $\x$ in Algorithm~\ref{algo:algo1} and Algorithm~\ref{algo:algo_EB_BDB}.

% ******************

Next we consider updating $\gam$, where the associated cost function conveniently decouples so we may solve for each $\gamma_i$ independently.  For this purpose, we use the fact that
\begin{equation}
\lambda + \Vert \bar{\k}\Vert^2_2 \gamma_i = \lambda \gamma_i \left( \frac{1}{\gamma_i} + \frac{\Vert \bar{\k}\Vert^2_2}{\lambda} \right)
\end{equation}
to obtain the following minimization problem for each $\gamma_i$:
\begin{equation}\label{eq:g_fun}
\min_{\gamma_i\ge 0}  \frac{x_i^2}{\gamma_i} + \log \gamma_i  + \log \left[ \frac{\Vert \bar{\k} \Vert^2_2}{\lambda} + \gamma_i^{-1} \right]  +  f(\gamma_i),
\end{equation}
where $\gamma_i$-independent terms are omitted.  Because no closed-form solution is available, we instead use basic principles from convex analysis to form a strict upper bound that will facilitate subsequent optimization.  In particular, we use
\begin{equation} \label{eq:gamma_bound}
\frac{z_i}{\gamma_i} - \phi^*(z_i) \geq  \log \left[ \frac{\Vert \bar{\k} \Vert^2_2}{\lambda} + \gamma_i^{-1} \right],
\end{equation}
which holds for all $z_i \geq 0$, where $\phi^*$ is the concave conjugate \citep{cvx} of the concave function $\phi(\alpha) \triangleq  \log \left[ \frac{\Vert \bar{\k} \Vert^2_2}{\lambda} + \alpha \right]$.  It can be shown that equality in (\ref{eq:gamma_bound}) is obtained using
\begin{eqnarray}\label{eq:update_z}
 \begin{split}
{z}_i^{\rm opt} = \left. \frac{\partial \phi}{\partial \alpha} \right|_{\alpha = \gamma_i^{-1}} = \frac{1}{\frac{\sum_j k_j^2 \bar{I}_{ji}}{\lambda} + \gamma_i^{-1}}, \forall i,
 \end{split}
\end{eqnarray}
where we have used the fact that $\Vert \bar{\k}\Vert^2_2 \triangleq \sum_{j} k_{j}^2 \bar{I}_{ji}$ is the squared norm of $\k$ reincorporating the $i$-dependent image boundary conditions (see Section~\ref{sec:notation}), which will become somewhat relevant for a more comprehensive version of the proof.  Plugging (\ref{eq:gamma_bound}) into (\ref{eq:g_fun}) we obtain the revised problem
\begin{equation}\label{eq:g_fun2}
\min_{\gamma_i\ge 0}  \frac{x_i^2 + z_i}{\gamma_i} + \log \gamma_i  + f(\gamma_i).
\end{equation}
This sub-problem can be handled in multiple ways.  First, if the underlying $g_x$ associated with $f$ (obtained from (\ref{eq:convex_prior})) is differentiable, then (\ref{eq:g_fun2}) has a convenient closed-form solution obtained as follows.  After a $\exp[-1/2(\cdot)]$ transformation (\ref{eq:g_fun2}) assumes the same variational form as the sparse prior given by (\ref{eq:convex_prior}) evaluated at the point $\sqrt{x_i^2 + z_i}$, ignoring irrelevant constants.  Consequently, based on \citep{Wipf_VEM_NIPS05} we know that the optimizing $\gamma_i$ is given by
\begin{equation} \label{eq:gam_update_app}
\gamma_i^{\rm opt} = \left. \frac{2\sigma}{\gx'({\sigma})} \right|_{\sigma = \sqrt{x_i^2 + z_i}}, \forall i.
\end{equation}
This covers the vast majority of practical sparse priors (and all of those amenable to Algorithm \ref{algo:algo1}). Secondly, if for some reason $g_x$ is not differentiable at some point(s), then (\ref{eq:g_fun2}) may still be solved numerically as a 1D optimization problem, or perhaps analytically leveraging the structure of $f$.  For example, if $f$ is a non-decreasing function (as motivated in Section \ref{sec:implicit_penalty}), then $g_x$ will not be differentiable at zero.  However, since $\gamma_i^{\rm opt} = 0$ whenever $x_i^2 + z_i = 0$, so this does not pose a problem.

%% new paragraph
\quad We now examine optimization over $\k$.  Isolating terms, this requires that we solve
\begin{equation} \label{eq:k_fun}
\min_{\k \geq 0}  \frac{1}{\lambda} \Vert \y - \k\ast \x \Vert_2^2 + \sum_i \log \left[ \frac{\Vert \bar{\k} \Vert^2_2}{\lambda} + \gamma_i^{-1} \right].
\end{equation}
There is no closed-form solution; however, as before we may use strict upper bounds derived from convex analysis for optimization purposes.  Accounting again for the fact that $\Vert \bar{\k}\Vert^2_2 \triangleq \sum_{j} k_{j}^2 \bar{I}_{ji}$ actually depends on $i$, we choose
\begin{equation} \label{eq:k_bound}
\Big( \sum_{j} k_{j}^2 \bar{I}_{ji} \Big) v_{i} - \varphi_i^*(v_i) \geq  \log \Big[ \frac{1}{\lambda} \Big(\sum_{j} k_{j}^2 \bar{I}_{ji} \Big) + \gamma_i^{-1} \Big],
\end{equation}
which holds for all $v_i \geq 0$, where $\varphi^*$ is the concave conjugate of the concave function $\varphi_i(\alpha) \triangleq  \log \left[ \frac{\alpha}{\lambda} + \gamma_i^{-1} \right]$.  Similar to the $\gam$ updates from above, it can be shown that equality in (\ref{eq:k_bound}) is obtained with the minimizing $v_i$ given by
\begin{eqnarray}\label{eq:update_v}
 \begin{split}
v_i^{\rm opt} = \left. \frac{\partial \varphi_i}{\partial \alpha} \right|_{\alpha = \sum_j k_j^2 \bar{I}_{ji}} = \frac{z_i}{\lambda}, \forall i.
 \end{split}
\end{eqnarray}
Plugging (\ref{eq:k_bound}) and (\ref{eq:update_v}) into (\ref{eq:k_fun}) leads to the quadratic optimization problem
\begin{equation}\label{eq:kernel_app}
\k^{\rm opt}  =  \arg \min_{\k\ge 0} \frac{1}{\lambda} \Vert \y - \W \k\Vert_2^2  +  \sum_i \frac{z_i}{\lambda} \left(\sum_{j} k_j^2 \bar{I}_{ji}  \right)  =  \arg \min_{\k\ge 0} \Vert \y - \W \k\Vert_2^2 +  \sum_{j} k_j^2 \left( \sum_i z_i \bar{I}_{ji} \right),
\end{equation}
%\begin{eqnarray}\label{eq:kernel_app}
%\begin{split}
%\k^{\rm opt} & =  \arg \min_{\k\ge 0} \frac{1}{\lambda} \Vert \y - \W \k\Vert_2^2  +  \sum_i \frac{z_i}{\lambda} \left(\sum_{j} k_j^2 \bar{I}_{ji}  \right)  \\
%& =  \arg \min_{\k\ge 0} \Vert \y - \W \k\Vert_2^2 +  \sum_{j} k_j^2 \left( \sum_i z_i \bar{I}_{ji} \right),
%\end{split}
%\end{eqnarray}
where $\W$ is the convolution matrix constructed from the image $\x$ (see Section~\ref{sec:notation}).  As a simple convex program, there exist many high-performance algorithms for solving~(\ref{eq:kernel_app}).

%The second equality is obtained using (\ref{eq:update_z}) and (\ref{eq:update_v}), from which it follows that  $\sum_i \tilde{v_{ij}} \equiv \sum_i z_{i+j} \bar{I}_{ji} = \sum_i C_{i+j,i+j} \bar{I}_{ji}$.

To review, we would originally like to minimize $\mathcal{L}(\x, \k, \gam)$ over $\x$, $\k$, and the latent variables $\gam$.  To simplify the optimization we introduce additional latent variables $\z \triangleq[z_1,\ldots,z_m]^T$ and $\v \triangleq [v_1,\ldots,v_m]^T$, such that, after combining terms from above we are now equivalently minimizing
\begin{eqnarray}
\begin{split}
&\mathcal{L}(\x, \k, \gam, \z, \v)  \triangleq   \frac{1}{\lambda} \Vert \y - \k\ast \x \Vert_2^2 \\
 &\qquad\qquad  +  \sum_i \left[ \frac{x_i^2 + z_i}{\gamma_i} + \log \gamma_i  + f(\gamma_i) - \phi^*(z_i) \right]  +  \sum_i \left[ \sum_{j} \left( k_{j}^2 \bar{I}_{ji} \right) v_{i} - \varphi_i^*(v_i) \right]
\end{split}
\end{eqnarray}
over $\x$, $\k$, and the latent variables $\gam$, $\z$, and $\v$. The associated coordinate descent updates rules, meaning the cyclic iteration of (\ref{eq:x_update_app}),  (\ref{eq:update_z}), (\ref{eq:gam_update_app}), (\ref{eq:update_v}), and (\ref{eq:kernel_app}), are guaranteed to reduce or leave unchanged $\mathcal{L}(\x, \k, \gam)$ by  standard properties of majorization-minimization algorithms.  And importantly, at least for our purposes, these updates are in one-to-one correspondence with those from Algorithm \ref{algo:algo1}, albeit with some inconsequential differences in notation and statistical interpretation.  Specifically, the $\gam$ update from (\ref{eq:gam_update_app}) is equivalent to the $\boldsymbol{\omega}$ update in Algorithm \ref{algo:algo1}, the $\x$ update from (\ref{eq:x_update_app}) is equivalent to the $\boldsymbol{\mu}$ update, the $\z$ update becomes equivalent to computing the diagonal of $\C$, and finally the $\k$ update from (\ref{eq:kernel_app}) is the same as that in Algorithm \ref{algo:algo1} but with the requisite boundary conditions explicitly incorporated via $\bar{\II}$.

Note that the $\boldsymbol{\omega}$ update from Algorithm \ref{algo:algo1} appears somewhat different from that originally presented in \citep{LevinWDF11}, which only considers the special case where the assumed image prior is a finite Gaussian scale mixture given by
\begin{equation}
p(x_i) = \sum_{j} \frac{\pi_j}{\sqrt{2 \pi \bar{\gamma}_j}} \exp \left[-\frac{1}{2} \frac{x_i^2}{\bar{\gamma}_j}   \right] ,
\end{equation}
where $\pi_j \geq 0$ and $\sum_j \pi_j = 1$.  However, using \citep{Wipf_VEM_NIPS05} it is easily shown that
\begin{equation}
\frac{2\sigma}{\gx'({\sigma})} =  \left( {\rm E}_{p(\gamma|x_i = \sigma)}[\gamma^{-1}] \right)^{-1} =  \frac{\sum_{j} \frac{\pi_j}{\sqrt{2 \pi \bar{\gamma}_j}} \exp \left[-\frac{1}{2} \frac{\sigma^2}{\bar{\gamma}_j}   \right]}{ \sum_{j} \frac{\pi_j}{\sqrt{2 \pi \bar{\gamma}_j}} \exp \left[-\frac{1}{2} \frac{\sigma^2}{\bar{\gamma}_j}   \right] \frac{1}{\bar{\gamma}_j }}
\end{equation}
such that formal equivalence with \citep{LevinWDF11} is maintained.

% ******************

In closing, we emphasize that the upper bounds utilized here were specifically chosen so as to establish a connection with Algorithm \ref{algo:algo1}.  However, once we have motivated that $\mathcal{L}(\x,\k,\gam)$ is an equally valid cost function, other bounds can be used to potentially improve the convergence rate or other properties of the algorithm.  This is a direction of future research.  \myendofproof

\subsection*{Proof of Corollary \ref{cor:vb_cost_reform}}
Here we omit the pixel-wise subscript $i$ for simplicity.  Likewise for later proofs where appropriate.  From the definition of $\gvb$ we know that $\gvb(x,0)  =  \min_{\gamma\ge 0} \frac{x^2}{\gamma}  + \log(\gamma ) +  f(\gamma)$.  After a $-2\log$ transformation of (\ref{eq:convex_prior}), and ignoring constant terms, we have $\gx (x) =  -2\log p(x) =  \min_{\gamma\ge 0} \frac{x^2}{\gamma}  + \log(\gamma ) +  f(\gamma)$, and so it follows that $\gvb(x, 0) = \gx (x)$.  \myendofproof

% Note that this somewhat surprising result is not nearly so obvious when we directly examine the original VB cost function from (\ref{eq:VB_DB})

\subsection*{Proof of Theorem \ref{thm:concave}}

We first assume that $f$ is a concave, non-decreasing function and express $\gvb(x, \rho)$ as
\begin{eqnarray} \label{eq:gvb_cost_new1}
\begin{split}
 \gvb(x,\rho) \triangleq \min_{\gamma \ge 0} \frac{x^2}{\gamma}  + \psi(\gamma),
\end{split}
\end{eqnarray}
where $\psi(\gamma) \triangleq \log (\rho+\gamma) + f(\gamma)$ is also a concave, non-decreasing function of $\gamma$ (because $\log (\rho+\gamma)$ is).  Thus we can always express $\psi(\gamma)$ as
\begin{equation}
\psi(\gamma) = \min_{z\ge 0} z \gamma - \psi^*(z),
\end{equation}
where $ \psi^*(z)$ is the concave conjugate \citep{cvx} of $\psi(\gamma)$.  Therefore, it follows that
\begin{equation} \label{eq:gvb_cost_new2}
\gvb(x, \rho)  = \min_{\gamma, z \ge 0} \frac{x^2}{\gamma}  + z  \gamma - \psi^*(z).
\end{equation}
Optimizing over $\gamma$ for fixed $x$ and $z$, the optimal solution is
\begin{eqnarray}
\gamma^{\rm opt} = z^{-1/2} |x|.
\end{eqnarray}
Plugging this result into (\ref{eq:gvb_cost_new2}) gives
\begin{equation}\label{eq:non_sep_reg_term}
 \gvb(x, \rho)  = \min_{z \ge 0} \frac{x^2}{z^{-1/2} |x|} + z z^{-1/2} |x|  - \psi^*(z) = \min_{z \ge 0}  2z^{1/2} |x| - \psi^*(z).
\end{equation}
This implies that $\gvb(x, \rho)$ can be expressed as a minimum over upper-bounding hyperplanes in $|x|$, with different $z$ implying different slopes.  Any function expressable in this form is necessarily concave, and also non-decreasing since \mbox{$z\ge0$} \citep{cvx}.

Now in the other direction, assume that $\gvb(x, \rho)$ is a concave, non-decreasing function of $|x|$.  It then follows that
\begin{equation}
\gvb(x, \rho) = \min_{z\geq 0} 2 z |x| + h(z)
\end{equation}
for some function $h$.  Using the fact that
\begin{equation}
2|x| = \min_{\alpha \geq 0} \frac{x^2}{\alpha} + \alpha
\end{equation}
and defining $\gamma \triangleq \alpha z^{-1}$, we can re-express $\gvb(x, \rho)$ as
\begin{eqnarray}
\gvb(x, \rho) & = & \min_{\alpha,z\geq 0} z \left[ \frac{x^2}{\alpha} + \alpha \right] + h(z)  =  \min_{\alpha,z\geq 0} \frac{x^2}{\alpha z^{-1}} + z \alpha + h(z) \nonumber \\
& = & \min_{\gamma,z\geq 0} \frac{x^2}{\gamma} + z^2 \gamma + h(z)  =  \min_{\gamma \geq 0} \frac{x^2}{\gamma} + \varphi(\gamma),
\end{eqnarray}
where $\varphi(\gamma) \triangleq \min_{z \geq 0} z^2 \gamma + h(z)$ is necessarily a concave, non-decreasing function of $\gamma$ by construction and arguments made previously.  This implies that $\psi(\gamma)$  from (\ref{eq:gvb_cost_new1}) must be a concave, non-decreasing function of $\gamma$ for all $\rho$.  Of course as $\rho \rightarrow \infty$, $\log(z + \rho)$ becomes arbitrarily flat, with derivative approaching zero for all $\gamma$.  Consequently, the only way to ensure that $\psi(\gamma)$ is concave and non-decreasing for any $\rho$ is to require that $f$ is a concave, non-decreasing function.

Finally, any locally minimizing solution $\x^{\rm opt}$ to (\ref{eq:vb_cost_reform}) must necessarily be a local minimum to
\begin{equation} \label{eq:min_over_x}
\min_{\x} \frac{1}{\lambda} \Vert \y - \HH \x \Vert_2^2 + \sum_i \gvb(x_i,\rho).
\end{equation}
If $f$ is concave and non-decreasing, then so is $\gvb(x_i,\rho)$ based on the arguments presented above, and so (\ref{eq:min_over_x}) is a canonical sparse estimation problem with a separable concave in $|\x|$ regularizer.  Based on \citep[Theorem 1]{RaoECPK03}, we may then conclude that $m-n$ elements of $\x^{\rm opt}$ will be zero at any local minimizer. \myendofproof

\subsection*{Proof of Corollary \ref{cor:concave}}
The proof in both directions follows from similar arguments to those used for proving Theorem \ref{thm:concave}.  We omit details for the sake of brevity.  \myendofproof

\subsection*{Proof of Theorem \ref{thm:general_sparse_promoting}}
Property (1) is very straightforward.  As $z \rightarrow \infty$, the optimizing $\gamma$ will become arbitrarily large regardless of the value of $\rho$.  In the regime where $\gamma$ is sufficiently large, the difference between the terms $\log(\gamma + \rho_1)$ and $\log(\gamma + \rho_2)$ must converge to zero.  It then follows that the difference between the corresponding minimizing $\gamma$ values, and therefore the cost function difference, converges to zero.

For property (2),  it will always be the case that $\grho{\rho_2}(z) - \grho{\rho_1}(z)$ for any $z$.  This occurs because for all $\gamma$, $\log(\gamma + \rho_2) > \log(\gamma + \rho_1)$.  Therefore if
\begin{equation}
\gamma_2^{\rm opt} \triangleq \arg\min_{\gamma} \frac{z}{\gamma} + \log(\gamma + \rho_2) + f(\gamma),
\end{equation}
then
\begin{equation}
\grho{\rho_2}(z) > \frac{z}{\gamma_2^{\rm opt}} + \log(\gamma_2^{\rm opt} + \rho_1) + f(\gamma_2^{\rm opt}) > \grho{\rho_1}(z).
\end{equation}
The minimizing value of $\gamma_1^{\rm opt}$ needed to produce the second inequality will always satisfy $\gamma_1^{\rm opt} < \gamma_2^{\rm opt}$.  This occurs because
\begin{equation}\nonumber
\gamma_1^{\rm opt}  =  \arg\min_{\gamma} \frac{z}{\gamma} + \log(\gamma + \rho_1) + f(\gamma)  =  \arg\min_{\gamma} \frac{z}{\gamma} + \log(\gamma + \rho_2) + \psi(\gamma) + \log \left( \frac{\gamma + \rho_1}{\gamma + \rho_2} \right).
\end{equation}
The last term, which is monotonically increasing from $\log\left(\rho_1/\rho_2 \right) < 0$ to zero, implies that there is always an extra monotonically increasing penalty on $\gamma$, when $\rho_1 < \rho_2$.  Since we are dealing with continuous functions here, the minimizing $\gamma$ will therefore necessarily be smaller.  Using results from convex analysis and conjugate duality, it can be shown that the minimizing $(\gamma_1^{\rm opt})^{-1}$ represents the gradient of $\grho{\rho_1}(z)$ with respect to $z$ (and likewise for $\gamma_2^{\rm opt}$), and we know that this gradient will always be a positive, non-increasing function.  We may therefore also infer that  $\grho{'\rho_1}(z)>\grho{'\rho_2}(z)$ at any point $z$.

We now consider a second point $z'>z$.  Because the gradient at every intermediate point moving from $\grho{\rho_1}(z)$ to $\grho{\rho_1}(z')$ is greater than the associated gradients moving from $\grho{\rho_2}(z)$ to $\grho{\rho_2}(z')$, it must be the case that $\grho{\rho_1}$ increased at a faster rate than $\grho{\rho_2}$, and so it follows that
\begin{equation}
\grho{\rho_2}(z) - \grho{\rho_1}(z) > \grho{\rho_2}(z') - \grho{\rho_1}(z'),
\end{equation}
thus completing the proof. \myendofproof

\subsection*{Proof of Theorem \ref{thm:special_case}}
For $f(\gamma) = b$, we have
\begin{eqnarray}\label{eq:g_const_f}
\begin{split}
 \gvb(x,\rho) \equiv \min_{\gamma\ge 0} \underbrace{\frac{x^2}{\gamma}  + \log(\rho + \gamma)}_{\varphi}
\end{split}
\end{eqnarray}
since constant terms are irrelevant.  We first calculate the optimal $\gamma$ by differentiating $\varphi$ and equating terms to zero.  Since
\begin{eqnarray}
\begin{split}
\frac{\partial \varphi}{\partial \gamma}= -\frac{x^2}{\gamma^2} + \frac{1}{\rho + \gamma},
\end{split}
\end{eqnarray}
it follows after some algebra that
\begin{eqnarray}\label{eq:opt_gam}
\gamma^{\rm opt} = \frac{x^2 + |x| \sqrt{x^2 + 4\rho}}{2}.
\end{eqnarray}
Based on the unimodality of $\varphi$ it follows that $\gamma^{\rm opt}$ represents the unique minimizer.  Substituting (\ref{eq:opt_gam}) into (\ref{eq:g_const_f}) and omitting irrelevant constant factors, we have
\begin{eqnarray}\nonumber
 \gvb(x,\rho) \equiv  \frac{2|x|}{|x| + \sqrt{x^2 + 4\rho}} + \log\big(2\rho + x^2 + |x|\sqrt{x^2 + 4\rho}\big).
\end{eqnarray}

\myendofproof

\subsection*{Proof of Corollary \ref{cor:special_case}}
Assuming $f(\gamma)=b$ and $\rho_1<\rho_2$, we want to show that $\grho{\rho_1} \prec \grho{\rho_2}$.  For this purpose it is sufficient to show that $\frac{\partial^2\grho{\rho}(x)}{\partial x^2}/\frac{\partial\grho{\rho}(x)}{\partial x}$ is an increasing function of $\rho$, which represents an equivalent condition for relatively concavity to one given by Definition \ref{def:1}, assuming the requisite derivatives exist \citep{RC_Palmer}.

Defining $\eta \triangleq \gamma^{-1}$, we have
\begin{eqnarray}\label{eq:g_k}
\begin{split}
 \grho{\rho}(x) &= \min_{\eta \ge 0} \eta{x^2}  + \log(\rho + \eta^{-1})
\end{split}
\end{eqnarray}
where the optimal $\eta^{\rm opt}$ is given by the gradient of $\grho{\rho}(x)$ with respect to $x^2$, which follows from basic concave duality theory.
Let $\hrho{\rho}(z)\triangleq \grho{\rho}(\sqrt{z})$.  Then $\eta^{\rm opt} =  \frac{\partial\hrho{\rho}(z)}{\partial z}$.
% Also, we have $\eta^* = (\gamma^*)^{-1}$.
With $z \triangleq x^2$, we can readily compute the expression for $\frac{\partial\grho{\rho}}{\partial x}(x)$ via
\begin{equation}\label{eq:dg}
\frac{\partial\grho{\rho}(x)}{\partial x} = \frac{\partial\hrho{\rho}(z)}{\partial z} \frac{dz}{dx} = 2x \frac{\partial\hrho{\rho}(z)}{\partial z} = \frac{x}{\rho} \left(\sqrt{1+\frac{4\rho}{x^2}} - 1\right).
\end{equation}
Using (\ref{eq:dg}) it is also straightforward to derive $\frac{\partial^2\grho{\rho}(x)}{\partial x^2}$  as
\begin{eqnarray}
\begin{split}
\frac{\partial^2\grho{\rho}(x)}{\partial x^2}&  = 2 \frac{\partial\hrho{\rho}(z)}{\partial z} - \frac{4}{x^2\sqrt{1+\frac{4\rho}{x^2}}}.
\end{split}
\end{eqnarray}
We must then show that
\begin{eqnarray}
\frac{\partial^2\grho{\rho}(x)/\partial x^2}{\partial\grho{\rho}(x)/\partial x} = \frac{1}{x} - \frac{\frac{4}{x^2\sqrt{1+\frac{4\rho}{x^2}}}}{\frac{x}{\rho} \left(\sqrt{1+\frac{4\rho}{x^2}} - 1\right)}
\end{eqnarray}
is an  increasing function of $\rho$.  By neglecting  irrelevant additive and multiplicative factors (and recall that $x\ge 0$ from the definition of $\grho{\rho}$), this is equivalent to showing that
\begin{eqnarray}
\xi(\rho) = \frac{1}{\rho} \left(\sqrt{1+\frac{4\rho}{x^2}} -1 \right)
\end{eqnarray}
is a  decreasing function of $\rho$.  It is easy to check that
\begin{eqnarray}
\begin{split}
\xi'(\rho)
%&=  1 - \sqrt{1+\frac{4\rho}{x^2}} + \frac{2\rho}{x^2\sqrt{1+\frac{4\rho^2}{x^2}}} \\
& = \frac{\sqrt{1+\frac{4\rho}{x^2}} - 1 - \frac{2\rho}{x^2}}{\sqrt{1+\frac{4\rho}{x^2}}} <0.
\end{split}
\end{eqnarray}
Therefore, $\xi(\rho)$ is a decreasing function of $\rho$, implying that $\frac{\partial^2\grho{\rho}(x)}{\partial x^2}/\frac{\partial\grho{\rho}(x)}{\partial x}$ is an increasing function of $\rho$, completing the proof. \myendofproof

\subsection*{Proof of Theorem \ref{thm:relative_concavity}}

For simplicity assume that $f$ is twice differentiable. From the definition of relative concavity, $\psirho{\rho_1}  \prec \psirho{\rho_2} $  if and only if  $\frac{\partial^2\psirho{\rho}(\gamma)}{\partial \gamma^2}/\frac{\partial\psirho{\rho}(\gamma)}{\partial \gamma}$  is
an increasing function of $\rho$ \citep{RC_Palmer}.  It is easy to show that
\begin{eqnarray}
\xi(\rho) \triangleq \frac{\partial^2\psirho{\rho}(\gamma)}{\partial \gamma^2}/\frac{\partial\psirho{\rho}(\gamma)}{\partial \gamma} = \frac{-\frac{1}{(\gamma + \rho)^2} + f''(\gamma)}{\frac{1}{\gamma + \rho} + f'(\gamma)}.
\end{eqnarray}
To avoid notation clutter, we let $\omega \triangleq \gamma + \rho$,  so that the objective is then to prove that
\begin{eqnarray}
\xi(\rho) =  \frac{-\frac{1}{\omega^2} + f''(\gamma)}{\frac{1}{\omega} + f'(\gamma)}
\end{eqnarray}
is an increasing function of $\rho$, for all $\gamma, \rho \ge 0$ if and only if $f''(\gamma) = 0$ and $f'(\gamma) \geq 0$, or equivalently that $f$ is affine with positive slope. For this purpose it suffices to examine conditions whereby
\begin{eqnarray} \label{eq:xi_gradient}
\xi'(\rho) =  \frac{f''(\gamma)\omega^2 + 2 f'(\gamma)\omega + 1}{\left(f'(\gamma) \omega^2 + \omega\right)^2} \ge 0, \forall \rho, \gamma \geq 0.
\end{eqnarray}

First, assume $f''(\gamma) = 0$.  We also have that $f'(\gamma) \geq 0$ by virtue of the Theorem statement.  Clearly (\ref{eq:xi_gradient}) will always be true and so $\xi(\rho)$ must be an increasing function of $\rho$.  In the other direction, assume that (\ref{eq:xi_gradient}) is true for all $\rho$ and $\gamma$.  Because $f$ is a concave function,  $f''(\gamma)\leq 0$.  Now consider the case where $f''(\gamma) < 0$.  The denominator of (\ref{eq:xi_gradient}) is always non-negative and can be ignored.  For the numerator, allow $\rho$ to become arbitrarily large while keeping $\gamma$ fixed.  The quadratic term will then dominate such that $\xi'(\gamma) < 0$, violating our assumption that $\xi'(\rho) \geq 0$.  Therefore it must be that \mbox{$f''(\gamma) = 0$}.

To conclude, $\psirho{\rho_1}  \prec \psirho{\rho_2} $  if and only if $f''(\gamma) = 0$ and $f'(\gamma) \geq 0$, which is equivalent to the requirement that $f(\gamma) = a\gamma+b$ with $a\geq0$.  \myendofproof

\subsection*{Proof of Theorem \ref{thm:invariance}}
Consider the VB cost function (\ref{eq:vb_cost_reform}) with $\gvb$ defined via (\ref{eq:general_f_penalty1}). Given an optimal solution pair $\{\x^*, \k^*\}$, we equivalently want to prove that $\{\alpha^{-1}\x^*, \alpha\k^*\}$ is also always an optimal solution pair if and only if $f(\gamma_i)=b$.

First we assume that $f$ is a constant.  It is easy to see that the value of the data fidelity term in (\ref{eq:vb_cost_reform}) is unchanged since
\begin{eqnarray}
\frac{1}{\lambda} \left\| \y - \k^* \ast \x^* \right\|_2^2 \equiv \frac{1}{\lambda} \left\| \y - \alpha\k^* \ast \frac{\x^*}{\alpha} \right\|_2^2.
\end{eqnarray}
For the penalty terms, after defining $\bar{\gamma}_i \triangleq \alpha^2\gamma_i$ for each $i$ and $\rho^* \triangleq \lambda/\Vert \bar{\k}^* \Vert_2^2$, we have
\begin{eqnarray}
\gvb\left(\frac{x_i^*}{\alpha},\frac{\rho^*}{\alpha^2} \right) + \log \left(\alpha^2 \left\| \bar{\k}^* \right\|_2^2 \right) & = & \min_{\gamma_i\ge 0} \frac{x_i^{*2}}{\alpha^2\gamma_i}  + \log\left(\frac{\rho^*}{\alpha^2} + \gamma_i \right) + \log \left(\alpha^2 \left\| \bar{\k}^* \right\|^2_2 \right) \nonumber \\
&= & \min_{\bar{\gamma}_i\ge 0} \frac{x_i^{*2}}{\bar{\gamma_i}}  + \log \left(\frac{\rho^*}{\alpha^2} + \frac{\bar{\gamma}_i}{\alpha^2} \right) + \log \alpha^2 + \log \left\| \bar{\k}^* \right\|^2_2 \nonumber \\
&= & \min_{\bar{\gamma}_i\ge 0} \frac{x_i^{*2}}{\bar{\gamma_i}}  + \log(\rho^* + \bar{\gamma}_i)  + \log \Vert\bar{\k}^* \Vert^2_2 \nonumber \\
& \equiv & \gvb(x_i^*,\rho^*) + \log \left( \Vert \bar{\k^*} \Vert_2^2 \right),
\end{eqnarray}
Therefore, the rescaled solution pair $\{\alpha^{-1}\x^*, \alpha\k^*\}$ does not change the cost function value, and must therefore also represent an optimal solution.

On the other hand, assume that $\{\alpha^{-1}\x^*, \alpha\k^*\}$ is an optimal solution for any $\alpha > 0$, from which it must follow that
$$
\gvb\left(\frac{x_i^*}{\alpha},\frac{\rho}{\alpha^2} \right) + \log (\alpha^2 \Vert \bar{\k}^* \Vert_2^2) = \gvb(x_i^*, \rho) + \log (\Vert \bar{\k^*} \Vert_2^2)
$$
and therefore
\begin{equation}
\min_{\bar{\gamma}_i\ge 0} \frac{x_i^{*2}}{\bar{\gamma_i}}  + \log(\rho + \bar{\gamma}_i)  + \log \Vert\bar{\k} \Vert^2_2 + f\left(\frac{\bar{\gamma}_i}{\alpha^2} \right) =  \min_{{\gamma}_i\ge 0} \frac{x_i^{*2}}{\gamma_i}  + \log(\rho + {\gamma}_i)  + \log \Vert\bar{\k} \Vert^2_2 + f({\gamma}_i).
\end{equation}
To satisfy the above equivalence for all possible $\x^*$, $\k^*$, and $\lambda$, $f$ must be a constant (with the exception of an irrelevant, zero-measure discontinuity at zero), completing the proof. \myendofproof

\section*{Appendix B: Noise Level Estimation}
%\subsection{Noise Level Estimation}

As introduced in Section \ref{sec:learn_lambda}, we would like to minimize the VB cost function (\ref{eq:vb_cost_reform}) after the inclusion of an additional $\lambda$-dependent penalty.  This is tantamount to solving
\begin{eqnarray}\label{eq:noise}\nonumber
\min_{\lambda \geq 0} \frac{1}{\lambda} \left[d + \Vert \y-\k\ast \x \Vert_2^2 \right] + n \log \lambda + \sum_i \log \Big( \frac{\Vert \bar{\k} \Vert^2_2}{\lambda} + \gamma_i^{-1} \Big),
\end{eqnarray}
where VB factors irrelevant to $\lambda$ estimation have been omitted.  We set $d = n\times10^{-4}$ for all simulations which leads to good performance.

%The last term is the additional barrier function motivated in Section \ref{sec:learn_lambda}.  In effect, it mitigates for case when $\Vert \y-\k\ast \x \Vert_2^2$ becomes to small with a proportional penalty.

While there is no closed-form, minimizing solution for $\lambda$, similar to the $\gam$ updates described in the proof of Theorem \ref{thm:vb_cost_reform}, we may utilize a convenient upper bound for optimization purposes.  Here we use
\begin{equation}
\frac{\theta}{\lambda} - \phi^*(\theta) \geq \sum_i \log \left( \frac{\Vert \bar{\k} \Vert^2_2}{\lambda} + \gamma_i^{-1} \right)
\end{equation}
where $\phi^*$ is the concave conjugate of $\phi(\theta) \triangleq \sum_i \log \left( \theta \Vert \bar{\k} \Vert^2_2 + \gamma_i^{-1} \right)$.  Equality is obtained with
\begin{eqnarray}\label{eq:noise_beta}
\theta^{\rm opt} = \left. \frac{\partial \phi}{\partial \theta} \right|_{\theta = \lambda^{-1}} = \sum_i \frac{\Vert \bar{\k} \Vert^2_2}{\frac{\Vert \bar{\k} \Vert^2_2}{\lambda} + \gamma_i^{-1}}.
\end{eqnarray}
To optimize over $\lambda$, we may iteratively solve
\begin{eqnarray}\label{eq:noise_var}
\min_{\lambda,\theta \geq 0} \frac{1}{\lambda} \left( \Vert \y-\k\ast \x \Vert_2^2 + d \right) + n \log \lambda + \frac{1}{\lambda} \theta - \phi^*(\theta).
\end{eqnarray}
For fixed $\theta$, the minimizing $\lambda$ is easily computed as
\begin{eqnarray}\label{eq:noise_rule}
\lambda^{\rm opt} = \frac{\Vert \y-\k\ast \x \Vert_2^2 + \theta + d}{n},
\end{eqnarray}
where $\lambda^{\rm opt}$ has a lower bound of $d/n$.  Thus we may set $d$ so as to reflect some expectation regarding the minimal about of noise or modeling error.  In practice, these updates can be merged into Algorithm \ref{algo:algo1} without disrupting the convergence properties (see Algorithm \ref{algo:algo_EB_BDB}).

\bibliographystyle{ieeetr}

% that's all folks
\end{document}